\pdfoutput=1

\documentclass[11pt]{article}

\usepackage[final]{acl}
\usepackage{hyperref}

\usepackage{times}
\usepackage{latexsym}

\usepackage[T1]{fontenc}

\usepackage[utf8]{inputenc}

\usepackage{microtype}

\usepackage{inconsolata}

\usepackage{graphicx}

\usepackage{amsmath}
\usepackage{bm}
\usepackage{tcolorbox}
\usepackage{tabularx}
\usepackage{booktabs}
\usepackage{subcaption}
\usepackage{setspace}
\usepackage[normalem]{ulem}
\usepackage{multirow} 
\usepackage{amsthm}
\usepackage{amsfonts}
\usepackage{mathabx}
\usepackage{amssymb}
\usepackage{colortbl}
\usepackage{tcolorbox}
\usepackage{xcolor} 
\usepackage{parskip}

\newtheorem{definition}{Definition}
\newtheorem{example}{Example}
\newtheorem{theorem}{Theorem}

\title{Bone Soups: A Seek-and-Soup Model Merging Approach \\for Controllable Multi-Objective Generation}

\author{
    Guofu Xie\textsuperscript{1},
    Xiao Zhang\textsuperscript{1}\thanks{Corresponding author:  Xiao Zhang.},
    Ting Yao\textsuperscript{2},
    Yunsheng Shi\textsuperscript{2} \\
    \textsuperscript{1}Gaoling School of Artificial Intelligence, \\
    Renmin University of China, Beijing, China \\
    \textsuperscript{2}Tencent \\
    \texttt{\{guofuxie, zhangx89\}@ruc.edu.cn},~  
    \texttt{\{tessieyao, yunshengshi\}@tencent.com}
}

\begin{document}

\maketitle

\newcommand{\guofuxie}[1]{\textcolor{blue}{#1 -- guofuxie}}
\newcommand{\zx}[1]{\textcolor{red}{#1 -- zhangxiao}}

\begin{abstract}
User information needs are often highly diverse and varied. A key challenge in current research is how to achieve controllable multi-objective generation while enabling rapid adaptation to accommodate diverse user demands during test time. Existing solutions, such as Rewarded Soup, focus on merging language models individually tuned on single objectives. While easy to implement and widely used, these approaches face limitations in achieving optimal performance due to their disregard for the impacts of competing objectives on model tuning. To address this issue, we propose \textbf{Bone Soup}, a novel model merging approach that first seeks a series of back\textbf{bone} models by considering the impacts of multiple objectives and then makes the \textbf{soup} (i.e., merge the backbone models). 
Specifically, Bone Soup begins by training multiple backbone models for different objectives using multi-objective reinforcement learning. Each backbone model is guided by a combination of backbone reward signals. To ensure that these models are optimal for the Pareto front, the backbone rewards are crafted by combining standard reward functions into basis vectors, which can then be modified through a rule-based construction method. Bone Soup leverages a symmetric circulant matrix mapping to generate the merging coefficients, which are used to merge the backbone models according to user preferences.
Extensive experimental results demonstrate that Bone Soup exhibits strong controllability and Pareto optimality in controllable multi-objective generation, providing a more effective and efficient approach to addressing diverse user needs at test time. Code is available at \url{https://github.com/andyclsr/BoneSoups}.

\end{abstract}

\section{Introduction}
\begin{figure}[t]
    \centering
    \includegraphics[width=\linewidth]{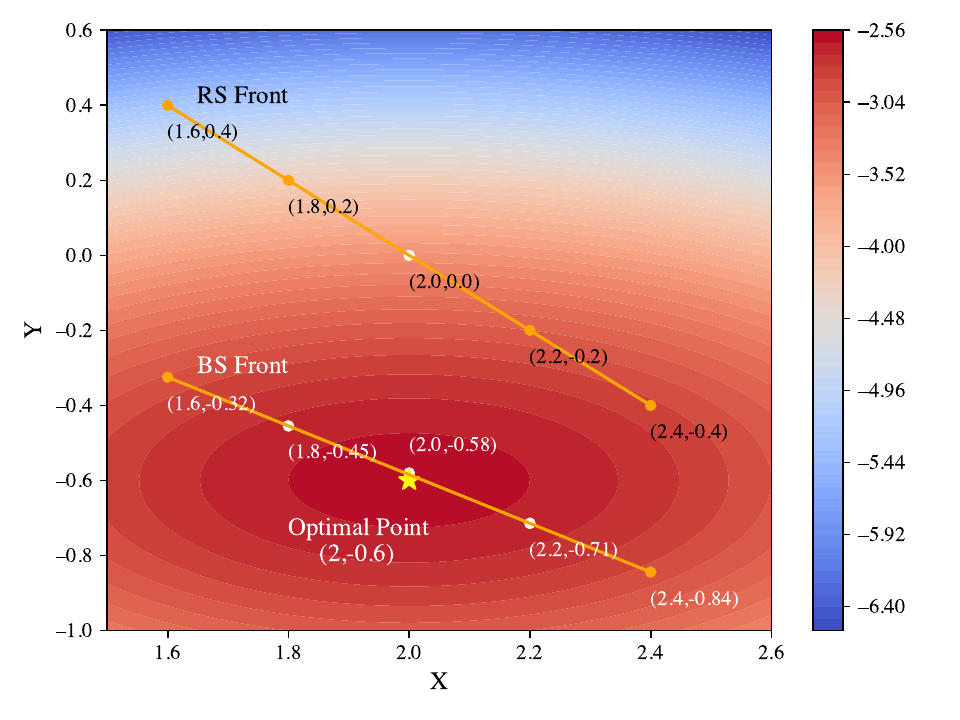}
    
    \caption{Illustration of Example 2.1. The RS front represents the front obtained by the existing soup-like approach~\cite{yang2024rewards}. The BS front represents the front obtained by our Bone Soup scheme that seeks the backbone models first. The heatmap indicates the magnitude of the testing reward as a function of two inputs $x$ and $y$. As shown in the figure, the points on the BS front are closer to the exact solution, highlighting the importance of constructing backbone models. }
    \label{fig:bonesoup:motivation_1}
\end{figure}

Human preferences and their information needs are highly diverse, and even for the same task, users may have distinct personalized demands in different scenarios~\cite{wu2024fine, rame2024rewarded, wang-etal-2024-arithmetic, shi2024decoding, Chen2024PADPA}. This diversity introduces a significant controllability challenge for AI service providers \cite{shen2024survey,chen2023controllable,shen2024generating}, who must develop learning models that can effectively adapt to a wide range of user preferences. 
A key area of research addressing this challenge is \emph{controllable multi-objective generation} (CMOG), which focuses on guiding the behavior of language models (LMs) to meet diverse and real-time user requirements without the need for retraining \cite{zhang2023survey, rame2024rewarded,shi2024decoding,wang-etal-2024-arithmetic}. 

For example, Bing Copilot offered users modes like ``More Accurate'', ``More Balanced'', and ``More Creative'', allowing for customization based on their \textit{discrete} requirements.

A straightforward approach to implementing CMOG is through prompt-based control~\cite{dong2023steerlm,ramnath2023tailoring,yang2024rewards,wang-etal-2024-arithmetic}, where LMs are guided to generate content according to user preferences for different objectives by modifying only the input prompts. These approaches can be seen as an \emph{implicit} control mechanism since it does not modify the model parameters at test time.
Recently, some \emph{explicit} approaches of controlling LMs have gained attention, known as model merging \cite{wortsman2022model, rame2024rewarded, tang2024merging, yu2024language, yadav2024ties, yang2023adamerging, wang2024localizing, ilharco2022editing}. In model merging approaches, model parameters from different LMs are combined at test time to accommodate varying user preferences. 

This form of test-time adaptation often provides more reliable control for CMOG, as it achieves control at the parameter level.

\begin{figure*}[t]
    \centering
    \includegraphics[width=\textwidth]{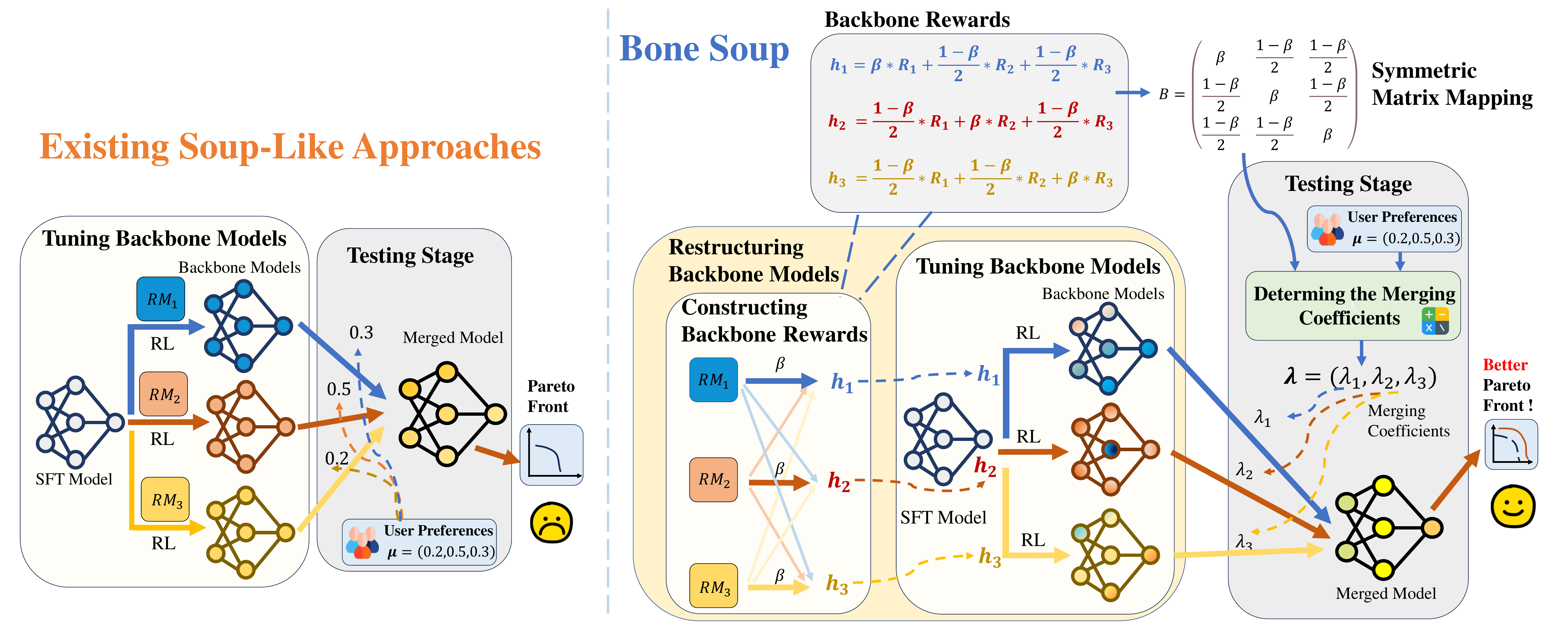}
    
    \caption{An overview of our method and a comparison between existing soup-like approaches and our Bone Soup method. Compared to existing methods, we incorporate the combined rewards and construct backbone rewards to guide the restructuring of backbone models. The merging coefficients are then determined based on the relationship between preference weights and the backbone rewards, improving the Pareto optimality and controllability of the merged model.}
    \label{fig:overview}
\end{figure*}

However, the performance of model merging heavily depends on the selection of base LMs and the determination of merging coefficients. This introduces a new challenge: \emph{how to effectively seek and merge base models based on users' preferences for multiple objectives?}  
We illustrate the existence of this challenge through an example, as shown in Figure~\ref{fig:bonesoup:motivation_1}. The figure presents two different trajectories, each interpolated from solutions optimized using distinct reward functions, accompanied by a heatmap that displays the testing rewards for user preferences. As shown, compared to solutions optimized with reward functions constructed by existing methods, \textit{there are superior solutions optimized with alternative reward functions that enable the model trajectories to more closely approximate the optimal testing reward.}

\begin{figure}[h]
    \centering
    \includegraphics[width=\linewidth]{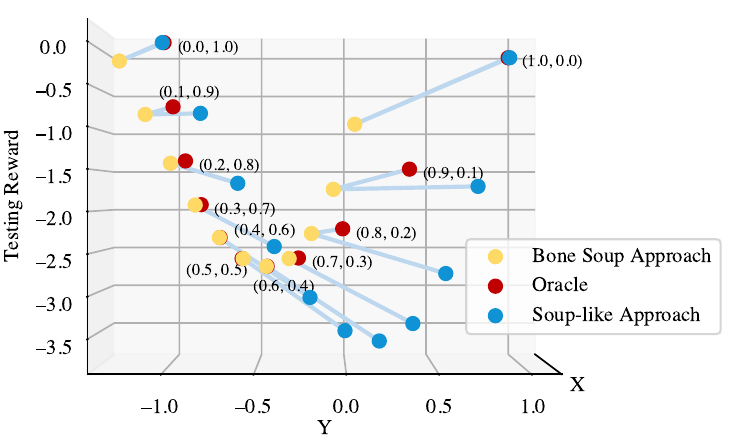}
    
    \caption{The solutions corresponding to the same preference across different methods are connected by blue lines. For each line, the closer the solution is to the red point (oracle), the better the result. Many of the yellow points in the middle are almost overlapping with the red point, indicating better solutions compared to the blue points further away. This highlights the importance of using backbone rewards to construct the backbone model.}
    \label{fig:bonesoup:motivation_3}
\end{figure}

To address the above challenge, we propose \textbf{Bone Soup}. Our proposed Bone Soup approach follows the model merging approaches seen in Rewarded Soup~\cite{rame2024rewarded} and Model Soup~\cite{wortsman2022model}. Overall, we first seek a series of backbone models, and then, based on the received user preferences at test time, we combine various backbones. Unlike Rewarded Soup, where models are tuned separately for each reward (with each reward corresponding to a specific objective) and then merged, our Bone Soup first identifies the optimal combination of rewards for different objectives. Then, these reward combinations are then used as supervision signals to train the backbone models. During inference, these backbones are adaptively merged based on given user preferences. \textit{This process is akin to selecting the right ingredients (bones) before making the soup.}
Moreover, we focus on the task of controllable multi-objective generation, where content is generated based on user-provided preference weights across different objectives at test time. In contrast, the Model Soup approach merges multiple models fine-tuned with different hyperparameters to improve model performances, while Rewarded Soup focuses on scenarios where the user's true preference (i.e., a single true label) is known, and explores how to represent it as a combination of rewards for different objectives (i.e., reward decomposition). 
We summarize our contributions as follows: 
\begin{itemize}
    \item We identify a key challenge in achieving controllable multi-objective generation through model merging, particularly when managing competing objectives, where existing approaches often fail to deliver optimal performance. 
    \item We propose Bone Soup, a novel model merging approach. By introducing combined rewards to guide the construction of backbone models, we enhance the merging process and optimize generation performance across multiple objectives, particularly in terms of controllability and Pareto optimality.
    \item Extensive experiments show that Bone Soup outperforms existing approaches, offering superior controllability, Pareto-optimal performance, and better adaptability to changes in user preferences.
\end{itemize}

\section{Problem Formulation and Analyses}
This section formulates the problem of controllable multi-objective generation through model merging and analyzes the sub-optimality of existing model merging approaches.

\subsection{Problem Formulation}
\label{sec:bonesoup:PF}
Consider $n$ objectives (e.g., factuality, relevance, completeness, etc.) that users care about, 
and each objective can be measured by a \textbf{reward function} $r_i$, $i \in \{1, 2, \ldots, n\}$.
The \textbf{preference weights} for these $n$ objectives can be represented as an $n$-dimensional vector $\bm \mu = [\mu_1, \mu_2, \ldots, \mu_n]^{\top} \in \Delta_n$, where $\Delta_n$ denotes the $n$-dimensional probability simplex.  
The problem of \textbf{controllable multi-objective generation} (CMOG) aims to enable language models (LMs) to dynamically adjust to changes in user-specified preference weights $\bm \mu$, allowing them to generate content that meets the user's requirements at test time.

To address the CMOG problem, the \textbf{model merging} approach first trains multiple base LMs using reward functions $\{r_i\}_{i=1}^n$, parameterized by $\bm \theta_i \in \Theta, i \in \{1, 2, \ldots, n\}$. Then, for satisfying user preferences, a merging strategy $\mathcal{M}$ is used to construct the model parameters for testing,  as follows:
\begin{equation}
\label{eq:bonesoup:merge_strategy}
\mathcal{M} \left( \{\bm \theta_i\}_{i=1}^n \right) = \sum_{i=1}^n \lambda_i \bm \theta_i,
\end{equation}
where $\bm \lambda = [\lambda_1, \lambda_2, \ldots, \lambda_n]^{\top}$ denotes the \textbf{merging coefficients}. 
Given an evaluation tool $\mathcal{H}$ for achieving optimal solutions, the base LMs $\{\bm \theta_i \}_{i=1}^n$ and their merging strategy aim to optimize the following expression: 
\begin{equation}
\label{eq:setup:eva}
    \operatornamewithlimits{arg\,max}_{\{\bm \theta_i \}_{i=1}^n, \mathcal{M}} \mathcal{H} \left( \mathcal{M} \left( \{\bm \theta_i\}_{i=1}^n \right) \right).
\end{equation}

In existing soup-like model merging approaches~\cite{rame2024rewarded, jang2023personalized}, for any objective $i \in \{1, 2, \ldots, n\}$, the base language model $\bm \theta_i$ is tuned with an individual reward function $r_i$ for that specific objective, making it a \textbf{specialized model} $\bm \theta_i$ for objective $i$.  
When applying these model merging approaches to CMOG, the merging coefficients in Eq.~\eqref{eq:bonesoup:merge_strategy} are directly set to the user's preference weights, i.e., $\bm \lambda = \bm \mu$, to combine the specialized models at test time. 
In Section~\ref{sec:bonesoup:PA}, we will demonstrate that merging the specialized models tuned individually with each reward does not lead to an optimal solution.

Overall, this paper explores model merging approaches for the CMOG problem, where model parameters are interpolated and merged based on user preference weights to achieve the following two goals: (1) \textit{Pareto Optimality} across multiple objectives (2) \textit{Controllability} that merged model parameters satisfy users' real-time needs.

To measure the two goals mentioned above, we define the evaluation tool $\mathcal{H}$ in Eq.~\eqref{eq:setup:eva} as the following \emph{testing reward}: 
given users' preference weights $\bm{\mu} = [\mu_1, \mu_2, \ldots, \mu_n ]^{\top}$ and corresponding rewards $\{r_i\}_{i=1}^n$, 
for merged model parameters $\bar{\bm \theta} :=\mathcal{M} \left( \{\bm \theta_i\}_{i=1}^n \right)$ defined in Eq.~\eqref{eq:bonesoup:merge_strategy}, the testing reward is defined as
\begin{equation}
    \label{eq:bonesoup:testing_reward}
    g_{\bm \mu} (\bar{\bm \theta}) := \sum_{i=1}^n \mu_i r_i(\bar{\bm \theta}).
\end{equation}
On one hand, maximizing Eq.~\eqref{eq:bonesoup:testing_reward} allows us to identify the convex Pareto front, reflecting the Pareto optimality of the merged model~\cite{zitzler1999multiobjective}. On the other hand, for any preference weights $\bm{\mu}$ provided at test time, the corresponding testing reward $g_{\bm{\mu}}$ is defined, and the merged model is required to adapt controllably to it.

\subsection{Problem Analyses}
\label{sec:bonesoup:PA}
As stated in Section~\ref{sec:bonesoup:PF}, in existing soup-like model merging approaches, the specialized models for each objective are tuned \emph{individually} with a \emph{single} reward, without considering whether incorporating other rewards could improve their training. 

\citet{rame2024rewarded} demonstrates that a global optimal solution can be derived using a single reward in certain cases, such as with quadratic rewards. However, for the CMOG problem we address, we show that individually tuning specialized models with a single reward and merging them using preference weights does not consistently yield, or even approximate, the global optimal solution.

\begin{example}
\label{exa:bonesoup:qua}
\textit{
Consider two objectives, respectively measured by the following two rewards:
\begin{align}
    r_1(x,y) &= -(x-1)^2-(y-1)^2 ~~\text{and}~~ \\
r_2(x,y) &= -(x-3)^2-4(y+1)^2,
\end{align}
which are maximized at $\bm \theta_{1}=(1,1)^{\top}$ and $\bm \theta_{2}=(3,-1)^{\top}$, respectively. 
Given preference weights $\bm \mu = (0.5, 0.5)^{\top}$ for the two rewards, 
the testing reward becomes $g_{\bm \mu} (x,y) := \bm \mu^{\top} [r_1, r_2]=0.5 \cdot r_1(x,y) + 0.5 \cdot r_2(x,y),$ and the exact solution for maximizing $g_{\bm \mu}$ is $\bm \theta^* = (2, -0.6)^{\top}$. 
However, using a soup-like approach, where the preference weights $\bm \mu$ are used to merge the individual solutions $\bm \theta_{1}$ and $\bm \theta_{2}$, the resulting solution is $\bar{\bm \theta} = 0.5 \cdot \bm \theta_1 + 0.5 \cdot \bm \theta_2 = (2, 0)^{\top}$, which significantly deviates from the exact solution $\bm \theta^*$. 
Now, instead of directly optimizing $r_1$ and $r_2$, we consider two backbone rewards that combine the rewards with different combination weights as follows:
\begin{align*}
    h_1(x,y) &= 0.4 \cdot r_1(x,y) + 0.6 \cdot r_2(x,y),~~\text{(prefer~obj~2)}\\
    h_2(x,y) &= 0.6 \cdot r_1(x,y) + 0.4\cdot r_2(x,y), ~~\text{(prefer~obj~1)}
\end{align*} 
with their respective optimal solutions, referred to as \emph{backbone models}, occurring at $\bm \theta_1^{\mathrm{bone}} = (2.2, - 5 / 7)^{\top}$ and at $\bm \theta_2^{\mathrm{bone}} = (1.8, - 5 / 11)^{\top}$.
Then, the merging solution is given by $\bar{\bm \theta}^{\mathrm{bone}} = 0.5 \cdot \bm \theta_1^{\mathrm{bone}} + 0.5 \cdot \bm \theta_2^{\mathrm{bone}} \approx (2,-0.58)^{\top}$, which is closer to the exact solution $\bm \theta^* = (2,-0.6)^{\top}$ than $\bar{\bm \theta} = (2, 0)^{\top}$.
}
\end{example}

In Example~\ref{exa:bonesoup:qua}, consider $\bm \theta_1$ and $\bm \theta_2$ as model parameters. If they are optimized solely for rewards $r_1$ and $r_2$ individually and then merged using preference weights $\bm \mu$, the result will not approximate the optimal solution $\bm \theta^*$ for the testing reward. However, if we first derive \emph{backbone rewards} $h_1$ and $h_2$ by combining the rewards, and then train \emph{backbone models} $\bm \theta_1^{\mathrm{bone}}$ and $\bm \theta_2^{\mathrm{bone}}$ on these backbone rewards, merging these backbone models with the preference weights can lead to a solution much closer to the optimal testing reward. 
Moreover, if the user's preference weights are given by $\bm{\mu}' = \{0.4, 0.6\}^{\top}$, then based on the relationship between $\bm{\mu}'$ and the combination weights in the backbone rewards $h_1$ and $h_2$, we can directly output $\bm{\theta}_1^{\mathrm{bone}}$ as the solution, obtaining the optimal solution for maximizing the testing reward $g_{\bm \mu'}$.

We also provide a comparison of the disparity between solutions of different methods and the oracle in Figure \ref{fig:bonesoup:motivation_3}.

Through the above example, we have demonstrated that BoneSoup can achieve solutions closer to the oracle. To further illustrate this point, we present the following theorem, which provides a lower bound on the interval where BoneSoup outperforms Rewarded Soup. This result proves that the front obtained by BoneSoup is, in most cases, superior to that of Rewarded Soup. We follow the setting in~\cite{rame2024rewarded} using quadratic reward functions and with Hessians proportional to identity matrices to derive the theorem.

\begin{theorem}
	Given quadratic reward functions with Hessians proportional to identity matrices:
	$$
	r_i(\theta) = r_i(\theta_i)-k_i\|\theta-\theta_i\|^2,i\in\{1,2\},
	$$where \( k_i \in \mathbb{R}_+ \) are distinct,and $\theta_i$ is the global maximum for reward $r_i$.
	Let the reward combination weight matrix be $B = \begin{pmatrix}
		\beta & 1-\beta \\
		1-\beta & \beta
	\end{pmatrix},\beta\in(\frac{1}{2},1)$, then the backbone rewards of the bone-soup approach can be denoted as $[h_1,h_2]^T = B[r_1,r_2]^T$.Let $\bm{\mu} = [\mu,1-\mu]^T$ be the user preference and the testing reward can written as $g_{\bm\mu}(\theta):=\bm\mu^T \begin{bmatrix}
	r_1 \\
	r_2
	\end{bmatrix} $.Denote the approximate solutions for the testing reward $g_{\mu}(\theta)$ of the soup-like approach and the bone-soup approach as $\bar\theta$ and $\bar\theta^{bone}$, respectively.Then,for any fixed $\beta \in (\frac{1}{2}, 1)$, when $\mu \in \left( \frac{1 - \sqrt{2\beta^2 - 2\beta + 1}}{2}, \frac{1 + \sqrt{2\beta^2 - 2\beta + 1}}{2} \right)$,
	\begin{align*}
	g_{\mu}(\bar\theta) < g_{\mu}(\bar\theta^{bone}) ,
	\end{align*}
	with interval length $\sqrt{2\beta^2 - 2\beta + 1} \geq \frac{\sqrt{2}}{2}$.
\end{theorem}

Therefore, constructing appropriate backbone rewards to train the backbone models is crucial for achieving Pareto optimality and controllability in CMOG.

\begin{proof}
    Please refer to Appendix~\ref{Appendix:proof}.
\end{proof}

\section{Bone Soup: The Proposed Approach}
In this section, we propose a novel approach to seek a series of superior backbone models, and then determine the merging coefficients for merging.

\subsection{Approach Overview}
We design and implement a more sophisticated merging-based approach Bone Soup for CMOG. Instead of directly interpolating between original base models, we propose to first \textit{seek the backbone models} which ensures better Pareto optimality, and then \textit{determine the merging coefficients} to contribute to better controllability. 

Figure~\ref{fig:overview} illustrates the overall workflow of our method.

\subsection{Restructuring the Backbone Models}
\label{sec:morlhf}

We begin by revisiting how specialized models $\bm \theta_i$ are obtained in existing works. Typically, these models are tuned through reinforcement learning from human feedback (RLHF)~\cite{stiennon2020learning,ouyang2022training,bai2022training}.

Existing soup-like model merging approaches ~\cite{jang2023personalized, rame2024rewarded} for CMOG individually tune the specialized models as above. \textit{However, when considering multi-objective reinforcement learning from human feedback (MORLHF), the approach used by existing methods represents just one specific case.}

Here, we extend tuning the backbone model from using a single reward to multiple rewards by introducing MORLHF: 
\begin{equation}
\label{eq:morl}
\bm \theta_i = \operatornamewithlimits{arg\,max}_{\pi_{\bm \theta}}\mathbb{E}_{s\thicksim\mathcal{D},a\thicksim\pi_{\bm \theta}(a|s)}\left[\bm {w}_i^{\top}\bm{r} -\eta\log\frac{\pi_{\bm \theta}(a|s)}{\pi_{\mathrm{sft}}(a|s)}\right],
\end{equation}
where $\bm{w}_i \in \Omega$ is the \textit{combination weight} of the reward models and $\Omega = \{\bm{a}\in \mathbb{R}^n| \sum_{i=1}^{n} a_i=1, a_i \geq 0\}$ is a $n$-simplex. $\bm{r}$ is a collection of all $n$ optimized reward models $\bm{r} = \{r_i(s,a)\}_{i=1}^n$. Then we define the\textit{ backbone reward} as $h_i(s,a) = \bm{w}_i^{\top} \bm r = \sum_{j=1}^n w_{i,j} \cdot r_j(s,a)$.
\textit{In this case, the single-reward setup in existing works is equivalent to setting $\bm{w}$ as a standard basis vector.}

\subsubsection{Obtaining Backbone Models}

Let $n$ denote the number of objectives to optimize,  $\bm{B} =[\bm{w}_1, \bm{w}_2, \ldots, \bm{w}_n] \in \mathbb{R}^{n \times n} $ denote a weight matrix composed of $n$ column vectors, with each column vector corresponding to the reward combination weight for tuning a new backbone model $\pi_{\bm \theta_{i}}$ in MORL as in Eq.~\eqref{eq:morl}.
Then, the combination weights $\{\bm w_i\}_{i=1}^n$ can be viewed as the basis vectors in the column space of $\bm B$, 
where $\bm w_i = \{ w_{i, j}\}_{j=1}^n$.

To simplify the search space from high-dimensional parameter space in Eq.~\eqref{eq:setup:eva} to a more manageable matrix space, we employ \textbf{a rule-based construction approach} to modify the matrix
$\bm{B}$ composed of $\{\bm{w}_i\}_{i=1}^n$ in Eq.~\eqref{eq:morl} from an identity matrix to matrices of basis vectors which achieve Pareto optimality:

\begin{align}
    \operatornamewithlimits{arg\,max}&_{\{\bm \theta_i \}_{i=1}^n, \mathcal{M}} 
    \mathcal{H} \left( \mathcal{M} \left( \{\bm \theta_i\}_{i=1}^n \right) \right) \notag \\
    \quad \longrightarrow  \quad
    &\operatornamewithlimits{arg\,max}_{\bm{B}, \mathcal{M}} 
    \mathcal{H} \left( \mathcal{M} \left( \{\bm \theta_i\}_{i=1}^n \right) \right). 
\end{align}

As mentioned earlier, introducing additional reward models may help restructure better backbone models, we introduce several rules 

to \textit{efficiently and effectively} determine matrix $\bm{B}$ :
\begin{itemize}
    \item 
    \textbf{Rule 1 (Dominance).} Each combination weight $\bm{w}_i \in \mathbb{R}^n$ should have exactly one dominant component value, denoted by $\beta_i$,  satisfying $\beta_i \in (1/n, 1)$. If we choose a small value for $\beta_i$, we will generate a set of backbone models with minor differences in abilities causing the poor \textit{Linear Mode Connectivity (LMC)}~\cite{wortsman2022model, frankle2020linear} properties and reducing controllability of the resulting solutions.
    To improve efficiency, the basis vectors should possess a similar structure and we set $\beta_i = \beta, \forall_i$. 
    \item 
    \textbf{Rule 2  (Invertibility).} Matrix $\bm{B}$ should be invertible. Since the subsequent step involves determining the merging coefficients, we require column vectors in $\bm {B}$ to be linearly independent to ensure the effectiveness of the inversion operation and to guarantee that the column space of $\bm B$ does not contain redundant information.
    \item 
    \textbf{Rule 3 (Normalization).} $\forall_{i} \sum_{j=1}^n w_{i,j} = 1$. This rule ensures that each $\bm w_i$ belongs to the $n$-simplex as defined in Eq.~\eqref{eq:morl}. 
    
\end{itemize}
To fulfill all the rules, we adopt a symmetric circulant matrix mapping. 
The \emph{symmetric circulant matrix mapping} $\bm B$ can be specified as follows: 
\begin{align}
\label{eq:Bone:B}
\bm B &:= \left[\bm{w}_1, \bm{w}_2, \ldots, \bm{w}_n\right] \notag \\
& = \begin{pmatrix}
\beta &\frac{1-\beta}{n-1}  & \dots & \frac{1-\beta}{n-1} \\
\frac{1-\beta}{n-1} & \beta & \dots & \frac{1-\beta}{n-1}  \\
\vdots & \vdots & \ddots & \vdots \\
\frac{1-\beta}{n-1}  & \frac{1-\beta}{n-1}  & \dots & \beta
\end{pmatrix} 
\in \mathbb{R}^{n \times n}.
\end{align}

In Eq.~\eqref{eq:Bone:B}, the non-dominant components are set as $(1-\beta)/(n-1)$. Taking $\bm{w}_1$ as an example, this can be interpreted as incorporating the original deterministic distribution~$\bm{o}_1 := (1, 0, \ldots, 0)^{\top}$ with a uniform distribution $\bm{u} := (1/n, 1/n, \ldots, 1/n)^{\top}$ using a mixup approach:  $\bm{w}_1 = \xi \bm{o}_1 + (1-\xi) \bm{u},$ where $\xi = (\beta n - 1)/(n-1) \in (0, 1)$.  
If we consider the basis vector $\bm{w}_i$ as a distribution for allocating rewards, this mixup method is equivalent to the exploration strategy employed in the Exp3.P algorithm~\cite{Bubeck2012Regret}. 

The next step is to select an approximate $\beta$ which is the only unknown parameter in the mapping $\bm B$. To satisfy Rule 1 and Rule 2, we constrain $\beta$ within the range $\beta \in (0.5, 1)$. Then, we train the backbone models in much smaller steps to determine which $\beta$ results in the most controllable and Pareto-optimal backbone models. Specifically, we define $\beta \in \mathcal{S}$, where $\mathcal{S}$ is a finite set with cardinality $m$, and for any $s_i \in \mathcal{S},$  $s_i$ is in the closed interval $[0.5,1]$. By adjusting $m$, we can balance the trade-off between efficiency and performance: 
$\beta = \operatornamewithlimits{arg\,max}_{\beta \in \mathcal{S}} \mathcal{H} \left( \mathcal{M_{\beta}} \left( \{\bm \theta_i\}_{i=1}^n \right) \right).$

We then use the symmetric circulant matrix mapping $\bm B$ to construct backbone rewards $h_i(s,a) = \sum_{j=1}^n w_{i,j} \cdot r_j(s,a)$ and use the reward to tune the backbone models $\{\bm \theta_i\}_{i=1}^n$.

\subsection{Determine the Merging Coefficients}
Having prepared the backbone models in the previous section, we now proceed to the merging stage. Given users' preference weights $\bm{\mu} = [\mu_1, \mu_2, \ldots, \mu_n]^\top$, our objective is to determine the merging coefficients $\bm{\lambda}$ for better controllability.

As we have trained the backbone model using backbone rewards combined with multiple rewards, a natural and straightforward approach for merging is then \textit{leveraging the reward relationship between the combination weights of backbone models and user preference weight $\bm{\mu}$} to merge the models accordingly which is achieved by mapping the combination weight vector of the backbone rewards to the user preference illustrated in Figure \ref{fig:method}. For instance, we will represent preference $\bm{\mu}$ by combination weights $\bm{w}_1$ and $\bm{w}_2$ and use the solution $\lambda_1$ and $\lambda_2$ to merge models. Specifically: $\bm \mu=\bm{B} \cdot \bm \lambda,$ and ~$\bm \lambda = \bm B^{-1} \bm \mu$ since $\bm B$ is invertible. Finally, we got the merged model parameters $\bar{\bm \theta}= \mathcal{M} \left( \{\bm \theta_i\}_{i=1}^n \right) = \sum_{i=1}^{n}\lambda_{i} \cdot \bm \theta_{i}$.

Existing soup-like model merging approaches ~\cite{jang2023personalized, rame2024rewarded} for CMOG combine specialized models linearly using $\bm{\mu}$ as the combination weight i.e. $\bm{\lambda} = \bm{\mu}$, which can also be interpreted as solving the linear equation in particular with $\bm{B}$ set as an identity matrix.

\begin{table*}[htbp]
    \centering
    \caption{Comparison of results across different methods for different trade-offs HH1~(Helpful vs Harmless), HH2~(Helpful vs Humor), and FP~(Faithful vs Preference 1).}
    \resizebox{\textwidth}{!}{
    \begin{tabular}{ccccccccccccccccccc}
        \toprule
         & \multicolumn{3}{c}{\textbf{Hypervolume \(\uparrow\)}} & \multicolumn{3}{c}{\textbf{Inner Product \(\uparrow\)}} & \multicolumn{3}{c}{\textbf{Controllability \(\uparrow\)}} & \multicolumn{3}{c}{\textbf{Length of Front \(\uparrow\)}} & \multicolumn{3}{c}{\textbf{Sparsity \(\downarrow\)}}  & \multicolumn{3}{c}{\textbf{Spacing \(\downarrow\)}} \\
        \multicolumn{1}{c}{\textbf{Method}} & HH1 & HH2 & FP & HH1 & HH2 & FP & HH1 & HH2 & FP & HH1 & HH2 & FP & HH1 & HH2 & FP & HH1 & HH2 & FP \\
        \midrule
        RS & 1.06 &\uline{ 1.12 }& 0.61 & 1.70 & \uline{1.89} & 1.13 & {0.84} & \textbf{1.00} & \textbf{1.00} & \textbf{\textbf{11}} & \textbf{\textbf{11}} & \textbf{\textbf{11}} & \uline{0.24} & \uline{0.24} & \textbf{0.17} & \uline{0.02} & \textbf{0.02} & \textbf{0.01} \\
        MOD & \uline{1.08} & 1.09 & 0.62 & \uline{1.83} & 1.85 & 1.17 & \textbf{1.00} & \textbf{1.00} & \textbf{1.00} & \textbf{11} & \textbf{11} & \textbf{11} & \uline{0.24} & \uline{0.24} & \uline{0.18} & \uline{0.02 }& \textbf{0.02 }&\uline{ 0.02} \\
        RiC & 0.45 & 0.66 & \textbf{1.23} & 1.09 & 1.52 & \textbf{2.03} & {0.85} & {0.80} & {0.82} & {8} & {6} & {6} &\textbf{ 0.07} & 0.25 & 0.39 & \textbf{0.01} & 0.07 & 0.08 \\
        \midrule
        
        Bone Soup & \textbf{1.24} & \textbf{1.24} & \uline{1.12} & \textbf{2.\textbf{11}} & \textbf{2.06} & \uline{1.89} & \textbf{1.00} & \textbf{1.00} & \textbf{1.00} & \textbf{\textbf{11}} & \textbf{\textbf{11}} & \textbf{\textbf{11}} & 0.33 & 0.27 & 0.29 & 0.07 & \uline{0.03} & 0.03 \\
        \bottomrule
    \end{tabular}
    }
    \label{tab:1}
\end{table*}

\begin{table*}[htbp]
    \centering
    \caption{Comparison of results across different methods for different trade-offs FR~(factuality vs relevance), CR~(completeness vs relevance), and FC~(factuality vs completeness).}
    \resizebox{\textwidth}{!}{
    \begin{tabular}{ccccccccccccccccccc}
        \toprule
         & \multicolumn{3}{c}{\textbf{Hypervolume \(\uparrow\)}} & \multicolumn{3}{c}{\textbf{Inner Product \(\uparrow\)}} & \multicolumn{3}{c}{\textbf{Controllability \(\uparrow\)}} & \multicolumn{3}{c}{\textbf{Length of Front \(\uparrow\)}} & \multicolumn{3}{c}{\textbf{Sparsity \(\downarrow\)}}  & \multicolumn{3}{c}{\textbf{Spacing \(\downarrow\)}} \\
        \multicolumn{1}{c}{\textbf{Method}} & FR & CR & FC & FR & CR & FC & FR & CR & FC & FR & CR & FC & FR & CR & FC & FR & CR & FC  \\
        \midrule
        MORLHF$^*$ & $0.27^*$ & $0.61^*$ & $0.16^*$ & 0.15 & 0.12 & 0.23 & 1.00 & 1.00 & 0.33 & 2 & 3 & 2 & 0.02 & 0.10 & 0.06 & 0.00 & 0.08 & 0.03 \\
        Rewarded Soups & 0.28 & 0.82 & 0.17 & 0.77 & 0.56 & 0.82 & 1.00 & 1.00 & 0.98 & 11 & 11 & 10 & 0.06 & 0.12 & 0.02 & 0.01 & 0.03 & 0.01 \\
        Bone Soup ($\beta=0.7$) & \uline{0.34} & \textbf{0.89} & 0.19 & \uline{0.81} & \textbf{0.61} & \textbf{0.88} & \uline{0.98} & \textbf{1.00} & \uline{0.98} & 10 & 11 & 10 & 0.06 & 0.13 & 0.02 & 0.01 & 0.05 & 0.01 \\
        Bone Soup & \textbf{0.35} & \uline{0.86} & \textbf{0.20} & \textbf{0.82} & \textbf{0.61} & \uline{0.85} & \textbf{1.00} & \uline{0.98} & 0.93 & 11 & 10 & 9 & 0.04 & 0.11 & 0.05 & 0.01 & 0.06 & 0.03 \\
        Bone Soup ($\beta=0.8$) & 0.33 & 0.83 & \textbf{0.21} & \textbf{0.82} & \textbf{0.61} & \textbf{0.88} & \textbf{1.00} & \textbf{1.00} & 0.96 & 11 & 11 & 10 & 0.06 & 0.14 & 0.04 & 0.01 & 0.07 & 0.02 \\

        \bottomrule
    \end{tabular}
    }
    \label{tab:2}
\end{table*}

Finally, We include the extrapolation-based approach which is firstly introduced in the paper~\cite{ilharco2022editing} to conduct unlearning or eliminate the effects on the expert model in specific tasks, and later used in ~\cite{zheng2024weak} to get a better-aligned model. We also apply extrapolation to the previously merged models as follows:

 \begin{equation}
    \hat{\bm{\theta}}^b = (1+\alpha)  \hat{\bm\theta} - \alpha \bm \theta_{\mathrm{sft}} = \hat{\bm{\theta}} + \alpha \Delta \bm{\theta},
    \label{eq:ex}
 \end{equation}
where $\bm \theta_{\mathrm{sft}}$ is the initial model used for PPO training and $\Delta \bm \theta = \hat{\bm{\theta}} - \bm{\theta_{\mathrm{sft}}}$. $\hat{ \bm {\theta}}^b$ represents the adjusted model after further diminishing the influence of the SFT model.

\section{Experiments}
\label{sec:Experiment}
In this section, we aim to evaluate the performance of Bone Soup and other latest typical controllable controllable multi-objective generation approaches.

\subsection{Experiments Setups}

\textbf{Task Setup.} We study three controllable multi-objective generation tasks using eight different rewards and two base models: \textbf{Long Form QA}~\cite{wu2024fine}, \textbf{Helpful Assistant}~\cite{bai2022training}, and \textbf{Reddit Summary}~\cite{stiennon2020learning}.  
We use the QA-Dataset~\cite{wu2024fine} and open-source reward models $\bm{R}_{\mathrm{fact}}$ (Factuality), $\bm{R}_{\mathrm{rele}}$ (Relevance), and $\bm{R}_{\mathrm{comp}}$ (Completeness), considering the trade-offs: factuality vs relevance, factuality vs completeness, and relevance vs completeness.  
For \textbf{Helpful Assistant} task, we use the HH-RLHF dataset~\cite{bai2022training,ganguli2022red} and two reward models from Huggingface $\bm{R_{\phi,1}}$ (helpful) and $\bm{R_{\phi,2}}$ (harmless) to explore trade-offs helpful vs harmless and helpful vs humor.  
Regarding \textbf{Reddit Summary} task, we use two reward models ``faithful'' and ``preference1'' trained on different datasets to evaluate human preference for summaries.
In this task, we seek controllability in trade-offs faithful vs preference1.

\textbf{Implementation Details.}
We use LLama-2 7B \cite{touvron2023llama} for Helpful Assistant task and Reddit Summary task and use T5-large~\cite{raffel2020exploring} for Long Form QA task. 

For all three tasks, we choose the best $\beta \in \{0.8,0.7,0.6\}$ by only training for 20\% total steps and evaluate the hypervolume.

As for the extrapolation of $\hat{\bm{\theta}}$, we also select the optimal $\alpha \in \{0.1, 0.2, 0.3, 0.4, 0.5\}$ using a validation datasets following the approach in \cite{zheng2024weak}. 

\textbf{Baselines.}
We consider three latest CMOG approaches including prompt-based approach Rewards-in-Context (RiC)~\cite{yang2024rewards}, decoding-time approach MOD~\cite{shi2024decoding} and merging-based method~\cite{rame2024rewarded} and follow their settings of the Hyperparameters. Detailed introduction and discussion about baselines are in Appendix \ref{baselines}.

\textbf{Evaluation Metrics.}
We provide both visualization and six numerical metrics for evaluation. To make the results more intuitive, we plot the Pareto Front of the rewards of each dimension for the evaluated set of preference vectors. 
A detailed introduction and discussion of all the metrics can be found in Appendix~\ref{sec:app:metrics}.

\begin{figure*}[ht]
    \centering
    \begin{subfigure}{0.24\textwidth}
        \centering
        \includegraphics[width=\linewidth]{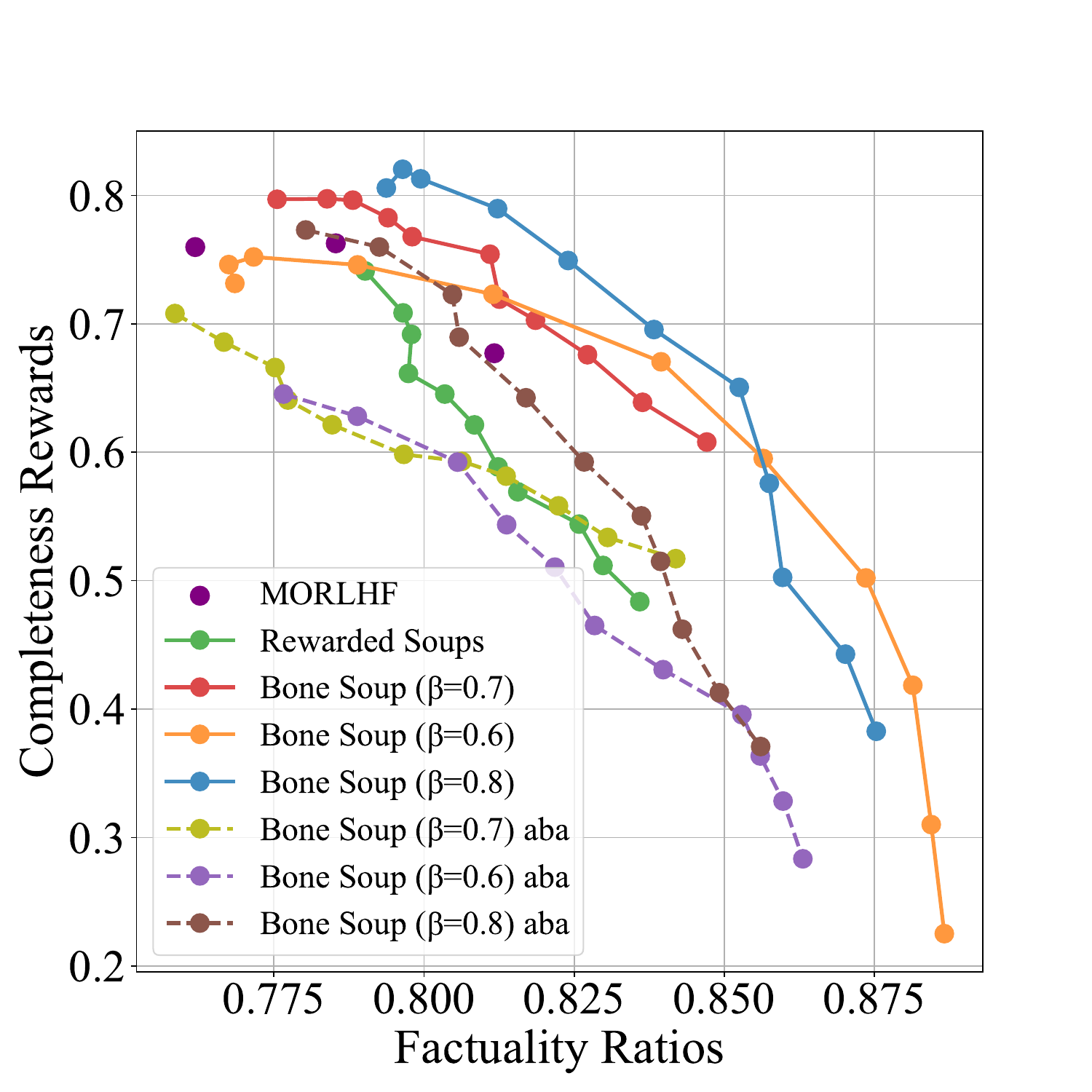} 
        \caption{factuality vs completeness}
        \label{fig:long_a}
    \end{subfigure}
    \hfill 
    \begin{subfigure}{0.24\textwidth}
        \centering
        \includegraphics[width=\linewidth]{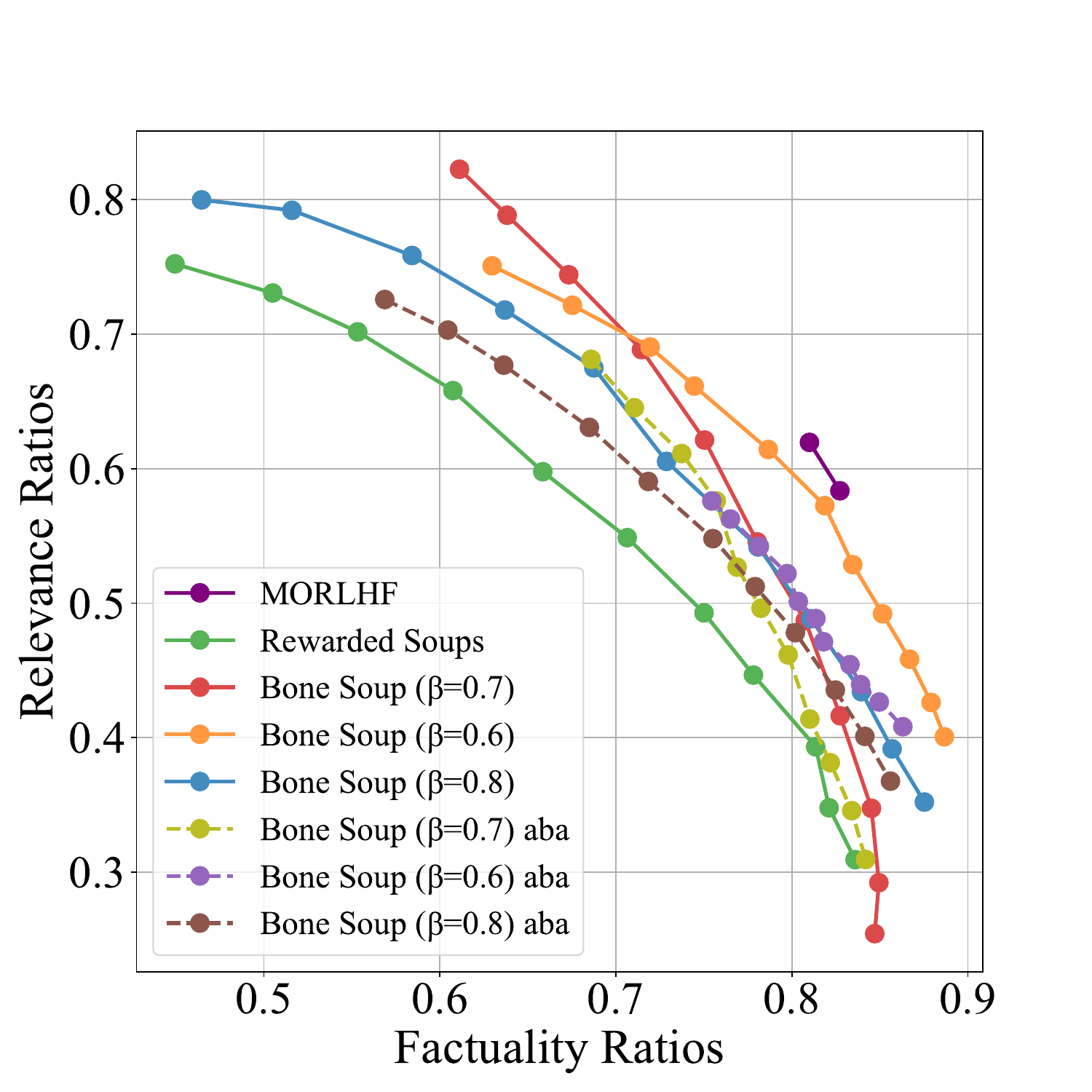} 
        \caption{factuality vs relevance}
        \label{fig:long_b}
    \end{subfigure}
    \hfill
    \begin{subfigure}{0.24\textwidth}
        \centering
        \includegraphics[width=\linewidth]{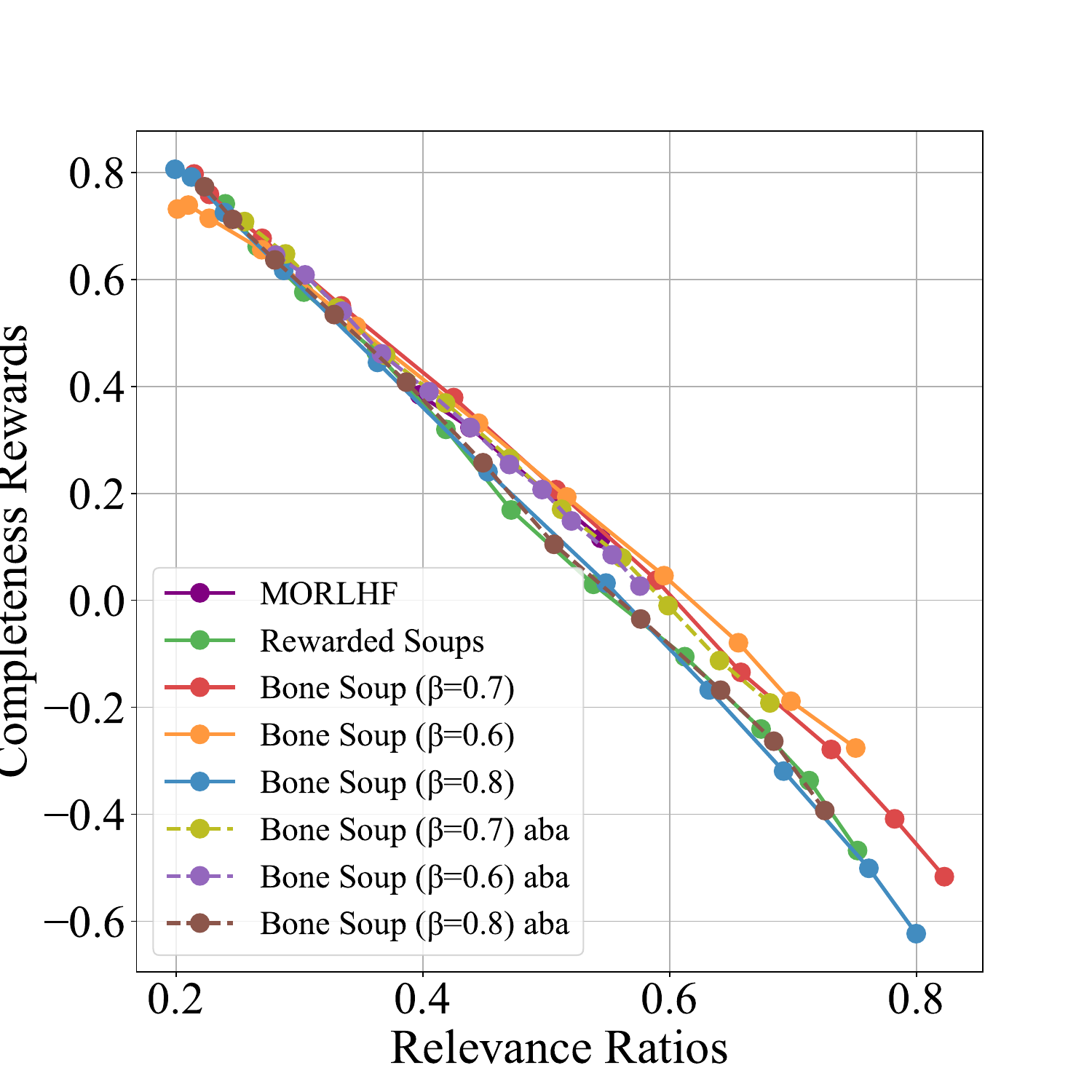} 
        \caption{relevance vs completeness}
        \label{fig:long_c}
    \end{subfigure}
    \hfill
    \begin{subfigure}{0.2\textwidth}
        \centering
        \includegraphics[width=\linewidth]{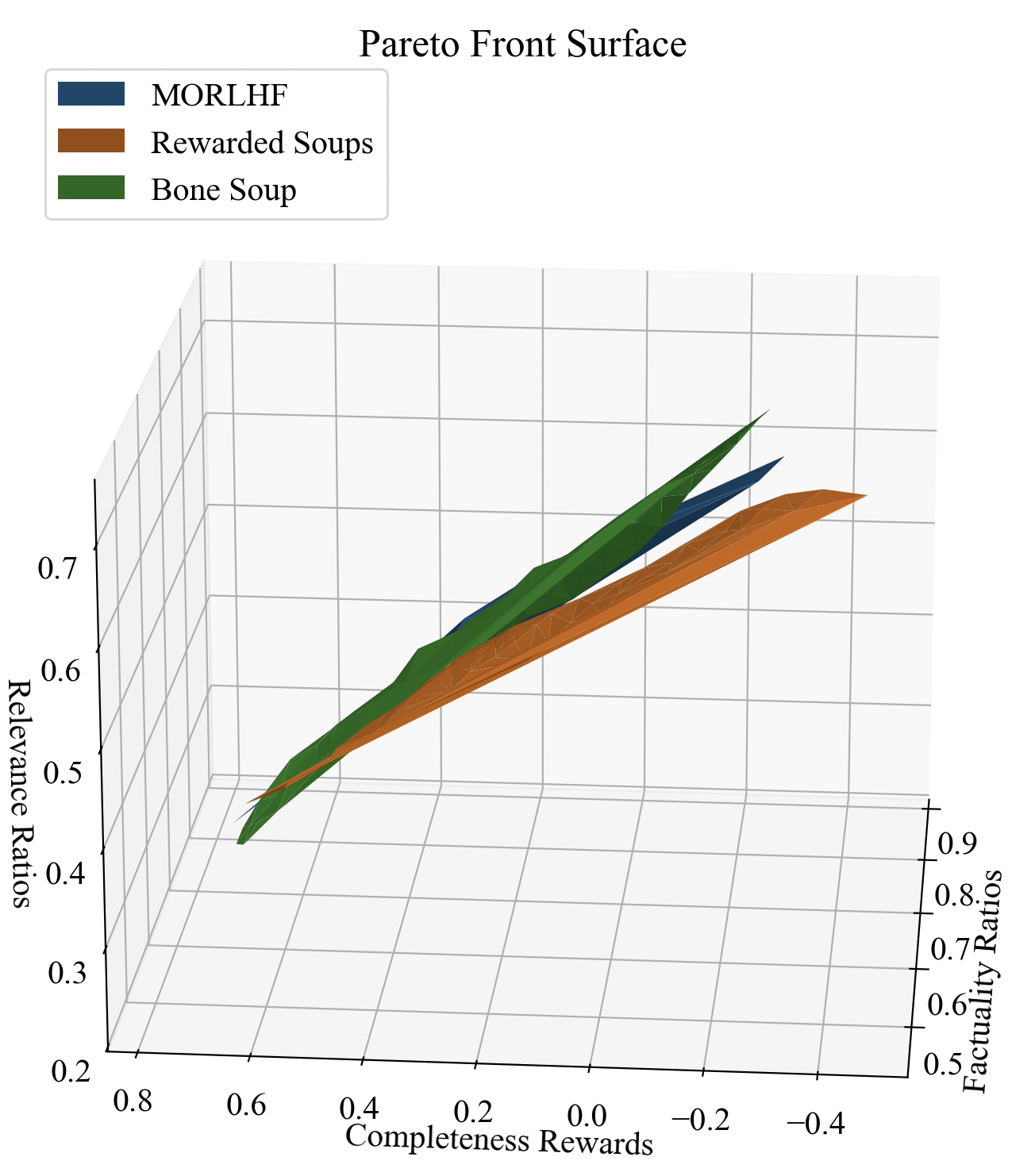} 
        \caption{factuality vs relevance vs completeness}
        \label{fig:long_d}
    \end{subfigure}
    
    \caption{Results of the Long Form QA task with (a) ``factuality vs. relevance'', (b) ``factuality vs. completeness'', (c) ``relevance vs. completeness'' and (d) ``factuality vs relevance vs. completeness''. We connect the points in the figure according to the order of the preference weight partial order relation. Bone Soup learns a better front than RS.}
    \label{fig:long}
    
\end{figure*}

\begin{figure*}[ht]
    \centering
    \begin{subfigure}{0.3\textwidth}
        \centering
        \includegraphics[width=\linewidth]{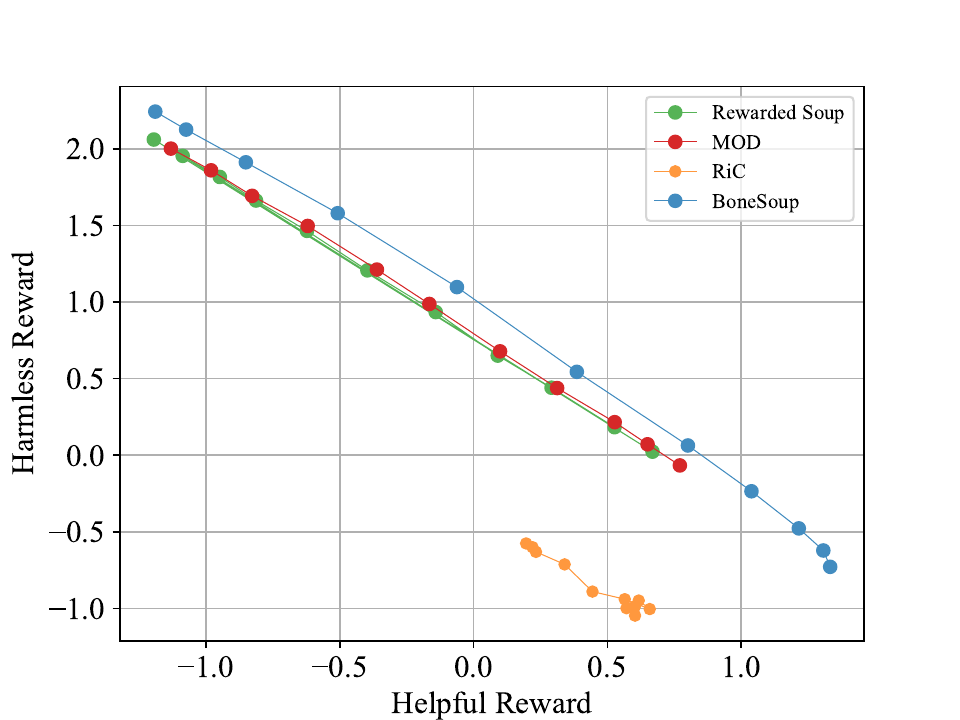} 
        \caption{helpful vs harmless}
        \label{fig:other_a}
    \end{subfigure}
    \hfill 
    \begin{subfigure}{0.3\textwidth}
        \centering
        \includegraphics[width=\linewidth]{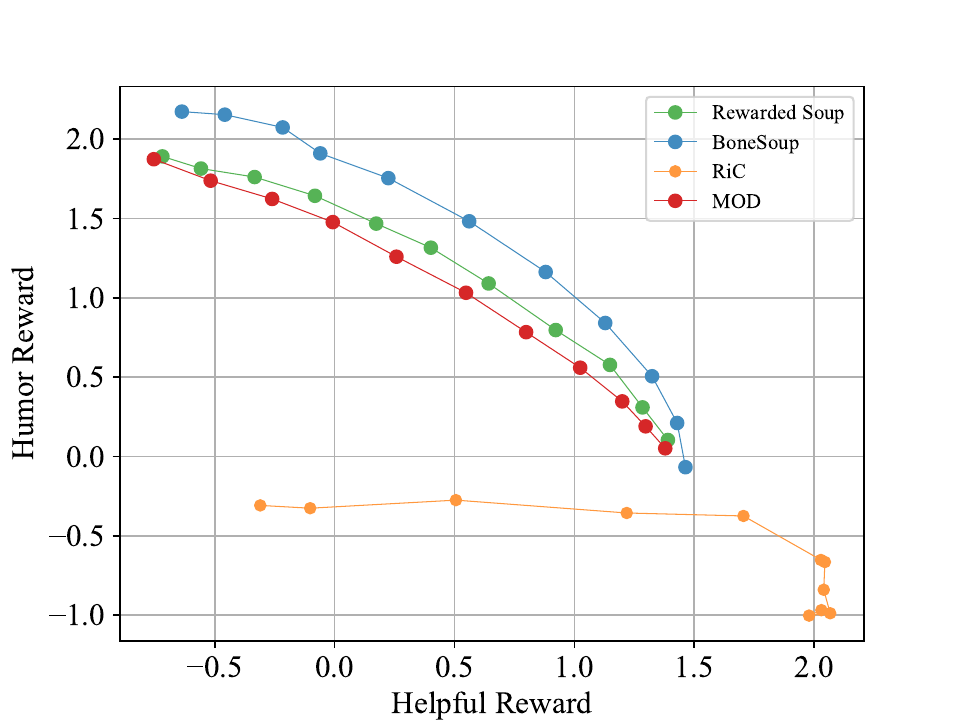} 
        \caption{helpful vs humor}
        \label{fig:other_b}
    \end{subfigure}
    \hfill
    \begin{subfigure}{0.3\textwidth}
        \centering
        \includegraphics[width=\linewidth]{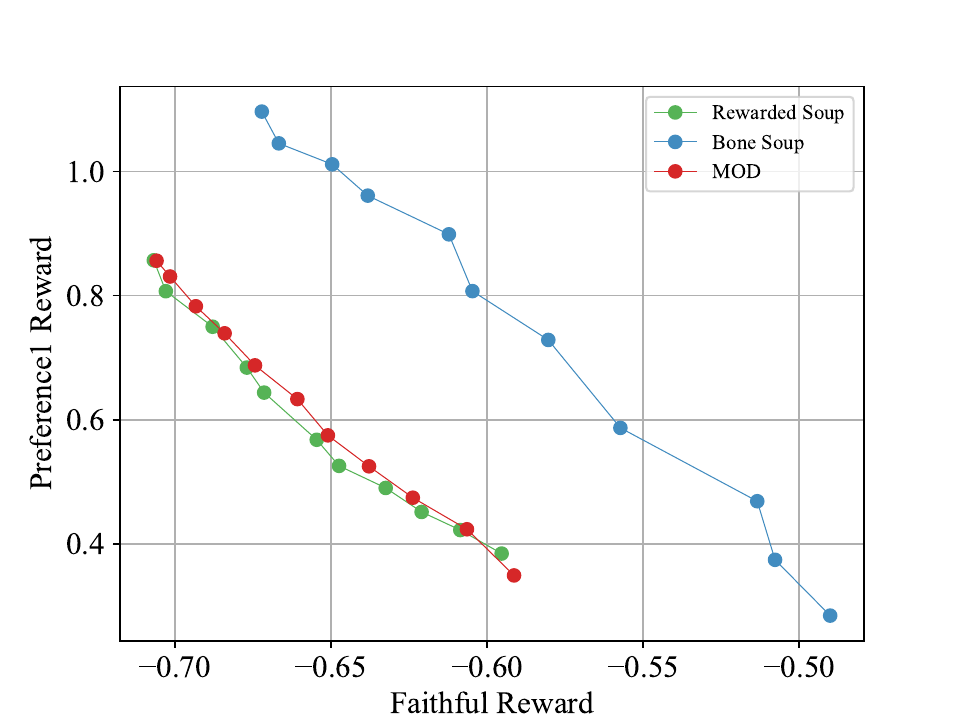} 
        \caption{faithful vs preference1}
        \label{fig:other_c}
    \end{subfigure}
    \caption{Results of Helpful Assistant task with (a) ``helpful vs. harmless'', (b) ``helpful vs. humor'', and  Reddit Summary task with (c) ``faithful vs. preference1''.}
    \label{fig:other}
\end{figure*}

\subsection{Results}
\subsubsection{Long Form QA task}
\label{sec:longform}
For the task of Long Form QA~\cite{wu2024fine},  As shown in Figure~\ref{fig:long}, each point in the front represents the average rewards of the solution corresponding to a specific user preference evaluated on test set.

In Figure~\ref{fig:long} and Table~\ref{tab:2}, we compare Bone Soup (BS) at different $\beta$ values with Rewarded Soup (RS) and MORLHF. The selection of $\beta$ is discussed in \ref{sec:app:beta1} and \ref{sec:app:beta2}. BS consistently outperforms RS and closely approximates Oracle MORLHF across three trade-offs. In factuality vs completeness and relevance vs completeness, BS even surpasses MORLHF, achieving a superior Pareto front.

Additionally, experiments in \ref{sec:app:comb} show that combining multiple rewards generally improves the backbone model. This led us to investigate merging backbone models based on user preferences, without considering the reward interplay in merging coefficients. As seen in Figure~\ref{fig:long}, the direct merge approach ("ABA") performs worse than RS in factuality vs completeness, but slightly outperforms it in the other two trade-offs, though still behind BS.

Overall, the superior performance of BS relative to MORLHF, together with the instability and suboptimality of ABA, validates the necessity and advantage of the two‐stage, seek‐and‐soup merging approach employed by Bone Soup.

Recent studies~\cite{mao2023gpteval, lambert2024rewardbench,sottana2023evaluation} have suggested that generative models can serve as unbiased evaluators—especially when ground-truth reward models are unavailable—making the use of models like GPT-4 a viable and effective evaluation approach. Therefore, in addition to using reward models, we incorporated GPT-based assessments to simulate more realistic evaluation scenarios. As shown in Figures~\ref{fig:gpt4:fact} and Figure~\ref{fig:gpt4:rele}, under the trade-off factuality vs relevance and across various user preferences, BoneSoup consistently outperforms Rewarded Soup, which is in line with our previous consequences.

We also conducted experiments in a three-objective setting. As shown in Figure~\ref{fig:long_d}, the front obtained by RS is dominated by that of MORLHF. Additionally, we observe that the front of BS is Pareto-dominant over that of MORLHF.

\subsubsection{Helpful Assistant}
\label{sec:helpful}
In this task, we focus on trade-offs ``Helpful vs Harmless'' (HH1), ``Helpful vs Humor'' (HH2). From Figure~\ref{fig:other_a}, Figure~\ref{fig:other_b} and Table~\ref{tab:1}, we can observe that the obtained front of RS approaches and MOD with similar shapes among which BS achieves the best front compared with all other baselines while RiC struggles with this task. The reason may lie in the difference between paradigms of RLHF and conditional SFT as RS and MOD all utilize the models tuned from RLHF and may obtain a similar shape.

From Figure~\ref{fig:other_a}, we can observe that Bone Soup consistently outperforms MOD which combines multiple backbone models' logits to achieve controllability. Compared to RS, both BS and MOD leverage different techniques to enhance the utilization of a set of backbone models, exploring how to better utilize these models for controllable multi-objective generation to varying degrees. However, BS provides a more comprehensive and fine-grained utilization of RLHF models therefore leading to a significantly better result.

\subsubsection{Reddit Summary}
In this task, we focus on trade-offs ``Faithful vs Preference 1'' (FP). 
From Table~\ref{tab:1} and Figure~\ref{fig:other_c}, We can see that BS significantly outperforms RS and MOD; however, in terms of hypervolume, BS falls short compared to RiC. Nevertheless, RiC performs poorly in controllability and has only 6 points on the front as shown in Figure~\ref{fig:ric}. Since the region of the front generated by RiC differs significantly from the front of BS, RS, and MOD, we therefore display RiC separately from BS, RS, and MOD for clarity.

\section{Related Work}
To be brief, our work is closely related to research on model merging, multi-objective optimization, and controllable generation. Due to space limitations, we provide a detailed discussion of these topics in Appendix~\ref{sec:related}.

\section{Conclusions}
In this work, we proposed Bone Soup, a novel model merging approach designed to address the challenges of controllable multi-objective generation. By introducing rule-based construction backbone models and combining rewards, we improved the merging process to achieve better controllability and Pareto-optimality. Extensive experiments show that Bone Soup outperforms existing methods, offering enhanced adaptability to dynamic user preferences and providing an effective and efficient solution for multi-objective generation tasks.

\section{Limitations}
Our work has the following limitations:(1) Our experiments primarily focus on controllable text generation based on human preferences, but we rely on automatic evaluators, including reward models and GPT-4, without conducting human evaluations. (2) Due to the relatively low additional complexity introduced in MORL \cite{wu2024fine}, along with the existence of multi-value head reward models \cite{wang2023helpsteer,kopf2023openassistant,wang-etal-2024-arithmetic}, our method is not significantly impacted by the number of objectives during training. As such, our approach can naturally scale to more than three objectives, but we have not conducted additional experiments with a larger number of objectives. (3)~While our model merging approach can be easily applied to fields such as computer vision and multimodal tasks, we have not conducted additional experiments to validate its performance in these areas.

\section{Acknowledgement}
This work was partially supported by the National Natural Science Foundation of China (No. 62376275, 92470205). Work partially done at Beijing Key Laboratory of Research on Large Models and Intelligent Governance, and Engineering Research Center of Next-Generation Intelligent Search and Recommendation, MOE. Supported by fund for building world-class universities (disciplines) of Renmin University of China.

\bibliography{sample}

\begin{thebibliography}{59}
\providecommand{\natexlab}[1]{#1}

\bibitem[{Achiam et~al.(2023)Achiam, Adler, Agarwal, Ahmad, Akkaya, Aleman, Almeida, Altenschmidt, Altman, Anadkat et~al.}]{achiam2023gpt}
Josh Achiam, Steven Adler, Sandhini Agarwal, Lama Ahmad, Ilge Akkaya, Florencia~Leoni Aleman, Diogo Almeida, Janko Altenschmidt, Sam Altman, Shyamal Anadkat, et~al. 2023.
\newblock Gpt-4 technical report.
\newblock \emph{arXiv preprint arXiv:2303.08774}.

\bibitem[{Bai et~al.(2022)Bai, Jones, Ndousse, Askell, Chen, DasSarma, Drain, Fort, Ganguli, Henighan et~al.}]{bai2022training}
Yuntao Bai, Andy Jones, Kamal Ndousse, Amanda Askell, Anna Chen, Nova DasSarma, Dawn Drain, Stanislav Fort, Deep Ganguli, Tom Henighan, et~al. 2022.
\newblock Training a helpful and harmless assistant with reinforcement learning from human feedback.
\newblock \emph{arXiv preprint arXiv:2204.05862}.

\bibitem[{Beltagy et~al.(2020)Beltagy, Peters, and Cohan}]{beltagy2020longformer}
Iz~Beltagy, Matthew~E Peters, and Arman Cohan. 2020.
\newblock Longformer: The long-document transformer.
\newblock \emph{arXiv preprint arXiv:2004.05150}.

\bibitem[{Bhat et~al.(2023)Bhat, Meng, Liu, Zhou, and Yavuz}]{bhat2023investigating}
Meghana~Moorthy Bhat, Rui Meng, Ye~Liu, Yingbo Zhou, and Semih Yavuz. 2023.
\newblock Investigating answerability of llms for long-form question answering.
\newblock \emph{arXiv preprint arXiv:2309.08210}.

\bibitem[{Boldi et~al.(2024)Boldi, Ding, Spector, and Niekum}]{boldi2024pareto}
Ryan Boldi, Li~Ding, Lee Spector, and Scott Niekum. 2024.
\newblock Pareto-optimal learning from preferences with hidden context.
\newblock \emph{arXiv preprint arXiv:2406.15599}.

\bibitem[{Bubeck and Cesa{-}Bianchi(2012)}]{Bubeck2012Regret}
S\'{e}bastien Bubeck and Nicol{\`{o}} Cesa{-}Bianchi. 2012.
\newblock Regret analysis of stochastic and nonstochastic multi-armed bandit problems.
\newblock \emph{Foundations and Trends$^{\textregistered}$ in Machine Learning}, 5(1):1--122.

\bibitem[{Casper et~al.(2023)Casper, Davies, Shi, Gilbert, Scheurer, Rando, Freedman, Korbak, Lindner, Freire et~al.}]{casper2023open}
Stephen Casper, Xander Davies, Claudia Shi, Thomas~Krendl Gilbert, J{\'e}r{\'e}my Scheurer, Javier Rando, Rachel Freedman, Tomasz Korbak, David Lindner, Pedro Freire, et~al. 2023.
\newblock Open problems and fundamental limitations of reinforcement learning from human feedback.
\newblock \emph{arXiv preprint arXiv:2307.15217}.

\bibitem[{Chen et~al.(2024)Chen, Zhang, Luo, Chai, and Liu}]{Chen2024PADPA}
Ruizhe Chen, Xiaotian Zhang, Meng Luo, Wenhao Chai, and Zuozhu Liu. 2024.
\newblock Pad: Personalized alignment at decoding-time.
\newblock \emph{arXiv preprint arXiv:arXiv:2410.04070}.

\bibitem[{Chen et~al.(2023)Chen, Wang, Wen, Li, Zhang, Zhang, Lin, Zhu, and Xu}]{chen2023controllable}
Sirui Chen, Yuan Wang, Zijing Wen, Zhiyu Li, Changshuo Zhang, Xiao Zhang, Quan Lin, Cheng Zhu, and Jun Xu. 2023.
\newblock Controllable multi-objective re-ranking with policy hypernetworks.
\newblock In \emph{Proceedings of the 29th ACM SIGKDD Conference on Knowledge Discovery and Data Mining}, pages 3855--3864.

\bibitem[{Christiano et~al.(2017)Christiano, Leike, Brown, Martic, Legg, and Amodei}]{christiano2017deep}
Paul~F Christiano, Jan Leike, Tom Brown, Miljan Martic, Shane Legg, and Dario Amodei. 2017.
\newblock Deep reinforcement learning from human preferences.
\newblock \emph{Advances in neural information processing systems}, 30.

\bibitem[{Dai et~al.(2023)Dai, Pan, Sun, Ji, Xu, Liu, Wang, and Yang}]{dai2023safe}
Josef Dai, Xuehai Pan, Ruiyang Sun, Jiaming Ji, Xinbo Xu, Mickel Liu, Yizhou Wang, and Yaodong Yang. 2023.
\newblock Safe rlhf: Safe reinforcement learning from human feedback.
\newblock \emph{arXiv preprint arXiv:2310.12773}.

\bibitem[{Deb et~al.(2002)Deb, Pratap, Agarwal, and Meyarivan}]{deb2002fast}
Kalyanmoy Deb, Amrit Pratap, Sameer Agarwal, and TAMT Meyarivan. 2002.
\newblock A fast and elitist multiobjective genetic algorithm: Nsga-ii.
\newblock \emph{IEEE transactions on evolutionary computation}, 6(2):182--197.

\bibitem[{Deng and Raffel(2023)}]{deng2023reward}
Haikang Deng and Colin Raffel. 2023.
\newblock Reward-augmented decoding: Efficient controlled text generation with a unidirectional reward model.
\newblock \emph{arXiv preprint arXiv:2310.09520}.

\bibitem[{Dong et~al.(2023)Dong, Wang, Sreedhar, Wu, and Kuchaiev}]{dong2023steerlm}
Yi~Dong, Zhilin Wang, Makesh~Narsimhan Sreedhar, Xianchao Wu, and Oleksii Kuchaiev. 2023.
\newblock Steerlm: Attribute conditioned sft as an (user-steerable) alternative to rlhf.
\newblock \emph{arXiv preprint arXiv:2310.05344}.

\bibitem[{Engstrom et~al.(2019)Engstrom, Ilyas, Santurkar, Tsipras, Janoos, Rudolph, and Madry}]{engstrom2019implementation}
Logan Engstrom, Andrew Ilyas, Shibani Santurkar, Dimitris Tsipras, Firdaus Janoos, Larry Rudolph, and Aleksander Madry. 2019.
\newblock Implementation matters in deep rl: A case study on ppo and trpo.
\newblock In \emph{International conference on learning representations}.

\bibitem[{Frankle et~al.(2020)Frankle, Dziugaite, Roy, and Carbin}]{frankle2020linear}
Jonathan Frankle, Gintare~Karolina Dziugaite, Daniel Roy, and Michael Carbin. 2020.
\newblock Linear mode connectivity and the lottery ticket hypothesis.
\newblock In \emph{International Conference on Machine Learning}, pages 3259--3269. PMLR.

\bibitem[{Ganguli et~al.(2022)Ganguli, Lovitt, Kernion, Askell, Bai, Kadavath, Mann, Perez, Schiefer, Ndousse et~al.}]{ganguli2022red}
Deep Ganguli, Liane Lovitt, Jackson Kernion, Amanda Askell, Yuntao Bai, Saurav Kadavath, Ben Mann, Ethan Perez, Nicholas Schiefer, Kamal Ndousse, et~al. 2022.
\newblock Red teaming language models to reduce harms: Methods, scaling behaviors, and lessons learned.
\newblock \emph{arXiv preprint arXiv:2209.07858}.

\bibitem[{Huang et~al.(2024)Huang, Liu, Thirukovalluru, Cohan, and Dhingra}]{huang2024calibrating}
Yukun Huang, Yixin Liu, Raghuveer Thirukovalluru, Arman Cohan, and Bhuwan Dhingra. 2024.
\newblock Calibrating long-form generations from large language models.
\newblock \emph{arXiv preprint arXiv:2402.06544}.

\bibitem[{Ilharco et~al.(2022)Ilharco, Ribeiro, Wortsman, Gururangan, Schmidt, Hajishirzi, and Farhadi}]{ilharco2022editing}
Gabriel Ilharco, Marco~Tulio Ribeiro, Mitchell Wortsman, Suchin Gururangan, Ludwig Schmidt, Hannaneh Hajishirzi, and Ali Farhadi. 2022.
\newblock Editing models with task arithmetic.
\newblock \emph{arXiv preprint arXiv:2212.04089}.

\bibitem[{Jang et~al.(2023)Jang, Kim, Lin, Wang, Hessel, Zettlemoyer, Hajishirzi, Choi, and Ammanabrolu}]{jang2023personalized}
Joel Jang, Seungone Kim, Bill~Yuchen Lin, Yizhong Wang, Jack Hessel, Luke Zettlemoyer, Hannaneh Hajishirzi, Yejin Choi, and Prithviraj Ammanabrolu. 2023.
\newblock Personalized soups: Personalized large language model alignment via post-hoc parameter merging.
\newblock \emph{arXiv preprint arXiv:2310.11564}.

\bibitem[{Khanov et~al.(2024)Khanov, Burapacheep, and Li}]{khanov2024args}
Maxim Khanov, Jirayu Burapacheep, and Yixuan Li. 2024.
\newblock Args: Alignment as reward-guided search.
\newblock \emph{arXiv preprint arXiv:2402.01694}.

\bibitem[{K{\"o}pf et~al.(2023)K{\"o}pf, Kilcher, Von~R{\"u}tte, Anagnostidis, Tam, Stevens, Barhoum, Nguyen, Stanley, Nagyfi et~al.}]{kopf2023openassistant}
Andreas K{\"o}pf, Yannic Kilcher, Dimitri Von~R{\"u}tte, Sotiris Anagnostidis, Zhi~Rui Tam, Keith Stevens, Abdullah Barhoum, Duc Nguyen, Oliver Stanley, Rich{\'a}rd Nagyfi, et~al. 2023.
\newblock Openassistant conversations-democratizing large language model alignment.
\newblock \emph{Advances in Neural Information Processing Systems}, 36:47669--47681.

\bibitem[{Lambert et~al.(2024)Lambert, Pyatkin, Morrison, Miranda, Lin, Chandu, Dziri, Kumar, Zick, Choi et~al.}]{lambert2024rewardbench}
Nathan Lambert, Valentina Pyatkin, Jacob Morrison, LJ~Miranda, Bill~Yuchen Lin, Khyathi Chandu, Nouha Dziri, Sachin Kumar, Tom Zick, Yejin Choi, et~al. 2024.
\newblock Rewardbench: Evaluating reward models for language modeling.
\newblock \emph{arXiv preprint arXiv:2403.13787}.

\bibitem[{Mao et~al.(2023)Mao, Chen, Zhang, Guerin, and Cambria}]{mao2023gpteval}
Rui Mao, Guanyi Chen, Xulang Zhang, Frank Guerin, and Erik Cambria. 2023.
\newblock Gpteval: A survey on assessments of chatgpt and gpt-4.
\newblock \emph{arXiv preprint arXiv:2308.12488}.

\bibitem[{Min et~al.(2020)Min, Michael, Hajishirzi, and Zettlemoyer}]{min2020ambigqa}
Sewon Min, Julian Michael, Hannaneh Hajishirzi, and Luke Zettlemoyer. 2020.
\newblock Ambigqa: Answering ambiguous open-domain questions.
\newblock \emph{arXiv preprint arXiv:2004.10645}.

\bibitem[{Mossalam et~al.(2016)Mossalam, Assael, Roijers, and Whiteson}]{mossalam2016multi}
Hossam Mossalam, Yannis~M Assael, Diederik~M Roijers, and Shimon Whiteson. 2016.
\newblock Multi-objective deep reinforcement learning.
\newblock \emph{arXiv preprint arXiv:1610.02707}.

\bibitem[{Ouyang et~al.(2022)Ouyang, Wu, Jiang, Almeida, Wainwright, Mishkin, Zhang, Agarwal, Slama, Ray et~al.}]{ouyang2022training}
Long Ouyang, Jeffrey Wu, Xu~Jiang, Diogo Almeida, Carroll Wainwright, Pamela Mishkin, Chong Zhang, Sandhini Agarwal, Katarina Slama, Alex Ray, et~al. 2022.
\newblock Training language models to follow instructions with human feedback.
\newblock \emph{Advances in neural information processing systems}, 35:27730--27744.

\bibitem[{Radford et~al.(2019)Radford, Wu, Child, Luan, Amodei, Sutskever et~al.}]{radford2019language}
Alec Radford, Jeffrey Wu, Rewon Child, David Luan, Dario Amodei, Ilya Sutskever, et~al. 2019.
\newblock Language models are unsupervised multitask learners.
\newblock \emph{OpenAI blog}, 1(8):9.

\bibitem[{Raffel et~al.(2020)Raffel, Shazeer, Roberts, Lee, Narang, Matena, Zhou, Li, and Liu}]{raffel2020exploring}
Colin Raffel, Noam Shazeer, Adam Roberts, Katherine Lee, Sharan Narang, Michael Matena, Yanqi Zhou, Wei Li, and Peter~J Liu. 2020.
\newblock Exploring the limits of transfer learning with a unified text-to-text transformer.
\newblock \emph{Journal of machine learning research}, 21(140):1--67.

\bibitem[{Rame et~al.(2023)Rame, Couairon, Dancette, Gaya, Shukor, Soulier, and Cord}]{rame2024rewarded}
Alexandre Rame, Guillaume Couairon, Corentin Dancette, Jean-Baptiste Gaya, Mustafa Shukor, Laure Soulier, and Matthieu Cord. 2023.
\newblock Rewarded soups: {T}owards {P}areto-optimal alignment by interpolating weights fine-tuned on diverse rewards.
\newblock \emph{Advances in Neural Information Processing Systems 36}.

\bibitem[{Ramnath et~al.(2023)Ramnath, Joshi, Hallinan, Lu, Li, Chan, Hessel, Choi, and Ren}]{ramnath2023tailoring}
Sahana Ramnath, Brihi Joshi, Skyler Hallinan, Ximing Lu, Liunian~Harold Li, Aaron Chan, Jack Hessel, Yejin Choi, and Xiang Ren. 2023.
\newblock Tailoring self-rationalizers with multi-reward distillation.
\newblock \emph{arXiv preprint arXiv:2311.02805}.

\bibitem[{Sanh(2019)}]{sanh2019distilbert}
V~Sanh. 2019.
\newblock Distilbert, a distilled version of bert: smaller, faster, cheaper and lighter.
\newblock \emph{arXiv preprint arXiv:1910.01108}.

\bibitem[{Schott(1995)}]{schott1995fault}
Jason~Ramon Schott. 1995.
\newblock \emph{Fault tolerant design using single and multicriteria genetic algorithm optimization}.
\newblock Ph.D. thesis, Massachusetts Institute of Technology.

\bibitem[{Schulman et~al.(2017)Schulman, Wolski, Dhariwal, Radford, and Klimov}]{schulman2017proximal}
John Schulman, Filip Wolski, Prafulla Dhariwal, Alec Radford, and Oleg Klimov. 2017.
\newblock Proximal policy optimization algorithms.
\newblock \emph{arXiv preprint arXiv:1707.06347}.

\bibitem[{Shen et~al.(2024{\natexlab{a}})Shen, Zhang, Shi, Zhang, Xie, and Xu}]{shen2024survey}
Chenglei Shen, Xiao Zhang, Teng Shi, Changshuo Zhang, Guofu Xie, and Jun Xu. 2024{\natexlab{a}}.
\newblock A survey of controllable learning: Methods and applications in information retrieval.
\newblock \emph{arXiv preprint arXiv:2407.06083}.

\bibitem[{Shen et~al.(2024{\natexlab{b}})Shen, Zhao, Zhang, Yu, He, and Fan}]{shen2024generating}
Chenglei Shen, Jiahao Zhao, Xiao Zhang, Weijie Yu, Ming He, and Jianping Fan. 2024{\natexlab{b}}.
\newblock Generating model parameters for controlling: Parameter diffusion for controllable multi-task recommendation.
\newblock \emph{arXiv preprint arXiv:2410.10639}.

\bibitem[{Shi et~al.(2024)Shi, Chen, Hu, Liu, Smith, Hajishirzi, and Du}]{shi2024decoding}
Ruizhe Shi, Yifang Chen, Yushi Hu, ALisa Liu, Noah Smith, Hannaneh Hajishirzi, and Simon Du. 2024.
\newblock Decoding-time language model alignment with multiple objectives.
\newblock \emph{Advances in Neural Information Processing Systems 37}.

\bibitem[{Siththaranjan et~al.(2023)Siththaranjan, Laidlaw, and Hadfield-Menell}]{siththaranjan2023distributional}
Anand Siththaranjan, Cassidy Laidlaw, and Dylan Hadfield-Menell. 2023.
\newblock Distributional preference learning: Understanding and accounting for hidden context in rlhf.
\newblock \emph{arXiv preprint arXiv:2312.08358}.

\bibitem[{Sottana et~al.(2023)Sottana, Liang, Zou, and Yuan}]{sottana2023evaluation}
Andrea Sottana, Bin Liang, Kai Zou, and Zheng Yuan. 2023.
\newblock Evaluation metrics in the era of gpt-4: reliably evaluating large language models on sequence to sequence tasks.
\newblock \emph{arXiv preprint arXiv:2310.13800}.

\bibitem[{Stelmakh et~al.(2022)Stelmakh, Luan, Dhingra, and Chang}]{stelmakh2022asqa}
Ivan Stelmakh, Yi~Luan, Bhuwan Dhingra, and Ming-Wei Chang. 2022.
\newblock Asqa: Factoid questions meet long-form answers.
\newblock \emph{arXiv preprint arXiv:2204.06092}.

\bibitem[{Stiennon et~al.(2020)Stiennon, Ouyang, Wu, Ziegler, Lowe, Voss, Radford, Amodei, and Christiano}]{stiennon2020learning}
Nisan Stiennon, Long Ouyang, Jeffrey Wu, Daniel Ziegler, Ryan Lowe, Chelsea Voss, Alec Radford, Dario Amodei, and Paul~F Christiano. 2020.
\newblock Learning to summarize with human feedback.
\newblock \emph{Advances in Neural Information Processing Systems}, 33:3008--3021.

\bibitem[{Tang et~al.(2024)Tang, Shen, Luo, Yin, Zhang, and Tao}]{tang2024merging}
Anke Tang, Li~Shen, Yong Luo, Nan Yin, Lefei Zhang, and Dacheng Tao. 2024.
\newblock Merging multi-task models via weight-ensembling mixture of experts.
\newblock \emph{arXiv preprint arXiv:2402.00433}.

\bibitem[{Touvron et~al.(2023)Touvron, Martin, Stone, Albert, Almahairi, Babaei, Bashlykov, Batra, Bhargava, Bhosale et~al.}]{touvron2023llama}
Hugo Touvron, Louis Martin, Kevin Stone, Peter Albert, Amjad Almahairi, Yasmine Babaei, Nikolay Bashlykov, Soumya Batra, Prajjwal Bhargava, Shruti Bhosale, et~al. 2023.
\newblock Llama 2: Open foundation and fine-tuned chat models.
\newblock \emph{arXiv preprint arXiv:2307.09288}.

\bibitem[{Wang et~al.(2024{\natexlab{a}})Wang, Lin, Xiong, Yang, Diao, Qiu, Zhao, and Zhang}]{wang-etal-2024-arithmetic}
Haoxiang Wang, Yong Lin, Wei Xiong, Rui Yang, Shizhe Diao, Shuang Qiu, Han Zhao, and Tong Zhang. 2024{\natexlab{a}}.
\newblock \href {https://doi.org/10.18653/v1/2024.acl-long.468} {Arithmetic control of {LLM}s for diverse user preferences: Directional preference alignment with multi-objective rewards}.
\newblock In \emph{Proceedings of the 62nd Annual Meeting of the Association for Computational Linguistics (Volume 1: Long Papers)}, pages 8642--8655, Bangkok, Thailand. Association for Computational Linguistics.

\bibitem[{Wang et~al.(2024{\natexlab{b}})Wang, Dimitriadis, Ortiz-Jimenez, Fleuret, and Frossard}]{wang2024localizing}
Ke~Wang, Nikolaos Dimitriadis, Guillermo Ortiz-Jimenez, Fran{\c{c}}ois Fleuret, and Pascal Frossard. 2024{\natexlab{b}}.
\newblock Localizing task information for improved model merging and compression.
\newblock \emph{arXiv preprint arXiv:2405.07813}.

\bibitem[{Wang et~al.(2023)Wang, Dong, Zeng, Adams, Sreedhar, Egert, Delalleau, Scowcroft, Kant, Swope et~al.}]{wang2023helpsteer}
Zhilin Wang, Yi~Dong, Jiaqi Zeng, Virginia Adams, Makesh~Narsimhan Sreedhar, Daniel Egert, Olivier Delalleau, Jane~Polak Scowcroft, Neel Kant, Aidan Swope, et~al. 2023.
\newblock Helpsteer: Multi-attribute helpfulness dataset for steerlm.
\newblock \emph{arXiv preprint arXiv:2311.09528}.

\bibitem[{Wortsman et~al.(2022)Wortsman, Ilharco, Gadre, Roelofs, Gontijo-Lopes, Morcos, Namkoong, Farhadi, Carmon, Kornblith et~al.}]{wortsman2022model}
Mitchell Wortsman, Gabriel Ilharco, Samir~Ya Gadre, Rebecca Roelofs, Raphael Gontijo-Lopes, Ari~S Morcos, Hongseok Namkoong, Ali Farhadi, Yair Carmon, Simon Kornblith, et~al. 2022.
\newblock Model soups: averaging weights of multiple fine-tuned models improves accuracy without increasing inference time.
\newblock In \emph{International conference on machine learning}, pages 23965--23998. PMLR.

\bibitem[{Wu et~al.(2024)Wu, Hu, Shi, Dziri, Suhr, Ammanabrolu, Smith, Ostendorf, and Hajishirzi}]{wu2024fine}
Zeqiu Wu, Yushi Hu, Weijia Shi, Nouha Dziri, Alane Suhr, Prithviraj Ammanabrolu, Noah~A Smith, Mari Ostendorf, and Hannaneh Hajishirzi. 2024.
\newblock Fine-grained human feedback gives better rewards for language model training.
\newblock \emph{Advances in Neural Information Processing Systems}, 36.

\bibitem[{Yadav et~al.(2024)Yadav, Tam, Choshen, Raffel, and Bansal}]{yadav2024ties}
Prateek Yadav, Derek Tam, Leshem Choshen, Colin~A Raffel, and Mohit Bansal. 2024.
\newblock Ties-merging: Resolving interference when merging models.
\newblock \emph{Advances in Neural Information Processing Systems}, 36.

\bibitem[{Yang et~al.(2023)Yang, Wang, Shen, Liu, Guo, Wang, and Tao}]{yang2023adamerging}
Enneng Yang, Zhenyi Wang, Li~Shen, Shiwei Liu, Guibing Guo, Xingwei Wang, and Dacheng Tao. 2023.
\newblock Adamerging: Adaptive model merging for multi-task learning.
\newblock \emph{arXiv preprint arXiv:2310.02575}.

\bibitem[{Yang et~al.(2024)Yang, Pan, Luo, Qiu, Zhong, Yu, and Chen}]{yang2024rewards}
Rui Yang, Xiaoman Pan, Feng Luo, Shuang Qiu, Han Zhong, Dong Yu, and Jianshu Chen. 2024.
\newblock Rewards-in-context: Multi-objective alignment of foundation models with dynamic preference adjustment.
\newblock \emph{International Conference on Machine Learning}.

\bibitem[{Yu et~al.(2024)Yu, Yu, Yu, Huang, and Li}]{yu2024language}
Le~Yu, Bowen Yu, Haiyang Yu, Fei Huang, and Yongbin Li. 2024.
\newblock Language models are super mario: Absorbing abilities from homologous models as a free lunch.
\newblock In \emph{Forty-first International Conference on Machine Learning}.

\bibitem[{Zhang et~al.(2023)Zhang, Song, Li, Zhou, and Song}]{zhang2023survey}
Hanqing Zhang, Haolin Song, Shaoyu Li, Ming Zhou, and Dawei Song. 2023.
\newblock A survey of controllable text generation using transformer-based pre-trained language models.
\newblock \emph{ACM Computing Surveys}, 56(3):1--37.

\bibitem[{Zheng et~al.(2024)Zheng, Wang, Ji, Huang, and Peng}]{zheng2024weak}
Chujie Zheng, Ziqi Wang, Heng Ji, Minlie Huang, and Nanyun Peng. 2024.
\newblock Weak-to-strong extrapolation expedites alignment.
\newblock \emph{arXiv preprint arXiv:2404.16792}.

\bibitem[{Zheng et~al.(2023)Zheng, Dou, Gao, Hua, Shen, Wang, Liu, Jin, Liu, Zhou et~al.}]{zheng2023secrets}
Rui Zheng, Shihan Dou, Songyang Gao, Yuan Hua, Wei Shen, Binghai Wang, Yan Liu, Senjie Jin, Qin Liu, Yuhao Zhou, et~al. 2023.
\newblock Secrets of rlhf in large language models part i: Ppo.
\newblock \emph{arXiv preprint arXiv:2307.04964}.

\bibitem[{Zhong et~al.(2024)Zhong, Ma, Zhang, Yang, Chen, Zhang, Qi, and Yang}]{zhong2024panacea}
Yifan Zhong, Chengdong Ma, Xiaoyuan Zhang, Ziran Yang, Haojun Chen, Qingfu Zhang, Siyuan Qi, and Yaodong Yang. 2024.
\newblock Panacea: Pareto alignment via preference adaptation for llms.
\newblock \emph{arXiv preprint arXiv:2402.02030}.

\bibitem[{Zhou et~al.(2024)Zhou, Liu, Shao, Yue, Yang, Ouyang, and Qiao}]{zhou-etal-2024-beyond}
Zhanhui Zhou, Jie Liu, Jing Shao, Xiangyu Yue, Chao Yang, Wanli Ouyang, and Yu~Qiao. 2024.
\newblock \href {https://doi.org/10.18653/v1/2024.findings-acl.630} {Beyond one-preference-fits-all alignment: Multi-objective direct preference optimization}.
\newblock In \emph{Findings of the Association for Computational Linguistics: ACL 2024}, pages 10586--10613, Bangkok, Thailand. Association for Computational Linguistics.

\bibitem[{Zitzler and Thiele(1999)}]{zitzler1999multiobjective}
Eckart Zitzler and Lothar Thiele. 1999.
\newblock Multiobjective evolutionary algorithms: {A} comparative case study and the strength {P}areto approach.
\newblock \emph{IEEE transactions on Evolutionary Computation}, 3(4):257--271.

\bibitem[{Zitzler et~al.(2003)Zitzler, Thiele, Laumanns, Fonseca, and Da~Fonseca}]{zitzler2003performance}
Eckart Zitzler, Lothar Thiele, Marco Laumanns, Carlos~M Fonseca, and Viviane~Grunert Da~Fonseca. 2003.
\newblock Performance assessment of multiobjective optimizers: An analysis and review.
\newblock \emph{IEEE Transactions on evolutionary computation}, 7(2):117--132.

\end{thebibliography}

\appendix

\section{Appendix}

\subsection{Related Work}
\label{sec:related}

\subsubsection{Multi-Objective Optimization and Generation}

Reinforcement Learning with Human Feedback (RLHF)~\cite{christiano2017deep, stiennon2020learning, ouyang2022training}, consisting of two stages—reward modeling and reinforcement learning—has become a powerful tool to align large language models (LLMs) with human preferences. Many existing models ~\cite{touvron2023llama,achiam2023gpt} utilize RLHF to enhance their performance. However, optimizing toward a single reward has notable limitations, such as its inability to handle complex, multifaceted preferences~\cite{casper2023open}, the challenge of satisfying all preferences with a single reward~\cite{jang2023personalized,rame2024rewarded}, and issues related to fairness in alignment\cite{siththaranjan2023distributional,boldi2024pareto,rame2024rewarded}. To address these shortcomings, multi-objective RLHF (MORLHF) has been introduced.

One of the most straightforward ways to adapt RLHF for multiple objectives is to combine all rewards linearly~\cite{mossalam2016multi}. However, due to the inefficiency of this approach in MORLHF, this paradigm struggles to quickly adapt to different preferences and achieve controllable multi-objective generation. 
Recently, an increasing number of studies have focused on controllable multi-objective generation. Methods for controllable multi-objective generation can be categorized into three main stages: pre-processing, in-processing, and post-processing. Pre-processing methods, like SteerLM~\cite{dong2023steerlm}, DPA~\cite{wang-etal-2024-arithmetic}, and RiC~\cite{yang2024rewards}, implement control through prompts, introducing multi-dimensional reward conditions. These methods use supervised fine-tuning to train the model to control outputs by prompts. The fine-tuning strategies and condition representations vary across methods, including rejection-sampling-based fine-tuning~\cite{wang-etal-2024-arithmetic, yang2024rewards} and representing conditions as unit vectors~\cite{wang-etal-2024-arithmetic} or by theoretical guarantee mapping~\cite{yang2024rewards}.

In-processing methods~\cite{rame2024rewarded, jang2023personalized} focus on model merging, where specialized models are combined using different merge coefficients to quickly generate models that cater to various preferences. This approach is straightforward to implement and computationally efficient.

Post-processing methods, such as Controlled Text Generation (CTG), primarily involve decoding-time algorithms~\cite{khanov2024args, deng2023reward, shi2024decoding}. These methods generate the next token by taking a linear combination of predictions from multiple base models, based on different objective weightings. Reward signals are used to find the optimal merging coefficients. For instance, MOD~\cite{shi2024decoding} identifies a closed-form solution using the Legendre transform, deriving an efficient decoding strategy, while ARGS~\cite{khanov2024args} and RAD~\cite{deng2023reward} achieves alignment by reward-guided search.

This paper focuses on introducing control during the in-processing phase, incorporating explicit control mechanisms into the model parameters to enable controllable generation.

\subsubsection{Model Merging}
We denote the policy LLM as $\pi_{\bm \theta}$ whose parameters are $\bm \theta \in \Theta \subseteq \mathbb{R}^d$. $\mathcal{X}$ and $\mathcal{Y}$ represent the input space (prompt space) and output space individually. 
We have summarized the current model merging techniques into the following three steps: \emph{determining the base models, merging the backbone models, and calibration after model merging}. We mainly focus on discussing the first two stages.

\emph{Determining the base models}, i.e., identifying the parameter space for interpolation. Denote the models to merge in the following step as $\{\pi_{\bm \theta_i}\}_{i=1}^m$. Here, it is generally assumed that the number of models to be merged is equal to the number of objectives or tasks, i.e., $m=n$. Moreover, these models are typically trained using a single loss~\cite{ilharco2022editing,yu2024language} or reward~\cite{wu2024fine, jang2023personalized}, meaning they can be regarded as expert models corresponding to each task or objective.

\emph{Merging the base models}.  
After obtaining $n$ specializing (expert models in a multi-task setting) with different focuses, the next step is to determine the interpolation coefficients $\lambda$ for model merging, $\bm \theta_{\mathrm{target}} = \sum_{i=1}^n \lambda_i \cdot \bm \theta_i$. Rewarded Soup~\cite{rame2024rewarded} proposes to merge the models optimized individually against a single objective. And $\lambda_i \in \Omega$ and $\Omega = \{\lambda_i \in \mathbb{R}^k| \sum_{i=1}^{n} \lambda_i=1, \lambda_i \geq 0\}$.

In the field of multi-task learning, various model merging approaches have been proposed. \citet{tang2024merging} using a dynamic routing mechanism trained in test-time adaptation to determine the model fusion coefficients. Some approaches exploit model parameter redundancy, leading to pruning-based approaches~\cite{yadav2024ties,yu2024language}.

AdaMerging~\cite{yang2023adamerging} employs an unsupervised approach to train merging coefficients and adjusts the merging coefficients at different layers of the models. 

TALL-masks \cite{wang2024localizing} generates a task-specific mask matrix using a predefined threshold derived from independent models. 

The key distinction between our approach and the above works lies in that they are not designed for, nor capable of, achieving controllable generation. 

In contrast, we have developed a series of techniques specifically aimed at optimizing for Pareto optimality and controllability.

\begin{figure}[t]
    \centering
    \includegraphics[width=\linewidth]{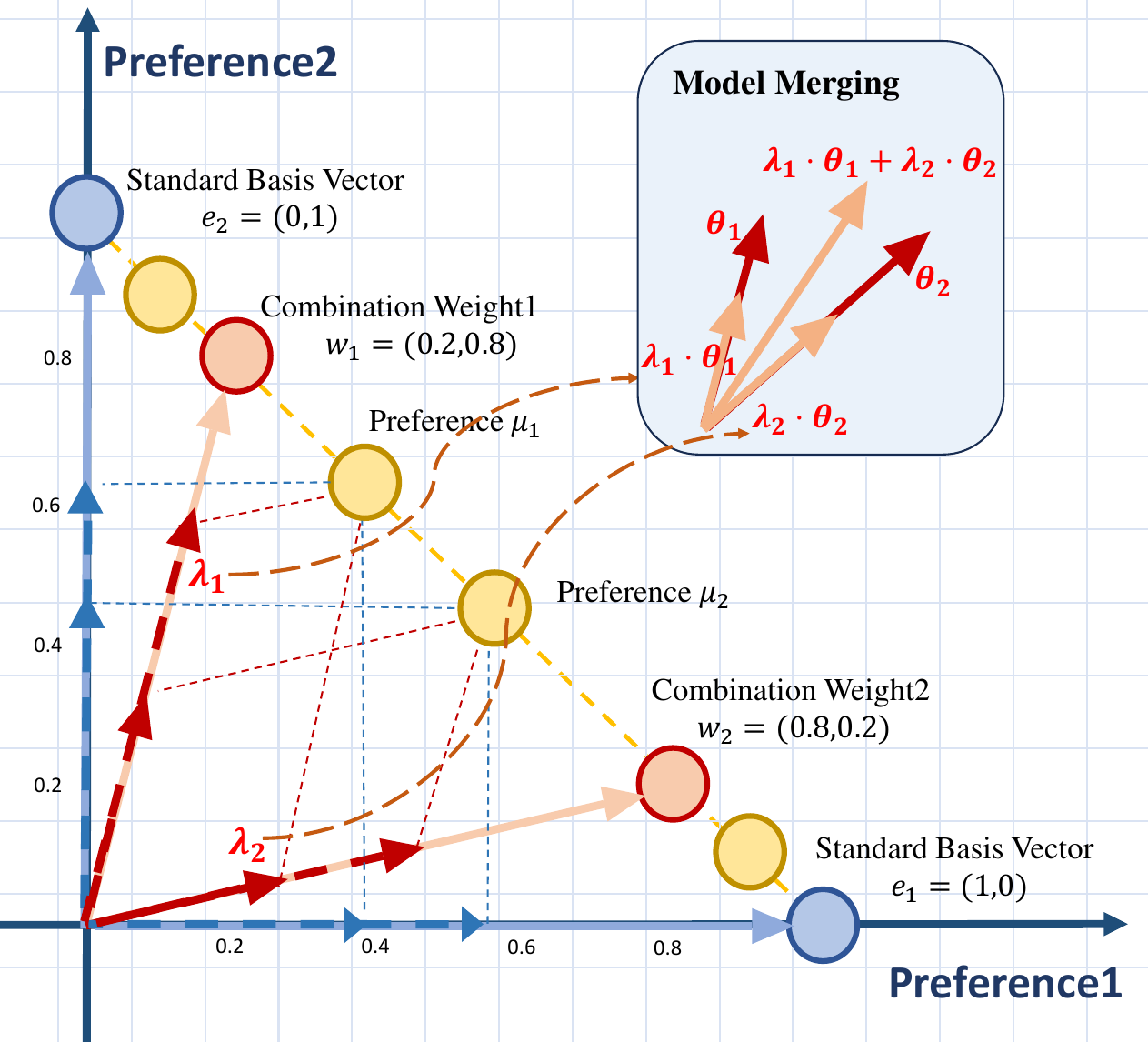}
    \caption{The illustration of how to leverage the relationship between user preferences and the rewards of backbone models to obtain the mapping between the target merged model and the backbone models.}

    \label{fig:method}
\end{figure}

\begin{figure}[t]
    \centering
    \includegraphics[width=\linewidth]{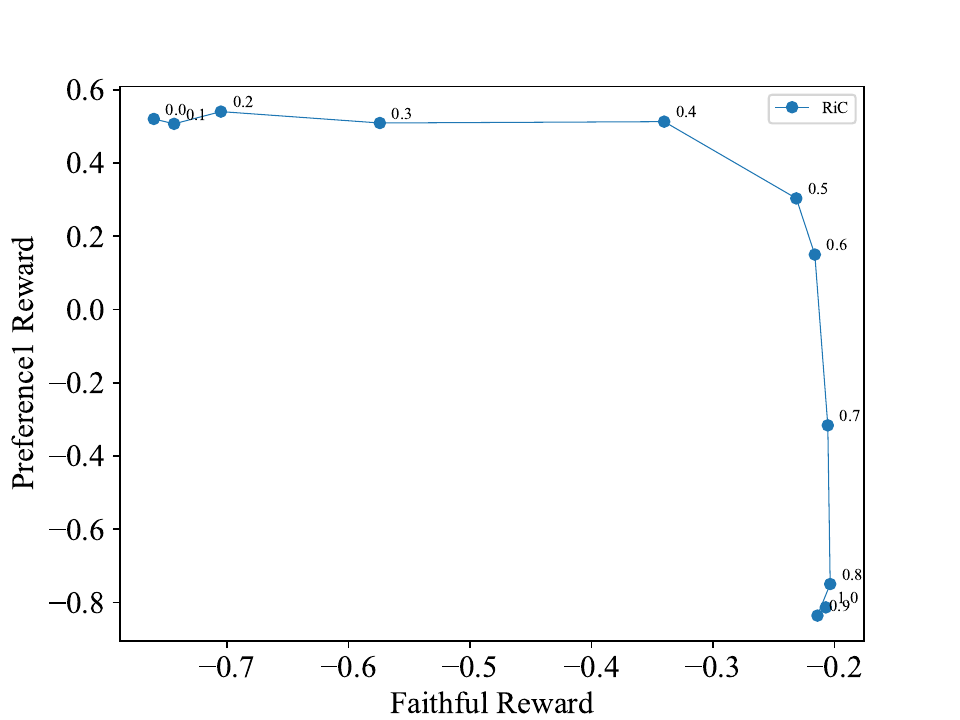}
    
    \caption{The front learned from RiC in trade-offs·`` faithful vs preference1'' (FP)}
    \label{fig:ric}
\end{figure}

\begin{figure}[t]
    \centering
    \includegraphics[width=\linewidth]{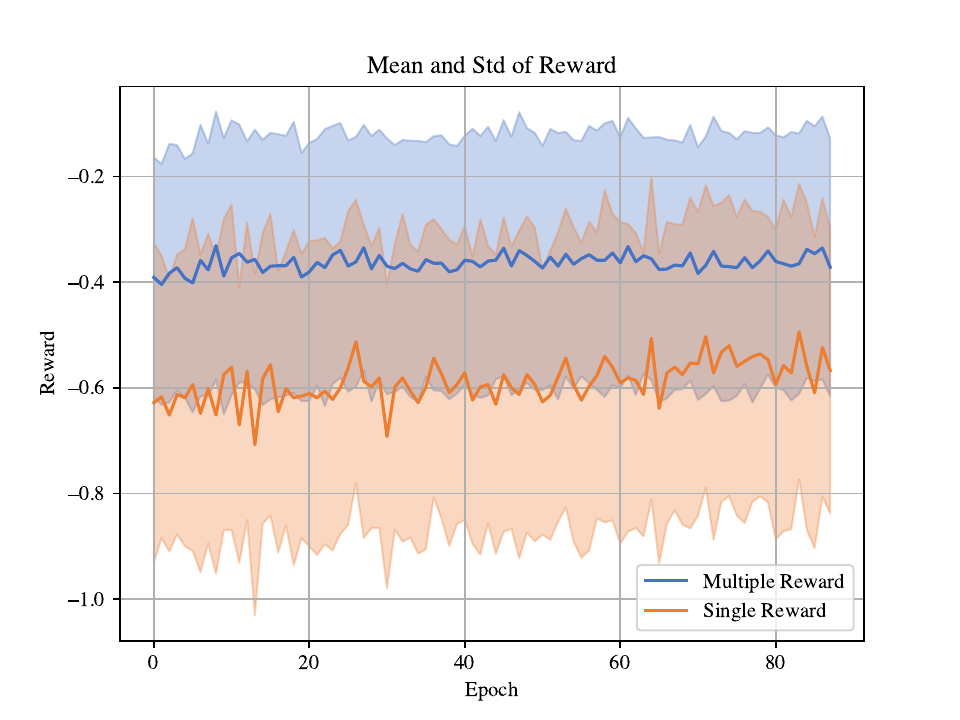}
    
    \caption{The mean and standard deviation of rewards per batch during the PPO optimization process for training the backbone model in Bone Soup (multiple rewards) and Rewarded Soup (single reward). The process of tuning the backbone model in Bone Soup is more stable compared to that in Rewarded Soup.}
    \label{fig:ppo}
\end{figure}

\begin{figure*}[h]
    \centering
    \begin{subfigure}{0.49\textwidth}
        \centering
        \includegraphics[width=\linewidth]{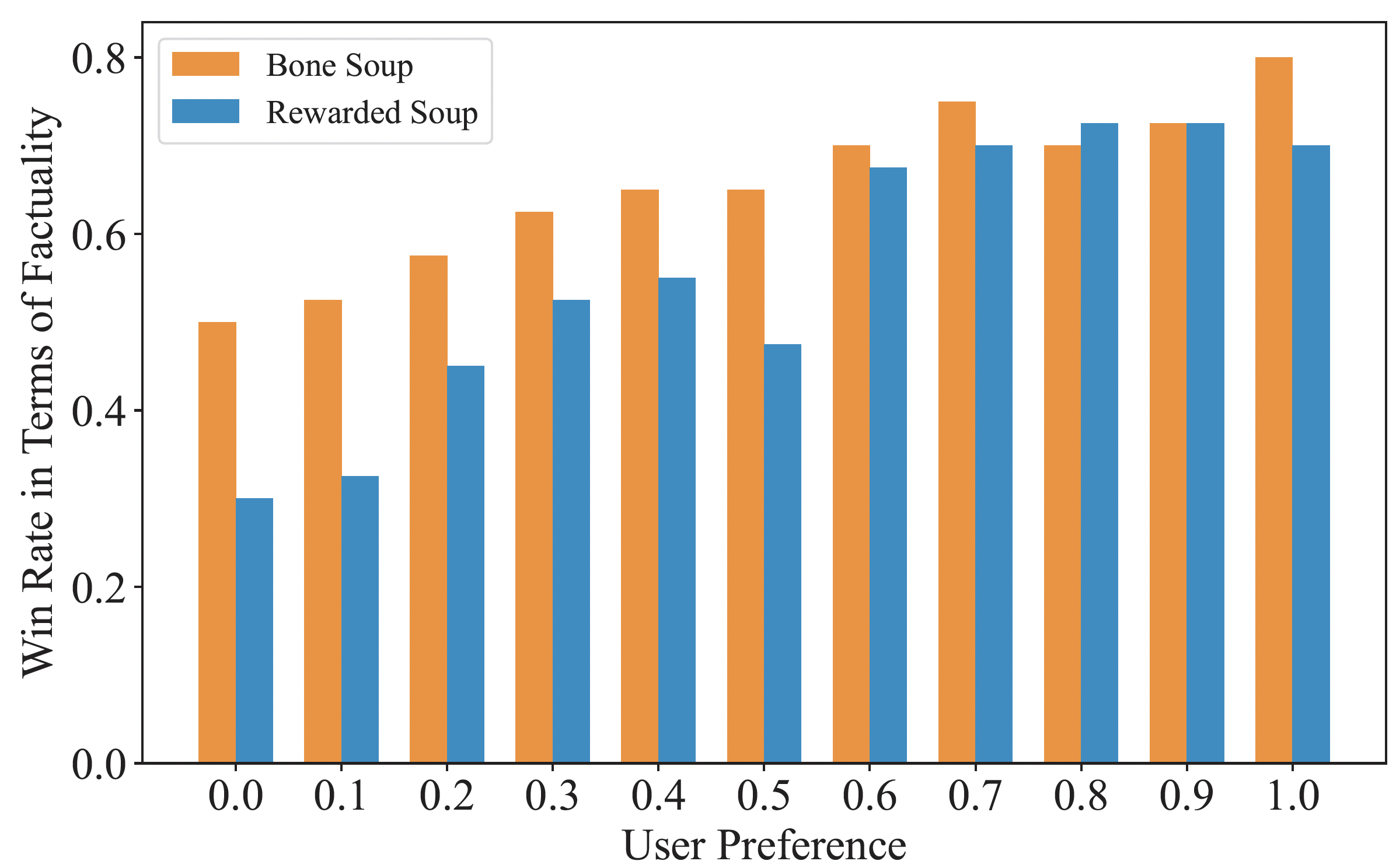}
        \caption{Factuality evaluation results}
        \label{fig:gpt4:fact}
    \end{subfigure}
    \hfill
    \begin{subfigure}{0.49\textwidth}
        \centering
        \includegraphics[width=\linewidth]{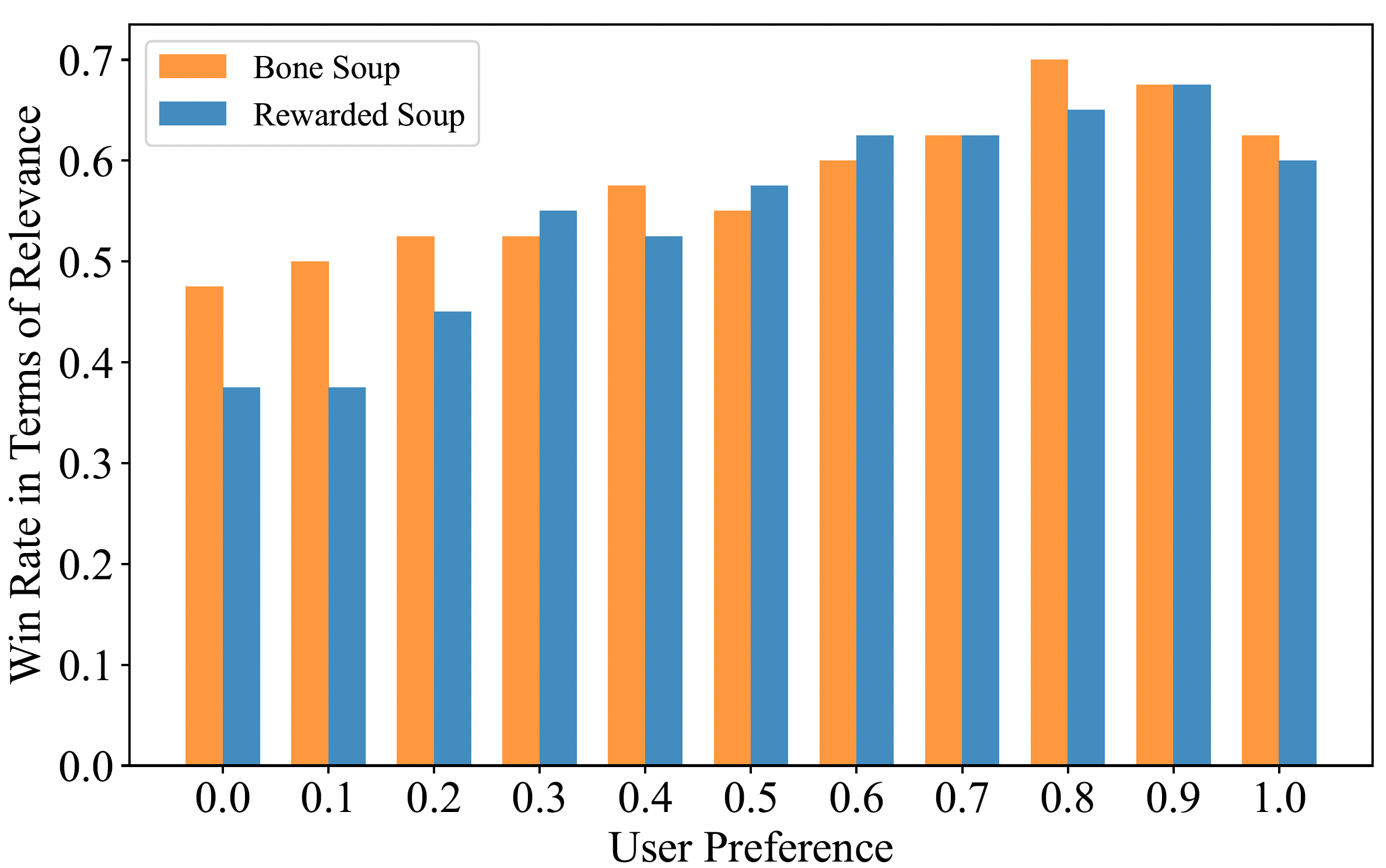}
        \caption{Relevance evaluation results}
        \label{fig:gpt4:rele}
    \end{subfigure}
    \caption{The real evaluation results by GPT-4 under different user preferences. Bone Soup achieves better performance compared with Rewarded Soup.}
\end{figure*}

\subsection{The Proof of Theorem 1}
\label{Appendix:proof}

\setcounter{theorem}{0}
\begin{theorem}
	Given quadratic reward functions with Hessians proportional to identity matrices:
	$$
	r_i(\theta) = r_i(\theta_i)-k_i\|\theta-\theta_i\|^2,i\in\{1,2\},
	$$where \( k_i \in \mathbb{R}_+ \) are distinct,and $\theta_i$ is the global maximum for reward $r_i$.
	Let the reward combination weight matrix be $B = \begin{pmatrix}
		\beta & 1-\beta \\
		1-\beta & \beta
	\end{pmatrix},\beta\in(\frac{1}{2},1)$, then the backbone rewards of the bone-soup approach can be denoted as $[h_1,h_2]^T = B[r_1,r_2]^T$.Let $\bm{\mu} = [\mu,1-\mu]^T$ be the user preference and the testing reward can written as $g_{\bm\mu}(\theta):=\bm\mu^T \begin{bmatrix}
	r_1 \\
	r_2
	\end{bmatrix} $.Denote the approximate solutions for the testing reward $g_{\mu}(\theta)$ of the soup-like approach and the bone-soup approach as $\bar\theta$ and $\bar\theta^{bone}$, respectively.Then,for any fixed $\beta \in (\frac{1}{2}, 1)$, when $\mu \in \left( \frac{1 - \sqrt{2\beta^2 - 2\beta + 1}}{2}, \frac{1 + \sqrt{2\beta^2 - 2\beta + 1}}{2} \right)$,
	\begin{align*}
	g_{\mu}(\bar\theta) < g_{\mu}(\bar\theta^{bone}) ,
	\end{align*}
	with interval length $\sqrt{2\beta^2 - 2\beta + 1} \geq \frac{\sqrt{2}}{2}$.
\end{theorem}

\begin{proof}
	The testing reward $g_{\mu} = \mu r_1+(1-\mu)r_2$ is quadratic thus has an unique global maximum $\theta^*$,that we find analytically:
    
        \resizebox{1.0\linewidth}{!}{
        \begin{minipage}{\linewidth}
	\begin{align*}
		\nabla_{\theta} g_{\mu}(\theta) = 0&\Rightarrow \mu k_1(\theta-\theta_1)+(1-\mu) k_2(\theta-\theta_2)=0 \\
		&\Rightarrow \theta^* = \frac{\mu k_1\theta_1+(1-\mu)k_2\theta_2}{\mu k_1+(1-\mu)k_2}
	\end{align*}
        \end{minipage}
        }
	The approximate solution $\bar\theta$ for the testing reward $g_\mu$ of the soup-like approach is formulated as,
	\begin{align*}
	 	\bar\theta = \mu \theta_1+(1-\mu)\theta_2
	\end{align*}
Consider the bone-soup approach,we have the backbone rewards and their corresponding global maximums as follows,
\resizebox{1.0\linewidth}{!}{
\begin{minipage}{\linewidth}
\begin{align*}
	h_1 = \beta r_1+(1-\beta)r_2,	\theta_1^{bone} =  \frac{\beta k_1\theta_1+(1-\beta)k_2\theta_2}{\beta k_1+(1-\beta)k_2} \\
	h_2 = (1-\beta) r_1 + \beta r_2,\theta_2^{bone} =  \frac{(1-\beta) k_1\theta_1+\beta k_2\theta_2}{(1-\beta) k_1+\beta k_2} 
\end{align*}
\end{minipage}
}

the merging coefficients can be calculated as $\bm\lambda = B^{-1}\bm\mu = [\lambda,1-\lambda]^T$,where $\lambda = \frac{\beta+\mu-1}{2\beta-1}$,then we have the soup-like approach's approximate solution $\bar\theta^{bone}$ for the testing reward $g_\mu$ as,   
\begin{align*}
	\bar\theta^{bone} = \lambda\theta_1^{bone} +(1-\lambda)\theta_2^{bone}
\end{align*}
We set the error function as $ E(\beta, \mu) = \|\bar\theta^{bone} - \theta^*\|^2$.Since the soup-like approach can be regarded as a special case of the bone-soup approach when $\beta=1$,we use $E(1,\mu)$ to denote the error of the soup-like approach.Under the current settings, the testing reward $g_{\mu}$ can be written as $g_{\mu} = c_1-c_2 \|\theta - \theta^*\|^2$,where $c_1$ and $c_2$ are constants, and $c_2 \in \mathbb{R}_+$.Therefore $g_{\mu}(\bar\theta) < g_{\mu}(\bar\theta^{bone}) \Leftrightarrow	E(\beta,\mu) < E(1,\mu)$.\\
\\The expressions for $E(\beta,\mu)$ and $ E(1,\mu)$ can be calculated as follows, \\

\resizebox{1.0\linewidth}{!}{
\begin{minipage}{\linewidth}
\begin{align*}
    &E(\beta,\mu) = \\
    & \|\lambda \frac{\beta k_1\theta_1+(1-\beta)k_2\theta_2}{\beta k_1+(1-\beta)k_2} + \\
    &(1-\lambda) \frac{(1-\beta) k_1\theta_1+\beta k_2\theta_2}{(1-\beta) k_1+\beta k_2} 
    - \frac{\mu k_1\theta_1+(1-\mu)k_2\theta_2}{\mu k_1+(1-\mu)k_2} \|^2 \\
    = & \left(\frac{k_1k_2(k_1-k_2)(\beta-\mu)(\beta+\mu-1)}{(\mu k_1+(1-\mu)k_2)(\beta k_1+(1-\beta)k_2)((1-\beta) k_1+\beta k_2)}\right)^2 \|\theta_1-\theta_2\|^2
\end{align*}
\end{minipage}
}
\begin{align*}
	E(1,\mu) = (\frac{(k_1-k_2)(1-\mu)\mu}{\mu k_1+(1-\mu)k_2})^2 \|\theta_1-\theta_2\|^2
\end{align*}

To compare$~E(\beta,\mu)$ and $~E(1,b)$, we have:

\resizebox{1.0\linewidth}{!}{
\begin{minipage}{\linewidth}
\begin{align*}
    &E(\beta,\mu) < E(1,\mu)\\
    &\Leftrightarrow (\frac{k_1k_2(\beta-\mu)(\beta+\mu-1)}{[\beta k_1+(1-\beta)k_2][(1-\beta) k_1+\beta k_2]})^2 < ((1-\mu)\mu)^2
\end{align*}
\end{minipage}
}

since
\begin{align*}
	&~[\beta k_1+(1-\beta)k_2][(1-\beta) k_1+\beta k_2] \\
    &= k_1k_2[2\beta^2-2\beta+1+\beta(1-\beta)(\frac{k_1}{k_2}+\frac{k_2}{k_1})]\\
    &\ge k_1k_2(2\beta^2-2\beta+1+2\beta(1-\beta)) = k_1k_2\\
\end{align*}
Therefore, we obtain: 

\resizebox{1.0\linewidth}{!}{
\begin{minipage}{\linewidth}
\begin{align*}
\left( \frac{k_1k_2(\beta - \mu)(\beta + \mu - 1)}{[\beta k_1 + (1 - \beta)k_2][(1 - \beta)k_1 + \beta k_2]} \right)^2 
&< \left( (\beta - \mu)(\beta + \mu - 1) \right)^2
\end{align*}
\end{minipage}
}

\begin{align*}
	&E(\beta,\mu) < E(1,\mu) \\
    &\Leftarrow (\mu(1-\mu))^2 - ((\beta-\mu)(\beta+\mu-1))^2 \\
	&= (\beta-\beta^2)(-2\mu^2+2\mu+\beta^2-\beta)> 0
\end{align*}
We can observe that,for any fixed $\beta \in (\frac{1}{2},1)$, the right-hand side of the equation holds, for all $ \mu \in\left( \frac{1 - \sqrt{2\beta^2 - 2\beta + 1}}{2}, \frac{1 + \sqrt{2\beta^2 - 2\beta + 1}}{2} \right)$, i.e. $E(\beta,\mu)<E(1,\mu)$ holds.Besides,we can find that the interval length   $\sqrt{2\beta^2 - 2\beta + 1} \geq \frac{\sqrt{2}}{2}$.Thus, the theorem is proved. 

\end{proof}
\subsection{Additional Experiments}

\subsubsection{RQ1: The Significance of Reconstructing Appropriate Backbone Models.} 
\label{sec:app:comb}

In Section~\ref{exa:bonesoup:qua}, we have already demonstrated the importance of reconstructing appropriate backbone models using a mathematical example. Here, we further illustrate the significance of basis reconstruction through some observations and empirical analysis.

\begin{center}
\begin{tcolorbox}[width=1.0\linewidth, boxrule=0pt, top=3pt, bottom=3pt, colback=gray!20, colframe=gray!20]
\textbf{Observation 1}: \textit{The specializing models trained with an individual reward for a single objective are not necessarily the optimal models under that specific reward function.}
\end{tcolorbox}
\label{observation1}
\end{center}

The effectiveness of model merging fundamentally depends on the quality and diversity of the backbone models. Relying solely on models trained for specific objectives may not yield optimal results, as such models might not fully explore or exploit the entire reward landscape. 

In Figure~\ref{fig:long}, we observe that the specializing models extrapolated by the two backbone models of Bone Soup consistently extend in two reward dimensions and outperform the specializing models tuned in RS, verifying the fact that models tuned with a specific reward may not always be the optimal ones for that reward and can be outperformed by models derived through various interpolation techniques.

\begin{center}
\begin{tcolorbox}[width=1.0\linewidth, boxrule=0pt, top=3pt, bottom=3pt, colback=gray!20, colframe=gray!20]
\textbf{Observation 2}: \textit{Incorporating additional rewards into the reward function enhances stability during model tuning.}
\end{tcolorbox}
\end{center}

In addition to the mathematical examples discussed in previous sections, we present some observations on incorporating additional rewards.
During the PPO tuning process, a single reward model may provide incorrect or unreasonable signals in specific situations due to its inherent limitations~\cite{casper2023open}, leading to significant fluctuations. By employing multiple reward models, these limitations can be mitigated through mutual complementation or correction, enhancing the stability of the tuning process. 

\begin{figure*}[h]
    \centering
    \begin{subfigure}{0.49\textwidth}
        \centering
        \includegraphics[width=\linewidth]{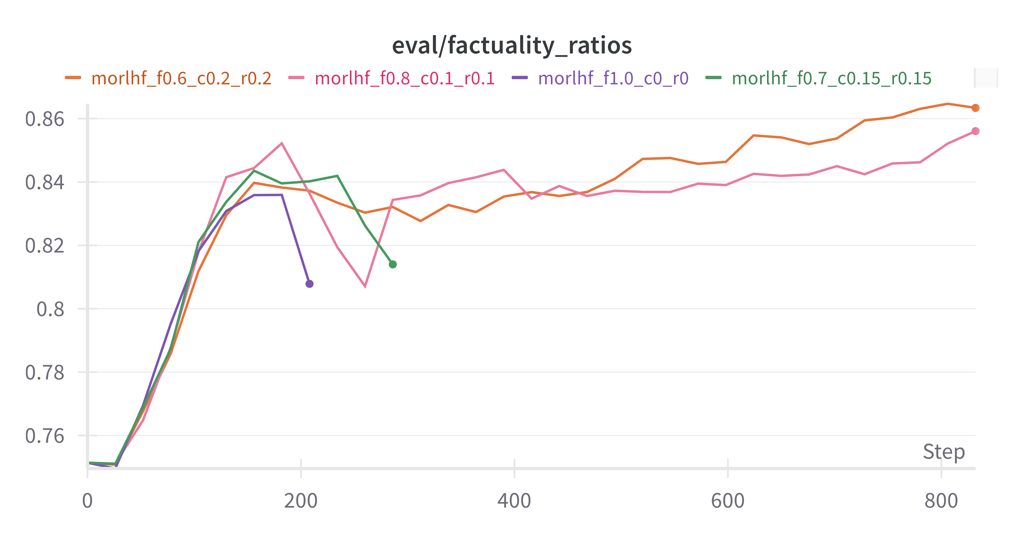} 
        \caption{Factuality rewards during PPO training}
        \label{fig:fact_f}
    \end{subfigure}
    \hfill 
    \begin{subfigure}{0.49\textwidth}
        \centering
        \includegraphics[width=\linewidth]{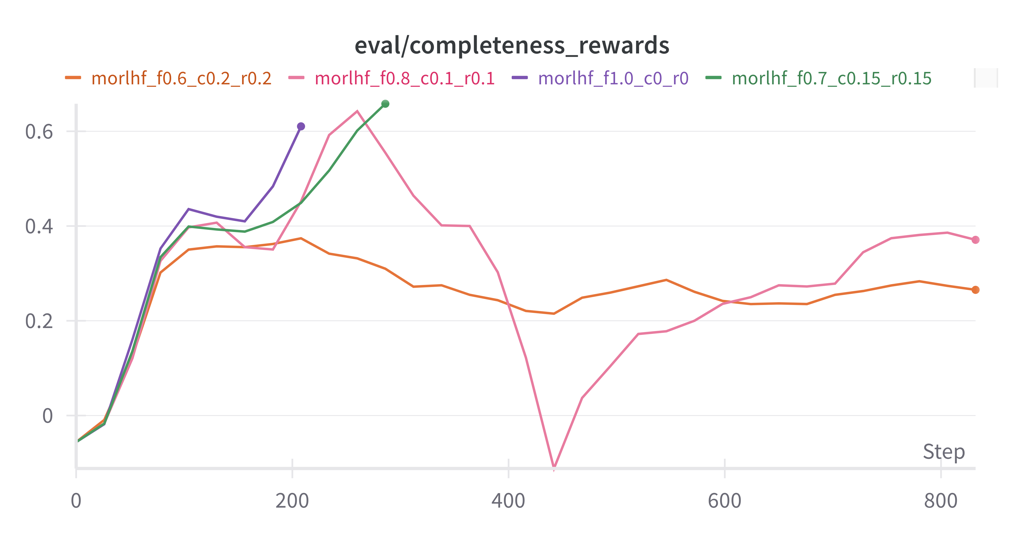} 
        \caption{Completeness rewards during PPO training}
        \label{fig:fact_c}
    \end{subfigure}
    \hfill
    \\
    \begin{subfigure}{0.49\textwidth}
        \centering
        \includegraphics[width=\linewidth]{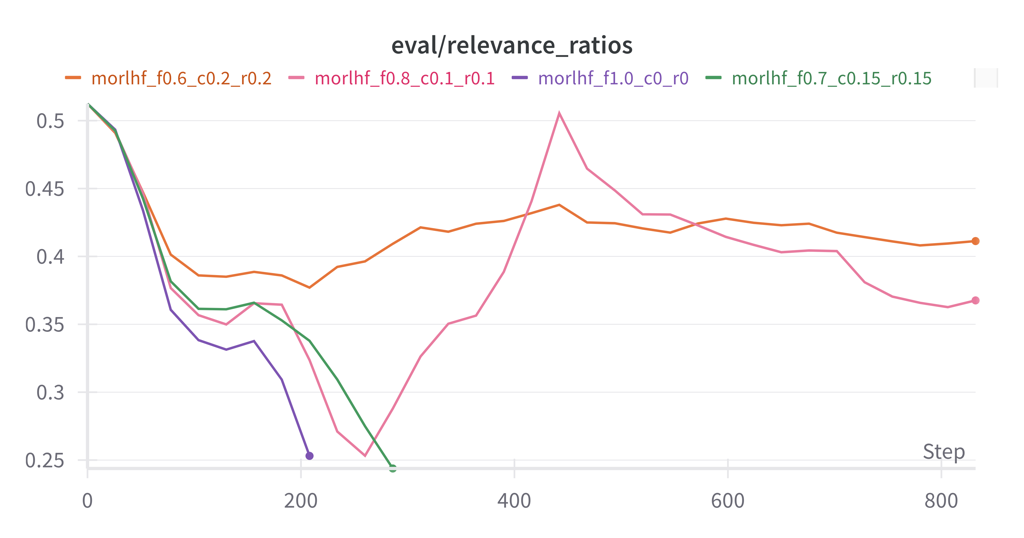} 
        \caption{Relevance rewards during PPO training}
        \label{fig:fact_r}
    \end{subfigure}
    \begin{subfigure}{0.49\textwidth}
        \centering
        \includegraphics[width=\linewidth]{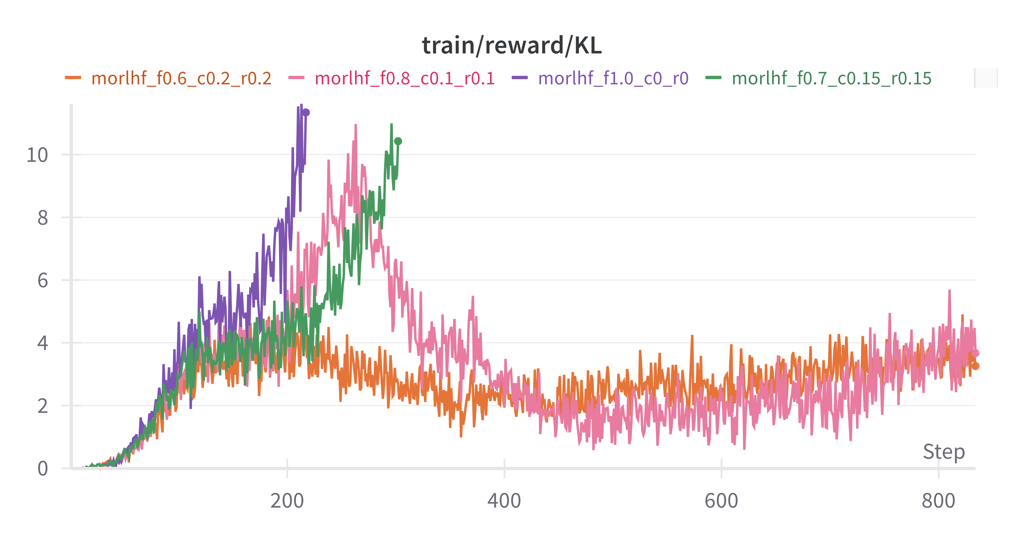} 
        \caption{KL divergence during PPO training}
        \label{fig:fact_kl}
    \end{subfigure}
    \\
    \begin{subfigure}{0.49\textwidth}
        \centering
        \includegraphics[width=\linewidth]{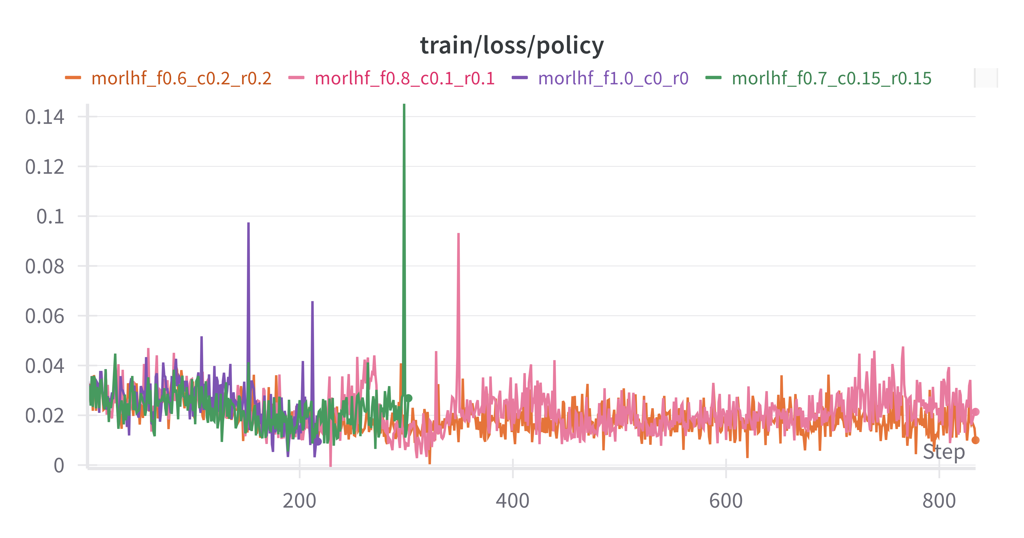} 
        \caption{Policy loss during PPO training}
        \label{fig:fact_ploss}
    \end{subfigure}
    \begin{subfigure}{0.49\textwidth}
        \centering
        \includegraphics[width=\linewidth]{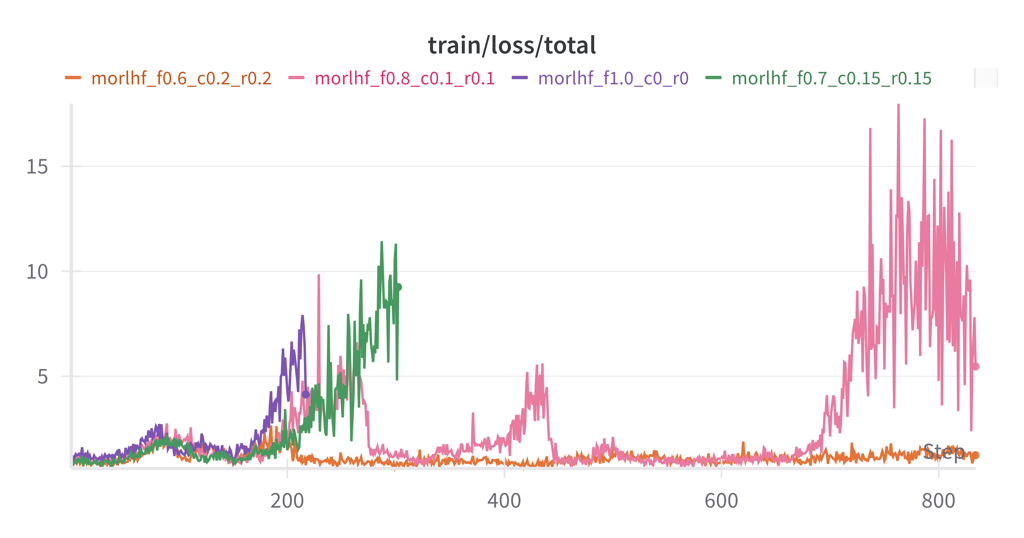} 
        \caption{Total loss during PPO training}
        \label{fig:fact_tloss}
    \end{subfigure}
    
    \caption{The factuality rewards, completeness rewards, relevance rewards, KL divergence, and policy loss during PPO training. All the subfigures are variations of different metrics of factuality-specialized model.}
    \label{fig:kl_fact_fcr}
\end{figure*}

We empirically validate that using multiple reward models can help smooth out the high variance problems introduced by a single model as shown in Figure~\ref{fig:ppo}. 
And Figure~\ref{fig:kl_fact_fcr} illustrates the training process of the factuality-specialized model using both combined rewards and a single reward during PPO training. The figure plots various metrics, including rewards, KL divergence, policy loss, and total loss. We observe that a spike in KL divergence during PPO training is indicative of model collapse, accompanied by a decline in the corresponding rewards, which suggests that early stopping is necessary. As shown in Figure~\ref{fig:fact_kl}, compared to the combined reward (especially at $\beta$ values of 0.6 and 0.8), the single reward leads to a more rapid increase in KL divergence, reaching the threshold sooner and triggering premature termination of training. Additionally, despite the combined reward placing relatively less emphasis on the factuality dimension—resulting in a lower focus on factuality during PPO training—it nonetheless delivers superior factuality performance. At the same time, the rewards across other dimensions are also enhanced compared to the rewarded soup approach, as evidenced by Figures \ref{fig:fact_c} and \ref{fig:fact_r}.

\begin{figure*}[h]
    \centering
    \begin{subfigure}{0.49\textwidth}
        \centering
        \includegraphics[width=\linewidth]{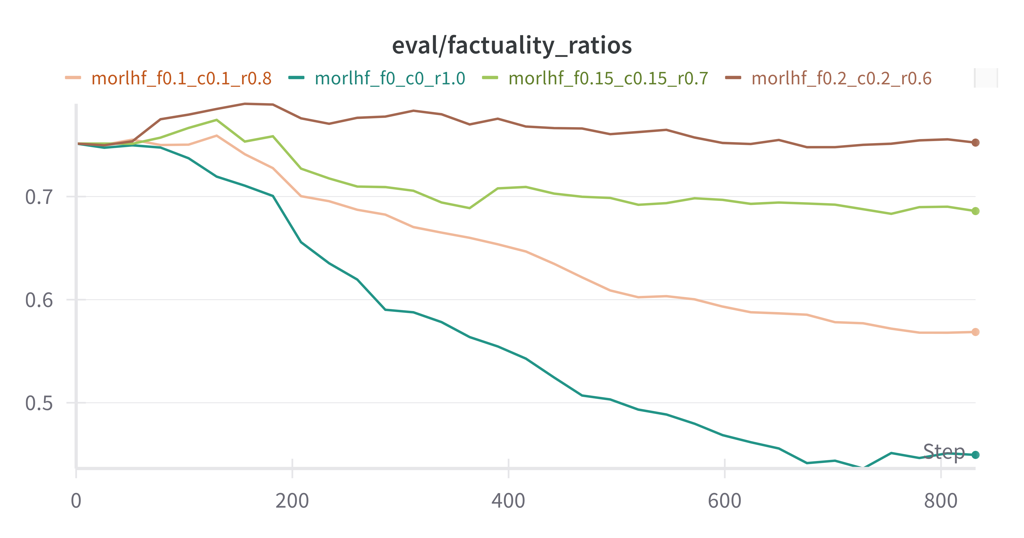} 
        \caption{Factuality rewards during PPO training}
        \label{fig:rele_f}
    \end{subfigure}
    \hfill 
    \begin{subfigure}{0.49\textwidth}
        \centering
        \includegraphics[width=\linewidth]{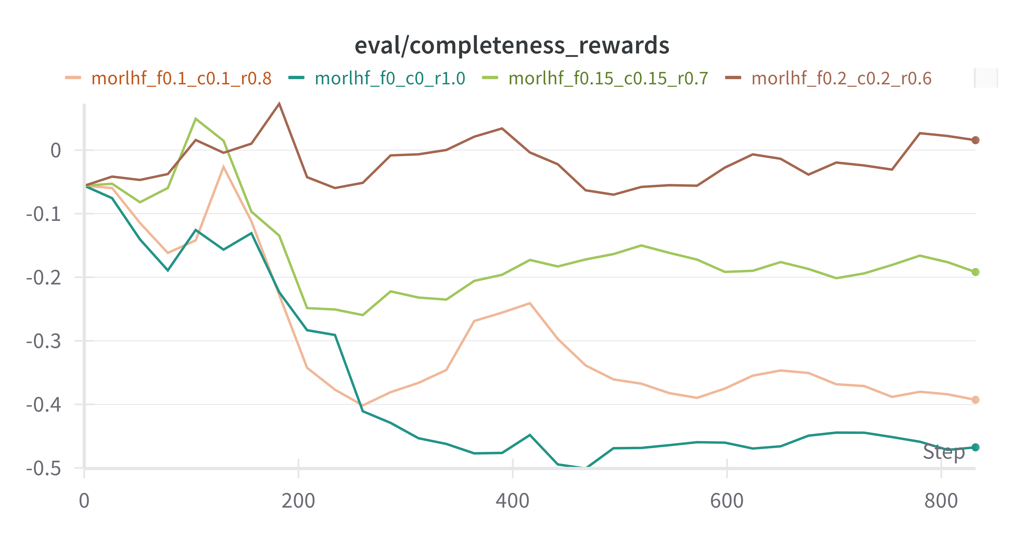} 
        \caption{Completeness rewards during PPO training}
        \label{fig:rele_c}
    \end{subfigure}
    \hfill
    \\
    \begin{subfigure}{0.49\textwidth}
        \centering
        \includegraphics[width=\linewidth]{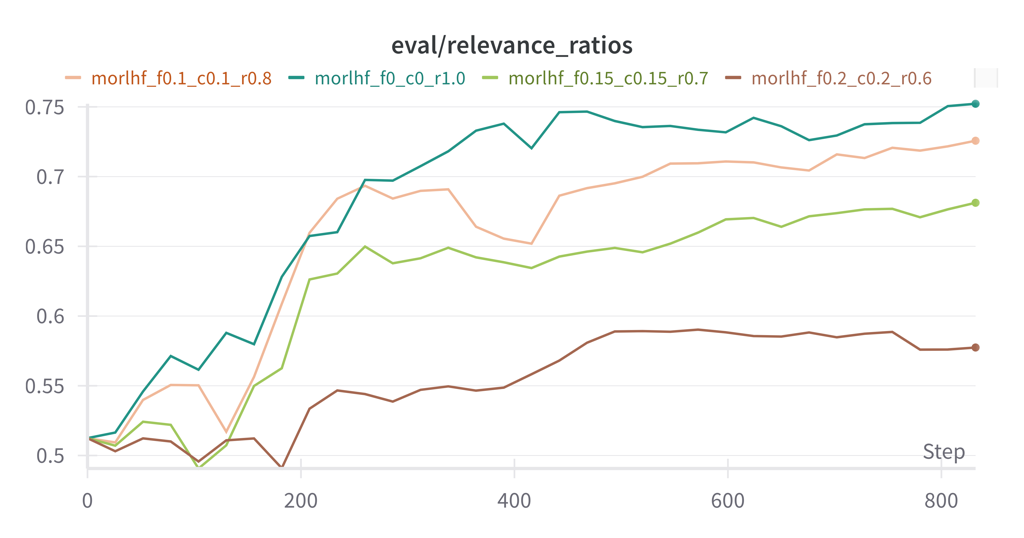} 
        \caption{Relevance rewards during PPO training}
        \label{fig:rele_r}
    \end{subfigure}
    \begin{subfigure}{0.49\textwidth}
        \centering
        \includegraphics[width=\linewidth]{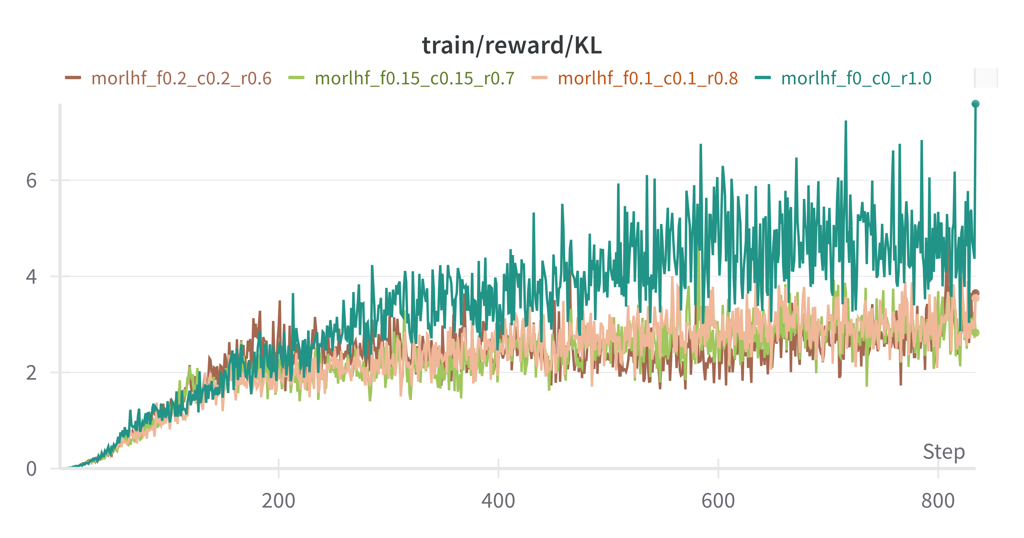} 
        \caption{KL divergence during PPO training}
        \label{fig:rele_kl}
    \end{subfigure}
    \\
    \begin{subfigure}{0.49\textwidth} 
        \centering
        \includegraphics[width=\linewidth]{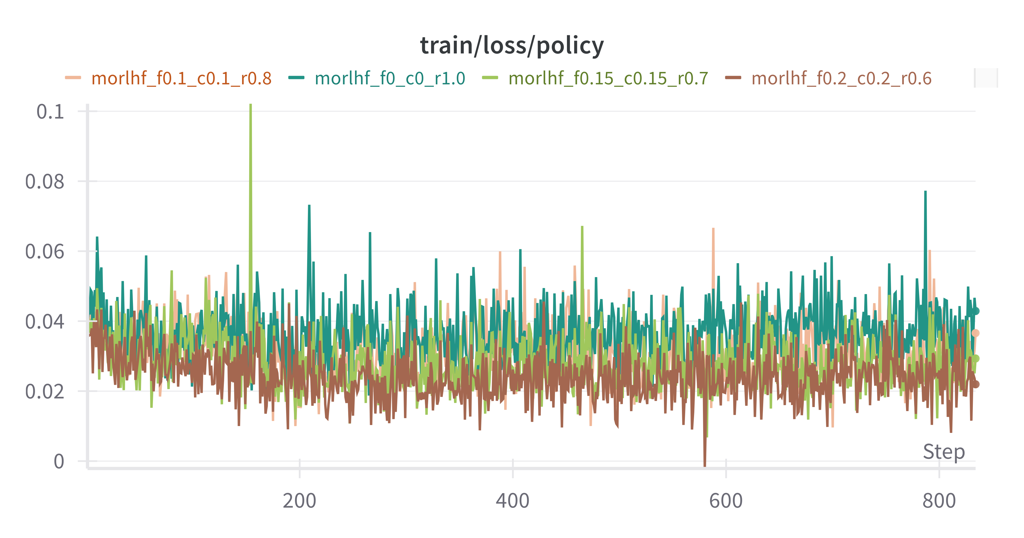} 
        \caption{Policy loss during PPO training}
        \label{fig:rele_ploss}
    \end{subfigure}
    \begin{subfigure}{0.49\textwidth}
        \centering
        \includegraphics[width=\linewidth]{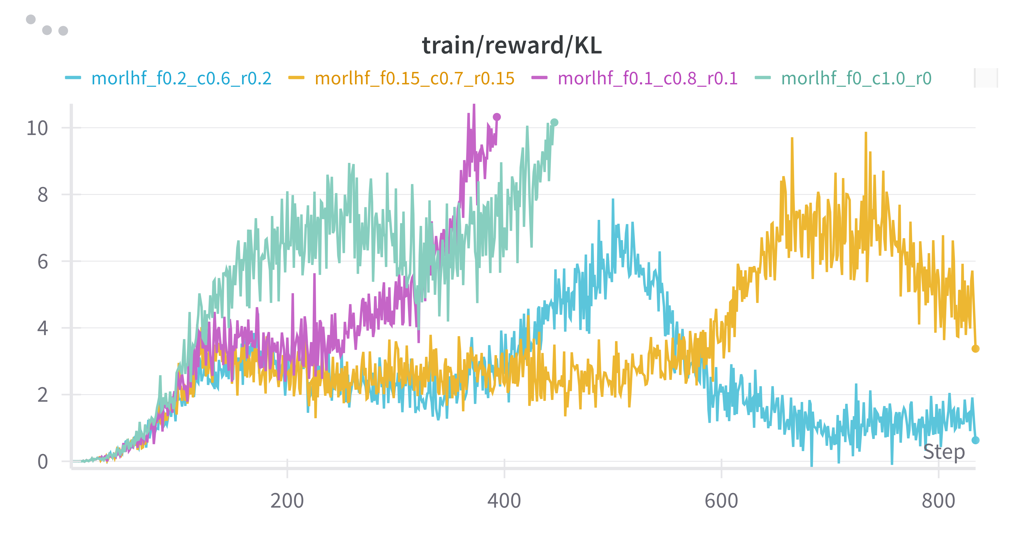} 
        \caption{KL divergence during PPO training of completeness-specialized model}
        \label{fig:comp_kl}
    \end{subfigure}

    \caption{The factuality rewards, completeness rewards, relevance rewards, KL divergence, and policy loss during PPO training. Subfigure (a), (b), (c), (d), and (e) depict the variations of different metrics during the PPO training process of the relevance-specialized model, while (f) shows the KL divergence changes during the PPO training process of the completeness-specialized model.}
    \label{fig:kl_rele_comp}
\end{figure*}

Moreover, Figures~\ref{fig:fact_kl},~\ref{fig:fact_ploss}, and ~\ref{fig:fact_tloss} clearly show that the training process with the combined reward is more stable. We attribute this stability to the integration of multiple rewards, which helps to counteract the incorrect and unstable signals that a single reward model might introduce. In essence, by blending multiple rewards, the potential instabilities during training are effectively mitigated.
And the similar results of training the completeness-specialized and relevance-specialized models are shown in Figure~\ref{fig:kl_rele_comp}. In Figure~\ref{fig:rele_kl} we can also observe a similar spike and higher KL compared with combined rewards. In Figure~\ref{fig:rele_ploss}, we also found a higher policy loss potentially representing the difficulty of convergence during training.

\begin{table*}[htbp]
    \centering
    \caption{The results of $\beta$ selection for three different trade-offs in the Long-form QA task. As shown in the table, $\beta$ = 0.6 achieves the best hypervolume(for simplicity of the method, we assume that hypervolume could represent the overall performance of the front.) on the small-scale validation set, and thus, $\beta$ = 0.6 is ultimately chosen for the three different trade-offs.}
    \resizebox{\textwidth}{!}{
    \begin{tabular}{ccccccccccccccccccc}
        \toprule
         & \multicolumn{3}{c}{\textbf{Hypervolume \(\uparrow\)}} & \multicolumn{3}{c}{\textbf{Inner Product \(\uparrow\)}} & \multicolumn{3}{c}{\textbf{Controllability \(\uparrow\)}} & \multicolumn{3}{c}{\textbf{Length of Front \(\uparrow\)}} & \multicolumn{3}{c}{\textbf{Sparsity \(\downarrow\)}}  & \multicolumn{3}{c}{\textbf{Spacing \(\downarrow\)}} \\
        \multicolumn{1}{c}{\textbf{Method}} & FR & CR & FC & FR & CR & FC & FR & CR & FC & FR & CR & FC & FR & CR & FC & FR & CR & FC  \\
        \midrule
        Bone Soup (6) select & \textbf{0.159} & \textbf{0.762} & \textbf{0.304} & \textbf{0.799} & \textbf{0.507} & \textbf{0.768} & 1.000 & 1.000 & 0.982 & 11 & 11 & 10 & 0.015 & 0.106 & 0.038 & 0.006 & 0.022 & 0.006 \\
        Bone Soup (7) select & 0.150 & 0.735 & 0.281 & 0.769 & 0.493 & 0.754 & 1.000 & 1.000 & 1.000 & 11 & 11 & 11 & 0.020 & 0.090 & 0.032 & 0.004 & 0.015 & 0.003 \\
        Bone Soup (8) select & 0.139 & 0.729 & 0.296 & 0.742 & 0.471 & 0.766 & 0.982 & 1.000 & 1.000 & 10 & 11 & 11 & 0.017 & 0.093 & 0.035 & 0.007 & 0.013 & 0.006 \\
        \bottomrule
    \end{tabular}
    }
    \label{tab:beta-select}
\end{table*}

\subsubsection{RQ2: Can the Small-Scale Selection Yield a $\beta$  Consistent with or Near the Optimal $\beta$?}
\label{sec:app:beta2}

In this section, we discuss the approach to selecting the $\beta$ parameter when constructing the matrix $\bm B$. Table~\ref{tab:beta-select} presents BoneSoup’s performance under various $\beta$ values for different trade-offs. We expect that a small-scale training run can effectively approximate the optimal beta found through full-scale training. Notably, Table~\ref{tab:beta-select} reveals that for all three trade-offs, a beta value of 0.6 consistently yields the best performance. Correspondingly, Table~\ref{tab:2} shows that, with full-scale training, the optimal performance is achieved at $\beta$ = 0.6 for the FR (factuality vs. relevance) and FC (factuality vs. completeness) trade-offs, while for the CR (completeness vs. relevance) trade-off, $\beta$ = 0.7 is optimal, with $\beta$ = 0.6 coming in as a close second. This strong alignment between small-scale and full-scale training results underscores the soundness and robustness of our $\beta$ selection strategy, enabling us to efficiently acquire a near-optimal $\beta$ that approximates the best possible performance.

\subsubsection{RQ3: Can Our Bone Soup Matrix Construction Method Effectively Identify High-Quality Matrices in a Vast Solution Space?}
\label{appendix:rq3}

To demonstrate the robustness and superiority of the Bone Soup matrix construction method, this section presents a performance comparison between the Bone Soup method and random matrix construction methods. Due to computational constraints, it is infeasible to exhaustively enumerate all potential backbone matrices $\bm B$. Therefore, we randomly constructed eight relatively representative matrices, as shown in Table~\ref{tab:random}.

The results are shown in Table~\ref{tab:avg-rank}. From the table, we can observe that under three different trade-offs, all Bone Soup variants utilizing multiple rewards outperform the naive Rewarded Soup method, which relies on a single reward. This advantage stems from the fact that multi-reward RL facilitates the construction of a superior backbone model, demonstrating that using combined rewards can better approximate the optimal solution.

Moreover, employing our proposed rule-based construction method in combination with the hypervolume-selection approach further surpasses the randomly selected reward matrices. Bone Soup achieves the best performance, while Bone Soup ($\beta=0.7$), which solely adopts the rule-based construction without adaptation, attains the second-best performance, still outperforming the randomly constructed method.

\begin{table}[h]
    \centering
    \caption{The Specific Representation of the Matrix $\bm B_i$}
    \resizebox{230pt}{!}{
    \begin{tabular}{cc}
        \toprule
        \multicolumn{2}{c}{\textbf{The randomly selected matrices $\bm B_i$}} \\
        \midrule
        $\bm B_{1} = \begin{pmatrix}
            0.7  & 0.2  & 0.15 \\
            0.15 & 0.6  & 0.15 \\
            0.15 & 0.2  & 0.7
        \end{pmatrix}$ & 
        $\bm B_{2} = \begin{pmatrix}
            0.7  & 0.1  & 0.1 \\
            0.15 & 0.8  & 0.1 \\
            0.15 & 0.1  & 0.8
        \end{pmatrix}$ \\
        
        $\bm B_{3} = \begin{pmatrix}
            0.7  & 0.15 & 0.1 \\
            0.15 & 0.7  & 0.1 \\
            0.15 & 0.15 & 0.8
        \end{pmatrix}$ & 
        $\bm B_{4} = \begin{pmatrix}
            0.7  & 0.15 & 0.2 \\
            0.15 & 0.7  & 0.2 \\
            0.15 & 0.15 & 0.6
        \end{pmatrix}$ \\
        
        $\bm B_{5} = \begin{pmatrix}
            0.8   & 0.15  & 0.2 \\
            0.1   & 0.7   & 0.2 \\
            0.1   & 0.15  & 0.6
        \end{pmatrix}$ & 
        $\bm B_{6} = \begin{pmatrix}
            0.7  & 0.2  & 0.1 \\
            0.15 & 0.6  & 0.1 \\
            0.15 & 0.2  & 0.8
        \end{pmatrix}$ \\
        
        $\bm B_{7} = \begin{pmatrix}
            0.7  & 0.2  & 0.2 \\
            0.15 & 0.6  & 0.2 \\
            0.15 & 0.2  & 0.6
        \end{pmatrix}$ & 
        $\bm B_{8} = \begin{pmatrix}
            0.8  & 0.2  & 0.2 \\
            0.1  & 0.6  & 0.2 \\
            0.1  & 0.2  & 0.6
        \end{pmatrix}$ \\
        \bottomrule
    \end{tabular}
    }
    \label{tab:random}
\end{table}

\begin{table}[htbp]
\centering
\caption{The comparison between Bone Soup and random matrix constction methods in 3 different trade-offs}
\label{tab:avg-rank}
\resizebox{230.pt}{!}{
\begin{tabular}{lrrrr}
\toprule
Method & RC\_Rank & FR\_Rank & FC\_Rank & Avg\_Rank \\
\midrule
Rewarded Soups            & 12.0 & 12.0 & 12.0 & 12.00 \\
\midrule
Bone Soup                 & 7.0  & 2.0  & 2.0  & \textbf{3.67} \\
Bone Soup ($\beta$=0.7)    & 1.0  & 3.0  & 8.0  & \uline{4.00} \\
Bone Soup ($\beta$=0.8)    & 11.0 & 8.0  & 1.0  & 6.67 \\
Bone Soup $\bm B_1$           & 3.0  & 1.0  & 10.0 & 4.67 \\
Bone Soup $\bm B_2$           & 9.0  & 11.0 & 4.0  & 8.00 \\
Bone Soup $\bm B_3$        & 8.0  & 9.0  & 6.0  & 7.67 \\
Bone Soup $\bm B_4$        & 2.0  & 6.0  & 9.0  & 5.67 \\
Bone Soup $\bm B_5$        & 4.0  & 5.0  & 3.0  & 4.00 \\
Bone Soup $\bm B_6$       & 10.0 & 10.0 & 7.0  & 9.00 \\
Bone Soup $\bm B_7$        & 4.0  & 6.0  & 11.0 & 7.00 \\
Bone Soup $\bm B_8$        & 6.0  & 4.0  & 5.0  & 5.00 \\
\bottomrule
\end{tabular}
}
\end{table}

\subsubsection{RQ4: Robustness Analysis of Bone Soup} 
\begin{figure*}[h]
    \centering
    \begin{subfigure}{0.32\textwidth}
        \centering
        \includegraphics[width=\linewidth]{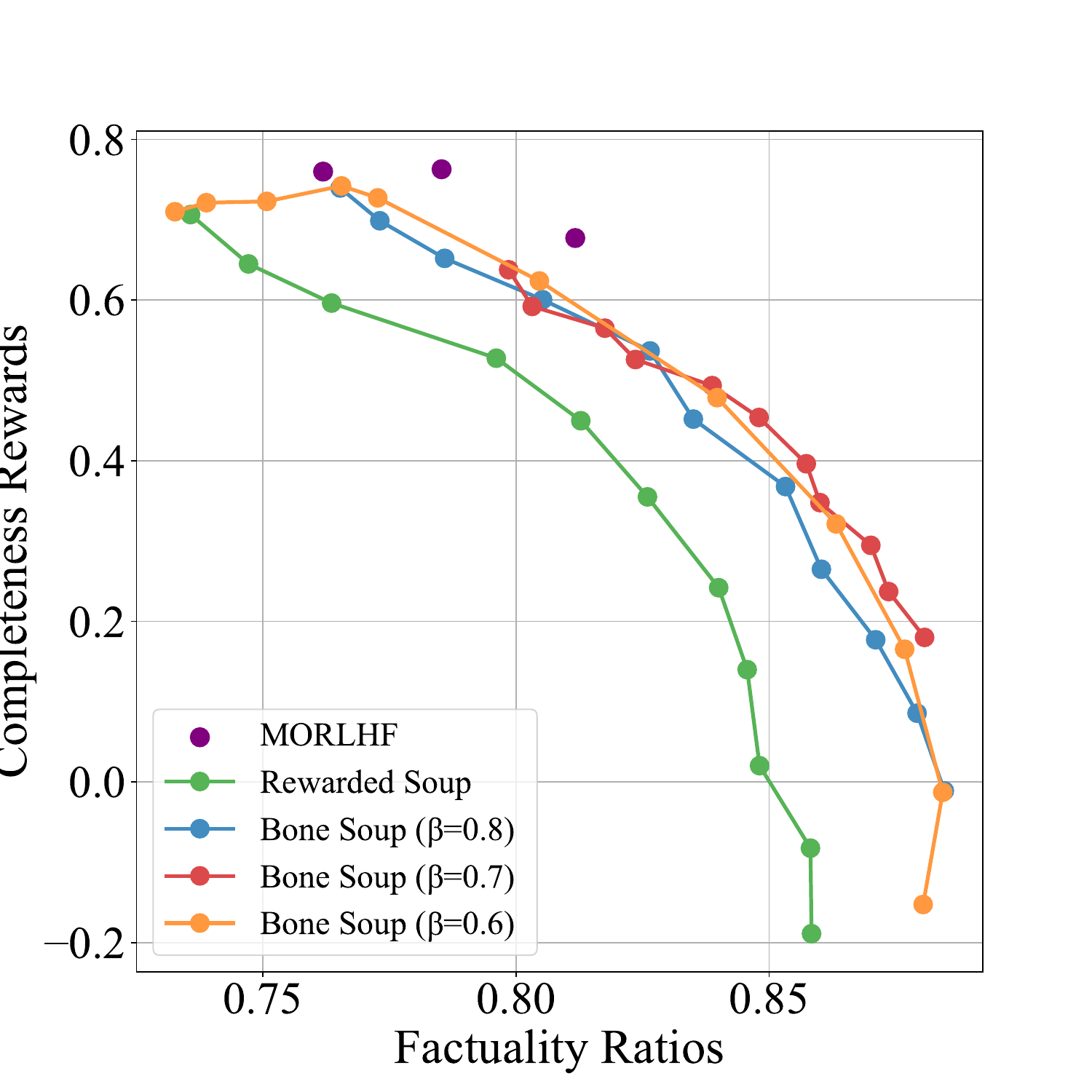} 
        \caption{Seed2: factuality vs completeness}
        \label{fig:seed2_a}
    \end{subfigure}
    \hfill 
    \begin{subfigure}{0.32\textwidth}
        \centering
        \includegraphics[width=\linewidth]{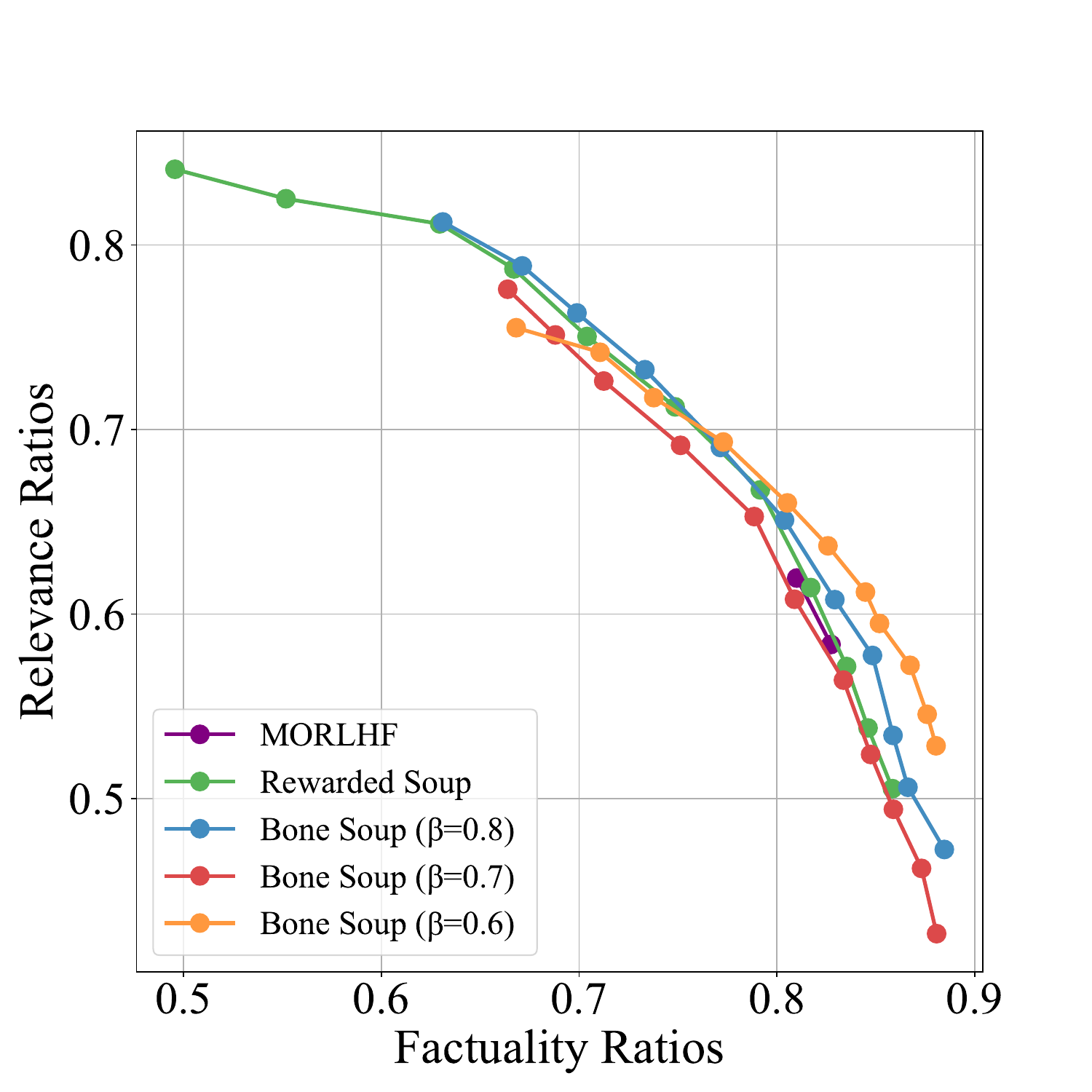} 
        \caption{Seed2: factuality vs relevance}
        \label{fig:seed2_b}
    \end{subfigure}
    \hfill
    \begin{subfigure}{0.32\textwidth}
        \centering
        \includegraphics[width=\linewidth]{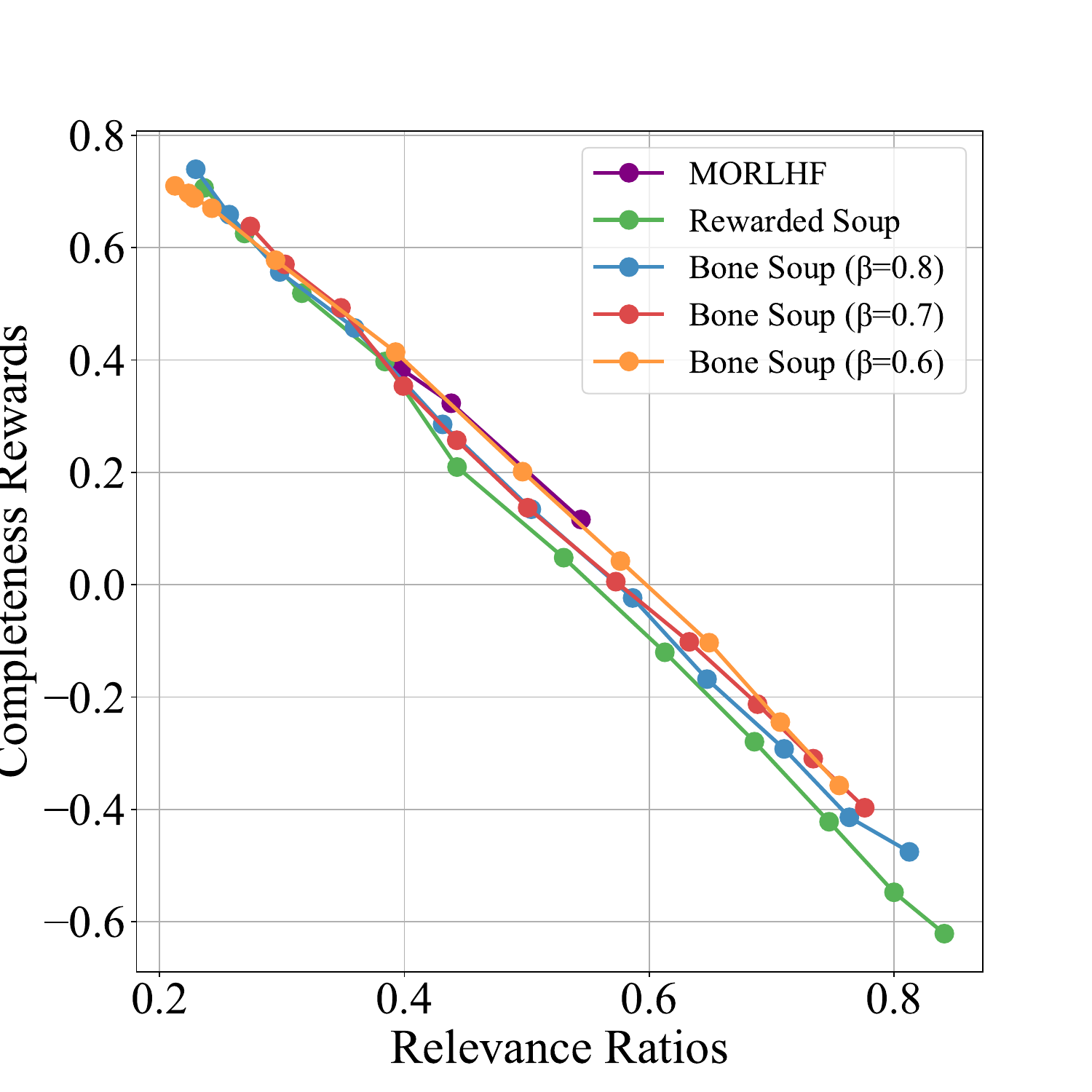} 
        \caption{Seed2: relevance vs completeness}
        \label{fig:seed2_c}
    \end{subfigure}
    \\
    \begin{subfigure}{0.32\textwidth}
        \centering
        \includegraphics[width=\linewidth]{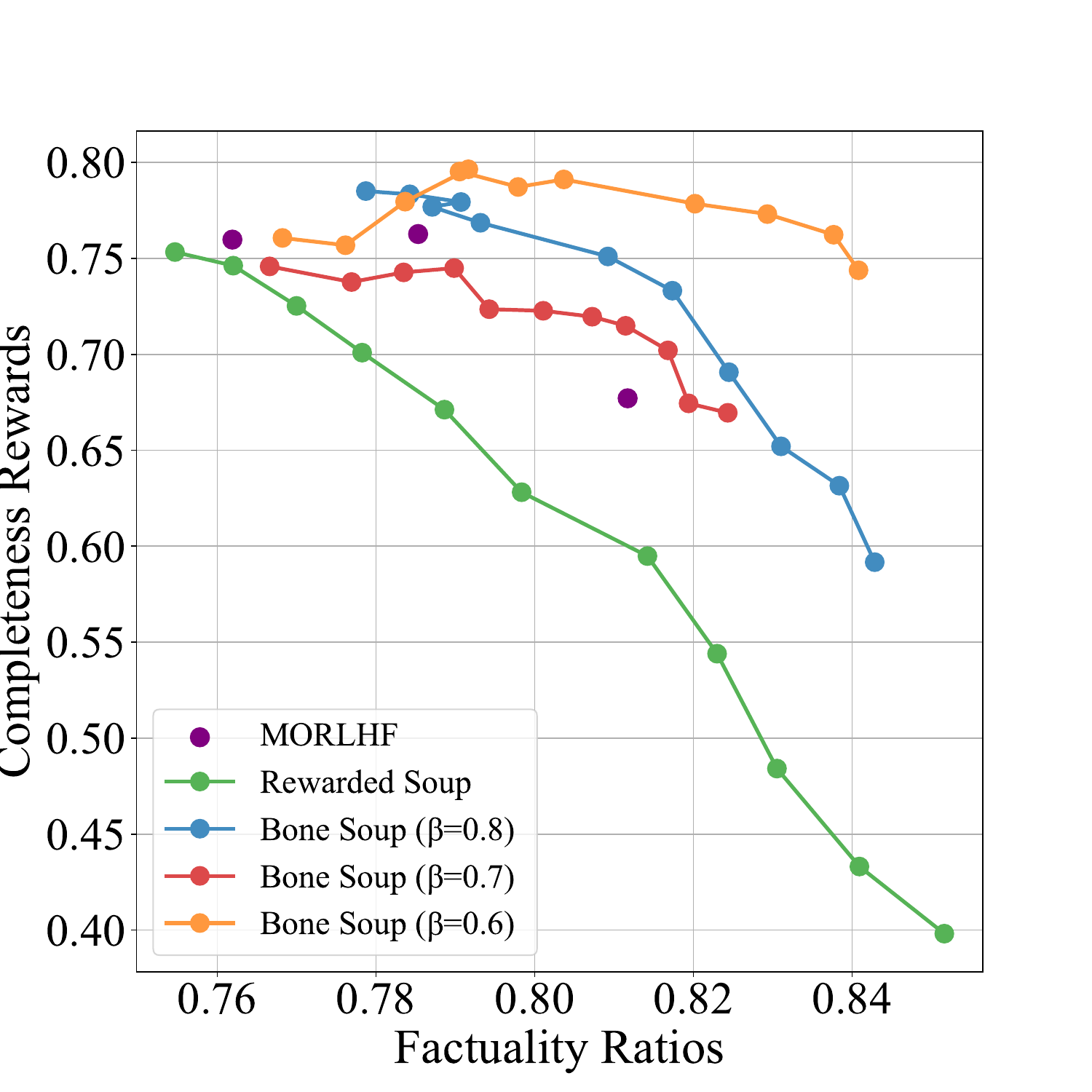} 
        \caption{Seed3: factuality vs completeness}
        \label{fig:seed3_a}
    \end{subfigure}
    \hfill 
    \begin{subfigure}{0.32\textwidth}
        \centering
        \includegraphics[width=\linewidth]{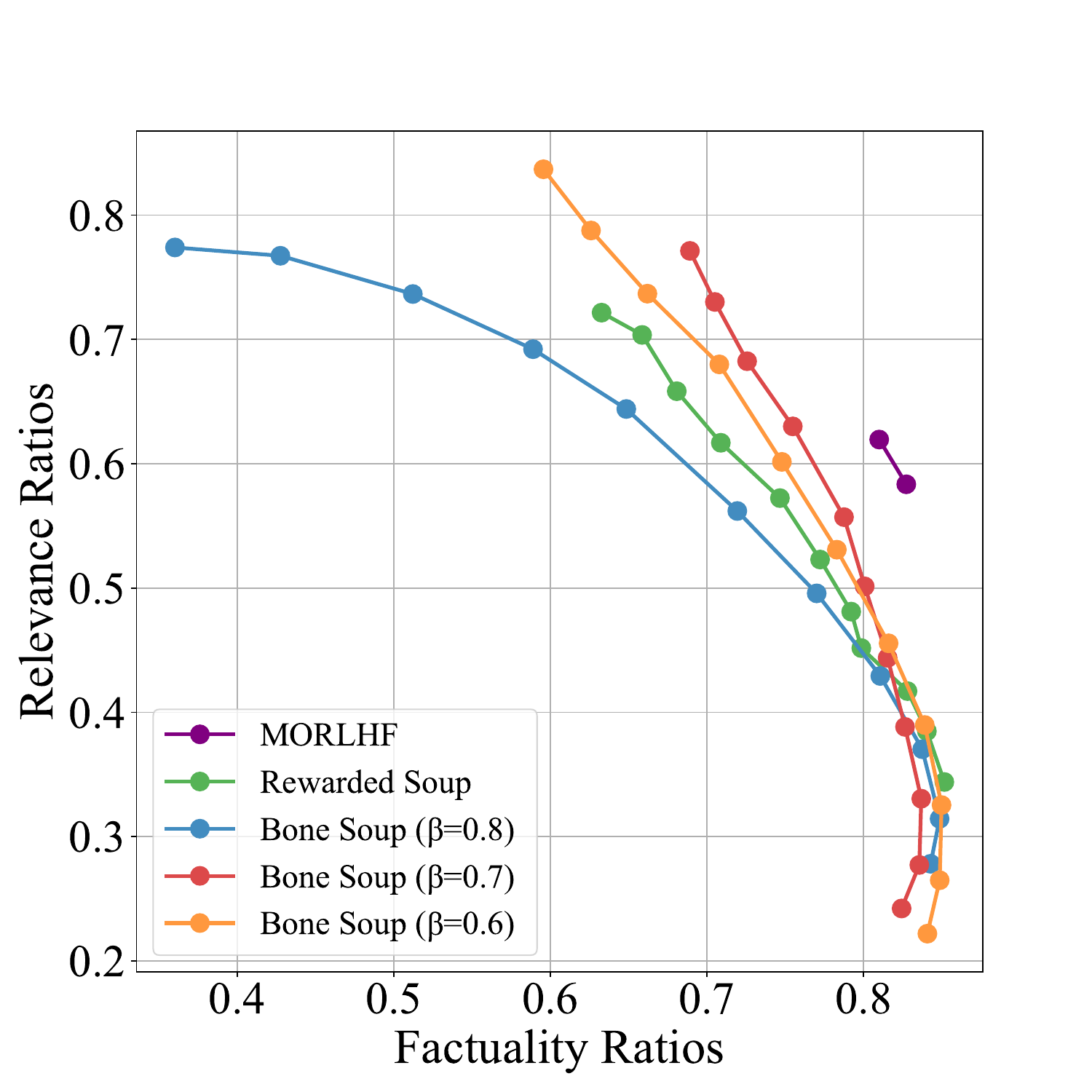} 
        \caption{Seed3: factuality vs relevance}
        \label{fig:seed3_b}
    \end{subfigure}
    \hfill
    \begin{subfigure}{0.32\textwidth}
        \centering
        \includegraphics[width=\linewidth]{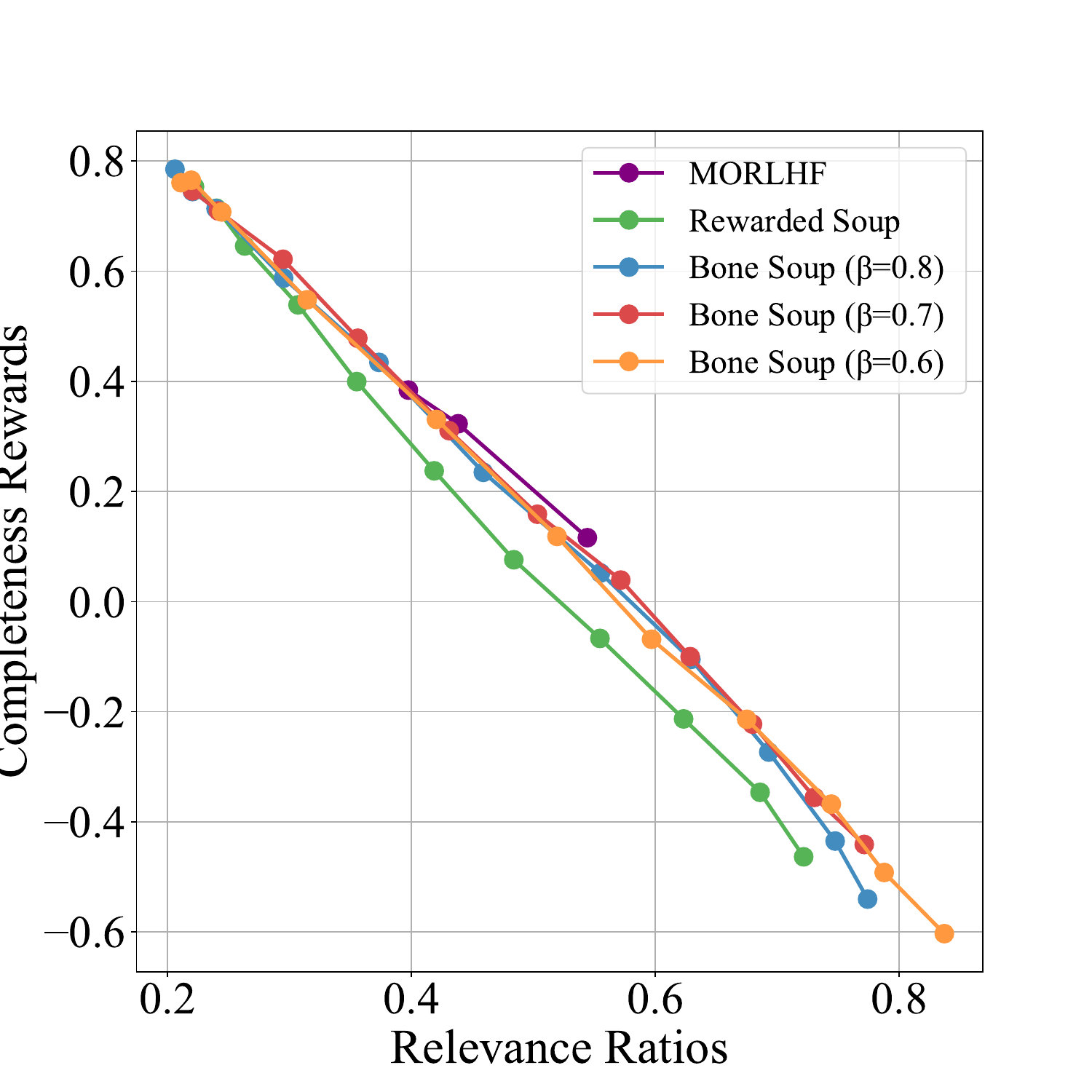} 
        \caption{Seed3: relevance vs completeness}
        \label{fig:seed3_c}
    \end{subfigure}
    
    \caption{Ablation study on different random seeds for the Long-Form QA task, evaluating ``factuality vs. relevance'', ``factuality vs. completeness'', ``relevance vs. completeness''. We connect the points in the figure according to the order of the preference weight partial order relation. We varied the random seeds and observed that Bonesoup demonstrates strong robustness and consistently outperforms other baselines.}
    \label{fig:seed3}
    
\end{figure*}

\begin{figure*}[h]
    \centering
    \begin{subfigure}{0.48\textwidth}
        \centering
        \includegraphics[width=\linewidth]{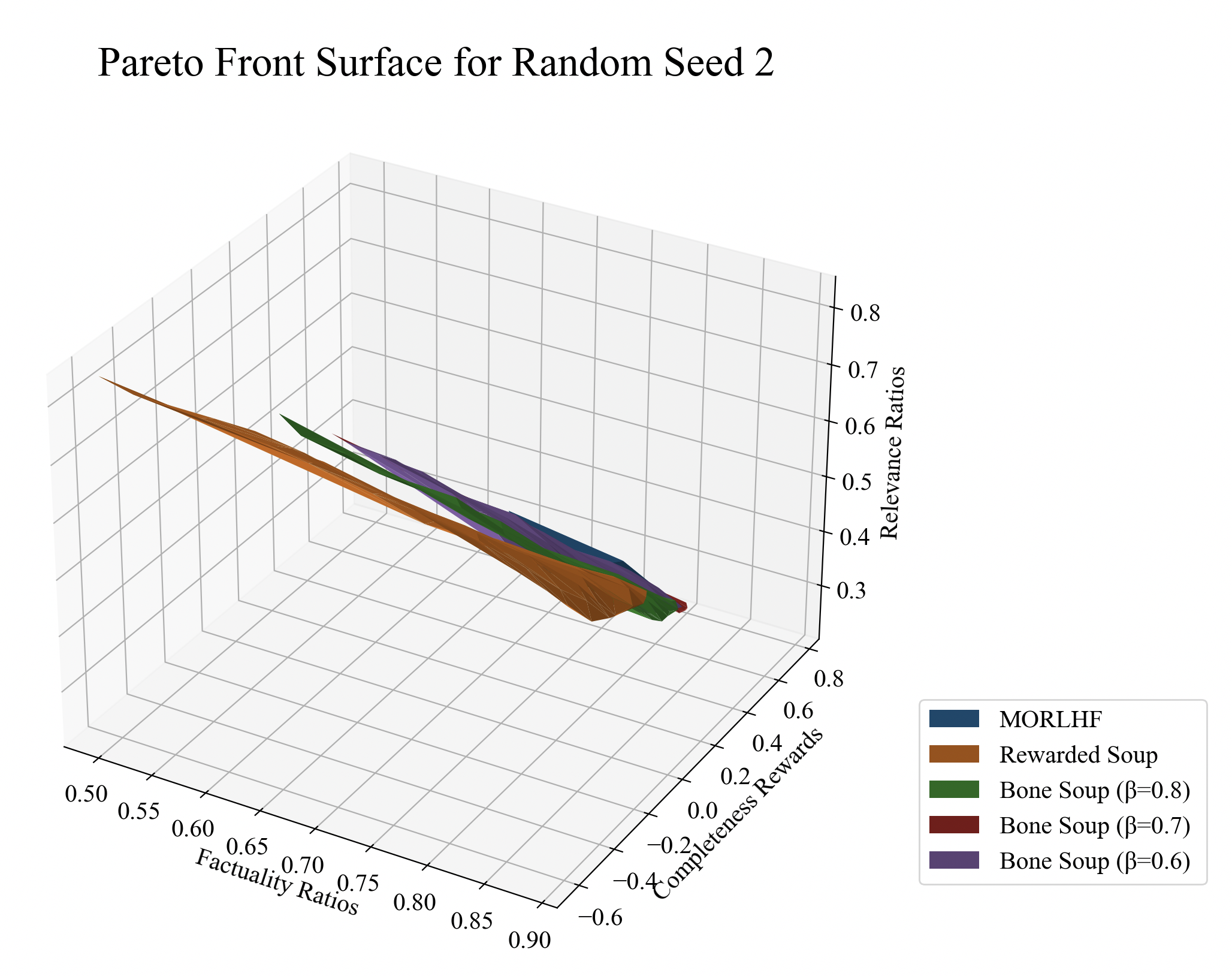} 
        \caption{factuality vs relevance vs completeness (seed 2)}
        \label{fig:seed_overall_a}
    \end{subfigure}
    \hfill 
    \begin{subfigure}{0.48\textwidth}
        \centering
        \includegraphics[width=\linewidth]{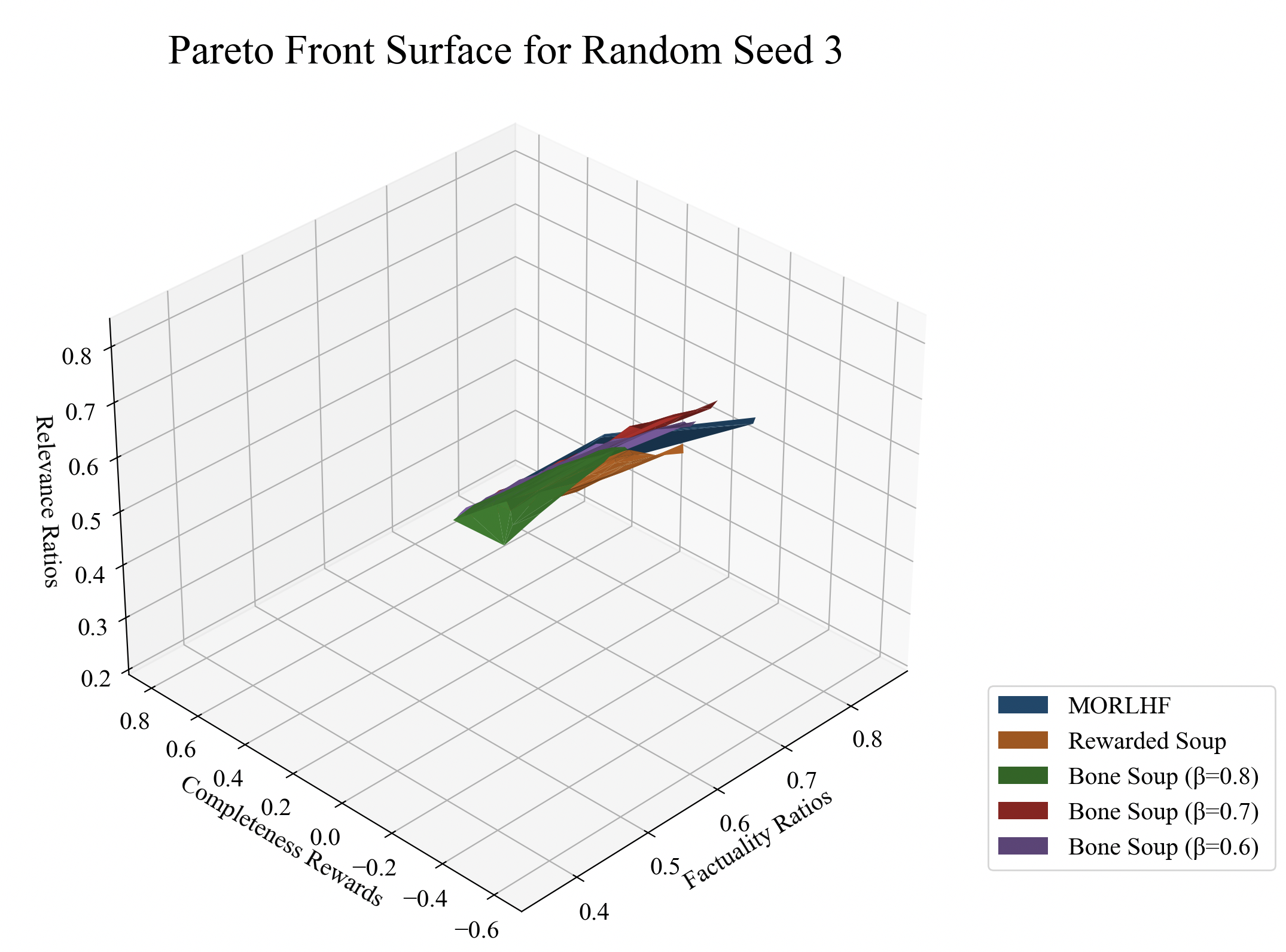} 
        \caption{factuality vs relevance vs completeness (seed 3)}
        \label{fig:seed_overall_b}
    \end{subfigure}
    \caption{Ablation study on different random seeds for the Long-Form QA task, evaluating ``factuality vs relevance vs. completeness''}
    \label{fig:seed_overall}
\end{figure*}

Due to the inherent instability and randomness of PPO optimization\cite{zheng2023secrets, casper2023open, engstrom2019implementation}, we randomly selected three seeds to rigorously assess the robustness of Bone‐Soup. As illustrated in Figure~\ref{fig:seed3}, even with varying seeds, our approach consistently outperforms Rewarded Soups and remains very close to Oracle MORLHF. In several cases—specifically in Figure \ref{fig:seed2_b}, \ref{fig:seed2_c}, \ref{fig:seed3_a}, and \ref{fig:seed3_c}—Bone‐Soup even achieves a Pareto front that surpasses MORLHF, further demonstrating its robustness. We also show the results in three-objective setting in Figure~\ref{fig:seed_overall}.

\subsubsection{RQ5: How Does the Determination of Merging Coefficients Affect the Performance of Bone Soup?}
\begin{figure*}[h]
    \centering
    \begin{subfigure}{0.49\textwidth}
        \includegraphics[width=\linewidth]{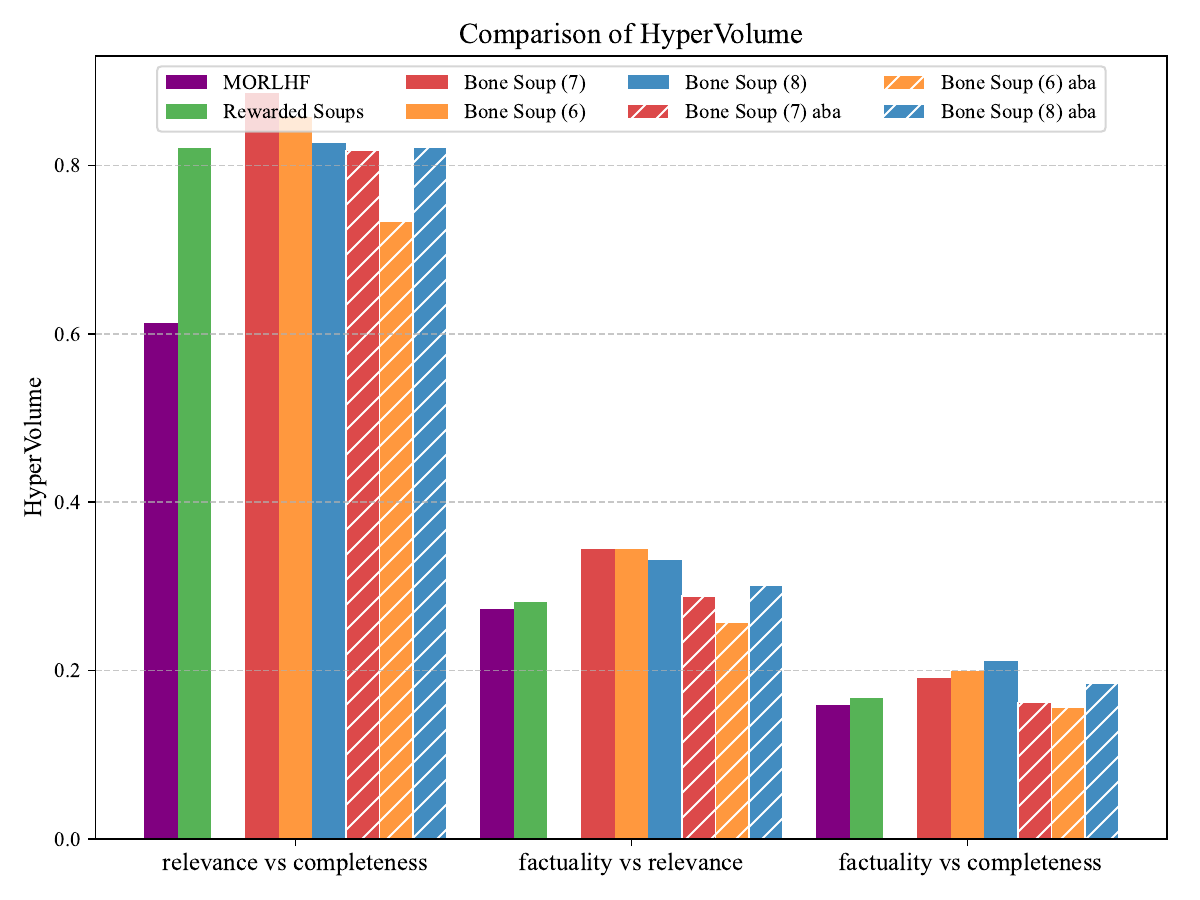}
        \caption{HyperVolume for different coefficient determination approaches.}
        \label{fig:aba_3}
    \end{subfigure}
    \begin{subfigure}{0.49\textwidth}
        \centering
        \includegraphics[width=\linewidth]{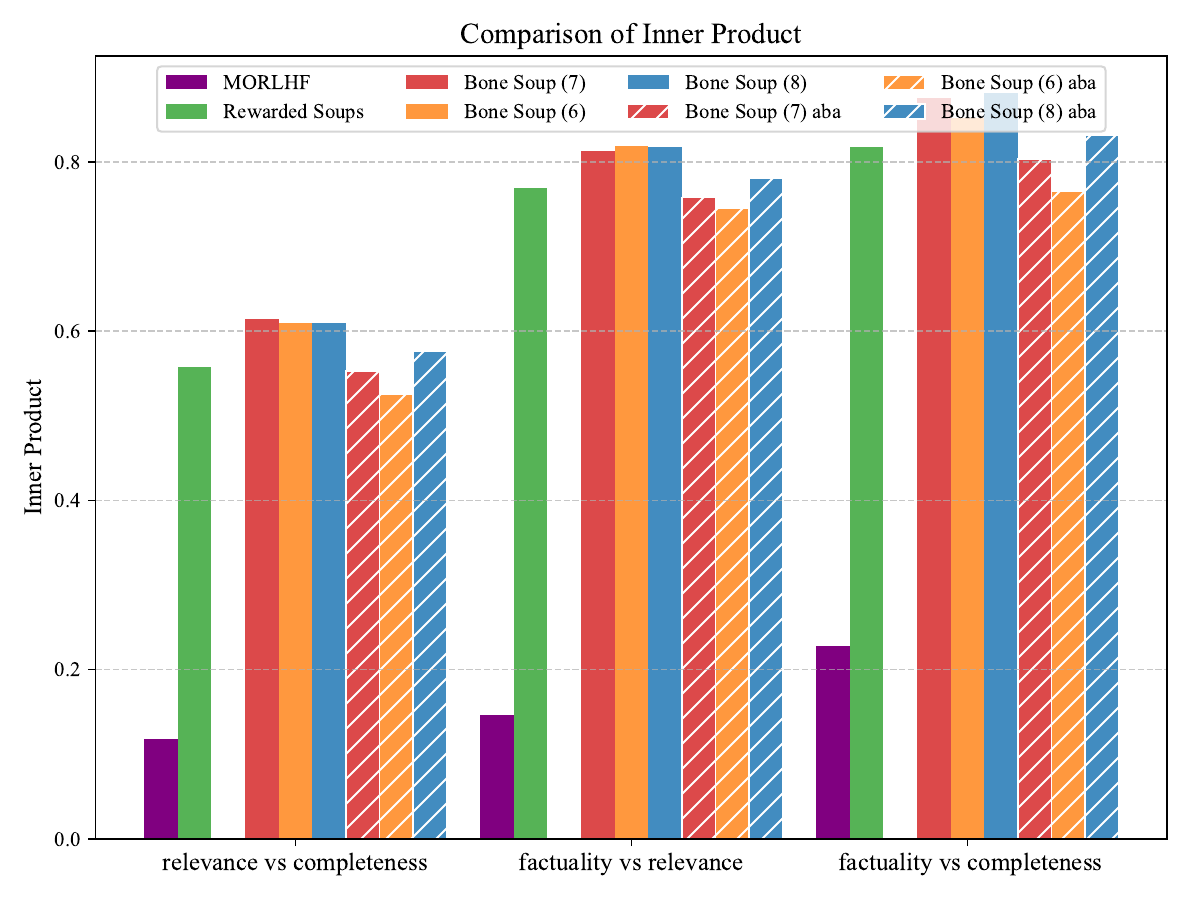}
        \caption{Inner Product for different coefficient determination approaches.}
        \label{fig:aba_4}
    \end{subfigure}
    \caption{Ablation of different approaches for coefficient determination.}
\end{figure*}

We conducted ablation experiments on determination of the merging coefficients. Bone Soup with the suffix 'aba' refers to using Bone Soup to obtain backbone models, and then setting $\bm{\mu} = \bm{\lambda}$ during merging, which means directly mapping the preference. The results in Figure~\ref{fig:aba_3} and~\ref{fig:aba_4} indicate that, even with better backbone models, the merged model still underperforms compared to RS, highlighting the importance of the merging coefficients determination process. This underscores the importance of establishing a strong link between rewards and user preferences, and further validates the critical role of the “soup” stage in our two-stage seek-and-soup approach.

\subsubsection{RQ6: How Do Different Values of $\beta$ Impact the Performance of Bone Soup?} 
\label{sec:app:beta1}
\begin{figure}[h]
    \centering
    \includegraphics[width=\linewidth]{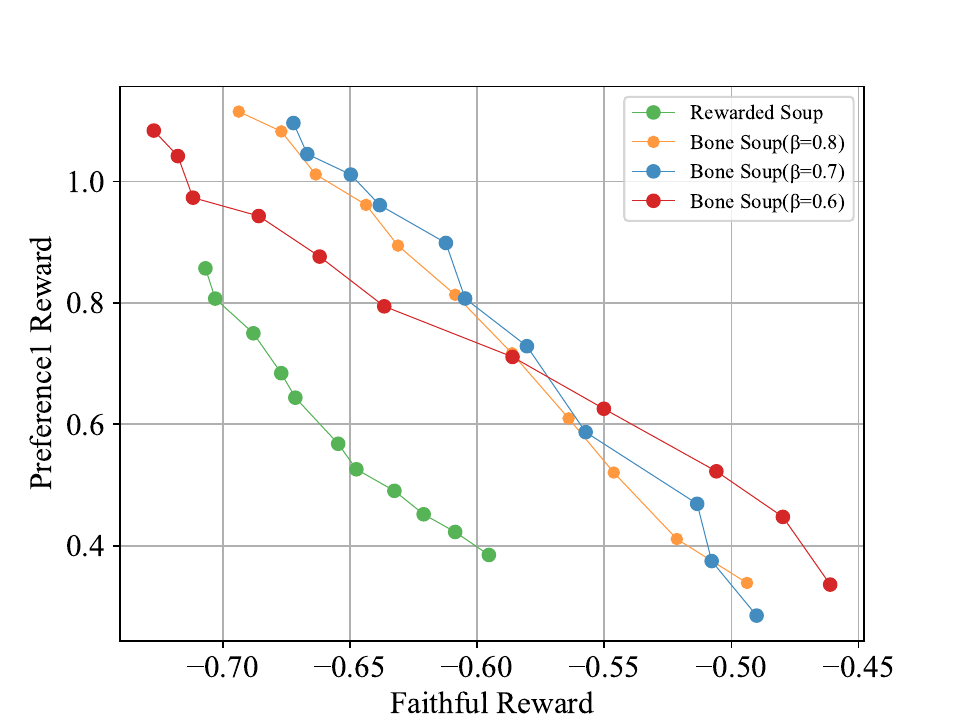}
    
    \caption{Ablation of the impacts of different $\beta$}
    \label{fig:aba_1}
\end{figure}

We conducted experiments on the different trade-offs in Long Form QA task and Reddit Summary task. In Figure~\ref{fig:long} and Figure~\ref{fig:aba_1}, we can see that by varying $\beta$, the resulting front consistently outperforms and dominates RS. This shows that the choice of $\beta$ does not significantly impact the performance of the front and a good $\beta$ only would further improve the upper bound of our method. This demonstrates the robustness in choosing the $\beta$ parameter, thereby underscoring both the lower bound performance and overall robustness of our method.

\begin{figure}[h]
    \centering
    \includegraphics[width=\linewidth]{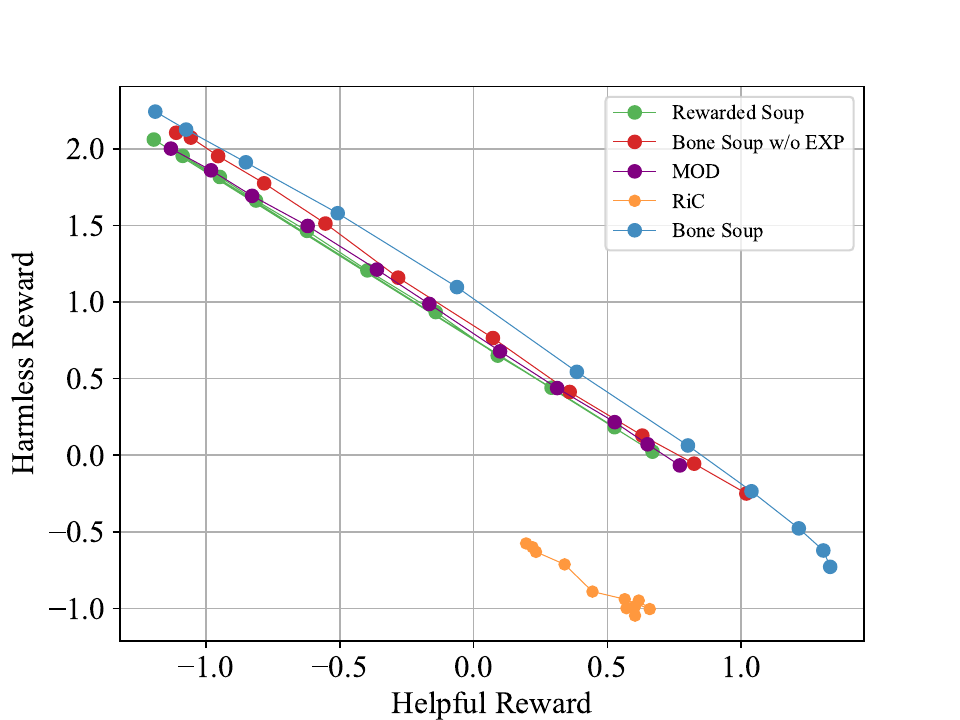}
    
    \caption{Ablation of Extrapolation}
    \label{fig:aba_2}
\end{figure}

\subsubsection{RQ7: What is the Impact of Extrapolation on Performance 
of Bone Soup?} 
We conducted experiments on the trade-off  HH1 (Helpful vs Harmless) on Helpful Assistant dataset. The results in Figure~\ref{fig:aba_2} show that after incorporating interpolation in Equation ~\ref{eq:ex}, the front obtained by BS is indeed improved, indicating that interpolation can further enhance the model's capabilities.

\subsubsection{Comparision with MODPO}
Regarding MODPO~\cite{zhou-etal-2024-beyond}, it is important to note that it is not an adaptive multi-objective generation method in the same sense as Bone Soup, MOD, RiC, or Rewarded Soup. MODPO requires re-training a model for each specific user preference, which is computationally expensive.

We also conduct additional experiments to include MODPO as a baseline for completeness. Since MODPO requires training for each preference, which incurs a huge overhead, we select 
 values of $\alpha \in \{0.1, 0.3, 0.5, 0.7, 0.9\}$ for the user preference $(\alpha, 1-\alpha)$. The experimental results are shown in Table~\ref{tab:modpo_1} and Table~\ref{tab:modpo_2}:

As can be seen, Bone Soup consistently outperforms MODPO across various metrics. Although MODPO and Bone Soup do not share overlapping solutions, MODPO’s Pareto front is significantly smaller, covering a more limited solution space. This limitation is also reflected in its poorer hypervolume performance.

Based on our analysis, we suspect that the relatively weak performance of MODPO can be attributed to the lack of sufficient distinction of the reward in the dataset. As described in the original paper on MODPO, in addition to the margin reward, a second reward model is required to label the preference dataset, where the original "chosen" and "rejected" responses might have low distinction under the current reward model. This reduces the effectiveness of the DPO training process. On the other hand, Bone Soup, Rewarded Soup, and other similar methods utilize reinforcement learning to obtain diverse backbone models. The process of sampling responses in RL helps mitigate the issue of low distinction in rewards. This could be the reason why MODPO performs worse in our experiments.

\begin{table}[htbp]
\centering
\caption{The comparison between Bone Soup and MODPO on Helpful vs Harmeless}
\label{tab:modpo_1}
\resizebox{230.pt}{!}{
\begin{tabular}{lrrrr}
\toprule
Method & HyperVolume & Inner Product & Controllability \\
\midrule
Bone Soup & \textbf{1.16} & \textbf{2.00} & \textbf{1.00} \\
MODPO & 0.93 & -0.57 & 1.00 \\
\bottomrule
\end{tabular}
}
\end{table}

\begin{table}[htbp]
\centering
\caption{The comparison between Bone Soup and MODPO on Faituful vs Preference1}
\label{tab:modpo_2}
\resizebox{230.pt}{!}{
\begin{tabular}{lccc}
\toprule
Method & HyperVolume & Inner Product & Controllability \\
\midrule
Bone Soup & \textbf{1.12} & \textbf{1.89} & \textbf{1.00} \\
modpo & 0.65 & 0.56 & 0.70 \\
\bottomrule
\end{tabular}
}
\end{table}

\subsection{Experiments Setup Details}
\subsubsection{Baselines}
\label{baselines}
We first introduce and compare the three major categories of CMOG methods, followed by a detailed description of the baselines and experimental settings used in our study.

\begin{itemize}
    \item Prompt-based methods: These require the LLM to understand fine-grained task descriptions encoded as numerical prompts (e.g., differences between 0.1 and 0.2). Achieving such precision through supervised fine-tuning is challenging, leading to poor controllability, as noted in \citet{yang2024rewards} and Table~\ref{tab:2} in our paper.
    \item Decoding-based methods: For example, \citet{shi2024decoding} controls outputs by combining logits from multiple aligned models and therefore introduces additional inference time and memory overhead. We believe that the utilization of aligned models only at the logit level is not sufficient. As shown in Table~\ref{tab:2} and Figure~\ref{fig:other} in the paper, decoding-based methods~\cite{shi2024decoding} only improve upon naive merging-based methods(RS) marginally.
    \item Merging-based methods: Our method improves controllability and performance by adjusting parameters at a deeper level and leveraging RLHF models more comprehensively, therefore resulting in significantly better fronts. The detailed analysis can be found in Section~\ref{sec:helpful}.
\end{itemize}

RiC achieves control by adding multiple reward conditions in the prompt and aligns the model through two-stage supervised training. Rewarded Soups (RS)~\cite{rame2024rewarded} trains specializing models separately for each reward and interpolates these models linearly. MOD~\cite{shi2024decoding} also needs to prepare multiple specializing models as RS did and achieve controllable alignment in decoding time by outputting the next token from a linear combination of predictions of all specializing models. For two objective setting, we utilize ten preferences $w_1 \in \{0.0, 0.1, \ldots, 1.0\}$ and $w_2 = 1-w_1$. For the three-objective setting, we uniformly selected $\binom{12}{2}=66$ points from the 3-simplex with a spacing of 0.1.

\subsubsection{Long Form QA task}
Long-form QA\cite{stelmakh2022asqa, min2020ambigqa, wu2024fine, bhat2023investigating, huang2024calibrating} requires the model to generate a complete and comprehensive answer and explanation based on one or more given texts. Since questions often have multiple meanings and can easily cause ambiguity, the required answers need to be complete and multi-faceted.

The FineGrainedRLHF~\cite{wu2024fine} dataset is obtained by reconstructing the ASQA\cite{stelmakh2022asqa} dataset and collecting human feedback, which is publicly available under the Apache 2.0 License. Our use of the dataset is consistent with its intended use. It consists of 2,853 training examples and 500 development examples, forming ``train\_feedback.json'' and ``dev\_feedback.json'' respectively. Each example consists of a question corresponding to four model-predicted outputs sampled from the initial policy model. The feedback includes fine-grained feedback for the first model output and preference feedback for the four model outputs. In the original paper, the authors obtained 1,000 samples from the ASQA dataset to form ``train\_1k.json'' for supervised training of the original policy model. We follow the setup of \citet{wu2024fine}, first performing supervised training on the initial policy model, and then using the reward models provided by \citet{wu2024fine} to conduct PPO~\cite{schulman2017proximal} training.

\textbf{Reward Models.} \citet{wu2024fine} provides rule-based reward models of three different granularities (\textbf{sub-sentence, sentence, full sequence}) based on error types. These reward models all use the encoder-only Longformer-base~\cite{beltagy2020longformer} as the backbone. Suppose the input format of the reward model is "question: q context: $p_1$ \: $p_2$ ... answer: [sep] $y_1^k$ [sep] $y_2^k$ ...", where k represents different granularity levels corresponding to different rewards $R_k$——for example, the relevance reward corresponds to sub-sentence granularity. \citet{wu2024fine} uses token-level classification loss to predict whether the segment of that granularity before each [sep] token contains errors corresponding to that reward. Therefore, the reward is set in a rule-based manner: for different rewards, it judges the different segments; if such an error exists, -1 is given at the [sep] position; otherwise, +1.

\textbf{$\bm{R}_1$ Relevance Reward}: . $\bm{R}_1$ is designed to predict whether there are errors such as irrelevance, repetition, or incoherence at the sub-sentence level. $\bm{R}_1$ gives a reward at each [sep] position; if there is no error, +1 is given at that position; otherwise, -1.

\textbf{$\bm{R}_2$ Factuality Reward}: $\bm{R}_2$ is designed to predict whether there are factual errors such as incorrect or unverifiable information at the sentence level. $\bm{R}_2$ gives a reward at each [sep] position; if there is no error, +1 is given at that position; otherwise, -1.

\textbf{$\bm{R}_3$: Completeness Reward}: $\bm{R}_3$ is designed to predict whether there are errors of information incompleteness at the full-sequence level. $\bm{R}_3$ gives a reward at each [sep] position; if there is no error, +1 is given at that position; otherwise, -1.

\subsubsection{Helpful Assistant Task}
The Helpful Assistant task requires the model to generate appropriate responses based on a given user conversation history, ensuring the response is as helpful as possible while maintaining safety. We use the hh-rlhf dataset for training and evaluation. The hh-rlhf dataset consists of 160k prompts, responses, and corresponding human annotations, which is publicly available under the MIT License. Our use of the dataset is consistent with its intended use. Additionally, we use three open-source reward models following \citet{yang2024rewards}

to assess the helpfulness, harmlessness, and humor of the responses generated by the model.
The backbones of first two reward models are GPT-2 model\cite{radford2019language}. The two reward models were trained on the Anthropic/hh-rlhf dataset using pair-wise feedback. The harmless reward model achieves a test set accuracy of 0.73698 and the helpful reward model achieves an accuracy of 0.72621 on the test set. The humor reward model is a fine-tuned version of distilbert-base-uncased~\cite{sanh2019distilbert} on a joke/no-joke dataset to detect humor. And there is no extra prompt for this task.

\subsubsection{Reddit Summary Task}
The Reddit Summary task~\cite{stiennon2020learning} focuses on summarizing Reddit posts, aiming to produce concise and coherent summaries that effectively capture the main content of the post. The dataset consists of Reddit threads, where the input comprises a post's title and body text, and the output is a human-written summary, which is publicly available under the MIT License. Our use of the dataset is consistent with its intended use. And the prefix of the prompt is "Generate a one-sentence summary of this post." according to \cite{yang2024rewards}. 
We use two open-source reward models \cite{yang2024rewards}

to assess the quality of the summary generated from two different aspects.

\subsubsection{Parameters Setting Details}
We list the values of parameters used in the experiment in Table~\ref{tab:param1} and Table~\ref{tab:param2} corresponding to Helpful Assitant, Reddit Summary Task, and Long Form QA task respectively.

\subsection{Evaluation Metrics Details}
\label{sec:app:metrics}

Numerical metrics include \textbf{hypervolume indicator}~\cite{zitzler2003performance}, \textbf{Inner Product}~\cite{zhong2024panacea}, \textbf{Sparsity(SP)}~\cite{deb2002fast, zhong2024panacea}, \textbf{Spacing}~\cite{schott1995fault, zhong2024panacea}, \textbf{controllability} and the \textbf{cardinality} of the Pareto front. 
\textbf{It is worth noting that a better front will also lead to greater Sparsity and Spacing since the coverage of that front is usually larger causing the bigger Sparsity and Spacing.} Therefore, we will use\textbf{ HV, Inner Product and controllability} as our main metrics and other metrics are for reference when the main metrics are very close.

Since Controllability is the fundamental and most important aspect of implementing CMOG, we will formally define it below.

\begin{definition}[Controllability]
Controllability measures the degree to which the model's output aligns with the desired human preference $\bm \mu$. It is calculated by exhaustively enumerating all pairs of preferences $(\bm{\mu_i}, \bm{\mu_j})$, and checking if their relative order $({\mu_{i,k}}, \mu_{j,k})$ matches the relative order of the evaluation~(rewards $r$) $(r_k(\mathcal{S}_i), r_k(\mathcal{S}_j))$ in each dimension k where $\mathcal{S}_i$ is the model solution obtained by CMOG approaches corresponding to the preference $\bm \mu_i$. For each pair~$(\bm{\mu_i}, \bm{\mu_j})$, if the above condition holds for all $k$ dimensions, we increment the controllability score by 1.
The final controllability score is normalized by dividing by the total number of solution pairs. The controllability score \(C\) is defined as:
\begin{equation}
\begin{aligned}
C = \frac{1}{N(N-1)} &\sum_{i \neq j} \mathbf{1}[\bigwedge_{k=1}^{n} \text{sign}\left(\mu_{i,k} - \mu_{j,k}\right) = \\
& \text{sign}\left(r_k(\mathcal{S}_{i}) - r_k\left(\mathcal{S}_{j}\right) \right) ],
\end{aligned}
\end{equation}
where $N$ is the total number of solutions, and $\mathbf{1}(\cdot)$ is the indicator function. The closer the score $C$ is to 1, the better the controllability of the model.
\end{definition}

\textbf{1.Hypervolume}

Hypervolume is a key performance indicator in multi-objective optimization, used to measure the volume of the space dominated by a set of solutions in the objective space. The hypervolume is defined as:

\begin{equation}
HV(S) = \text{Volume}\left( \bigcup_{i=1}^{n} \left[ \mathbf{r}, \mathbf{f}(x_i) \right] \right)
\end{equation}

where $\left[ \mathbf{r}, \mathbf{f}(x_i) \right]$ represents the hyper-rectangle region between the reference point $\mathbf{r}$ and each solution $\mathbf{f}(x_i)$. Hypervolume is widely used to evaluate the performance of multi-objective optimization algorithms. A larger hypervolume indicates a better coverage of the objective space.

\textbf{2.Inner Product}

The inner product between the preference vector and the corresponding reward vector serves as a metric for measuring their correspondence. From another perspective, this can be interpreted as a weighted sum of rewards, where the preference vector reflects the emphasis on different reward components.

Mathematically, this can be expressed as:

\begin{equation}
IP(\mathbf{\mu}, \mathbf{r}) = \sum_{i=1}^{n} \mu_i \cdot r_i
\end{equation}

where $ \mathbf{\mu} = [\mu_1, \mu_2, \dots, \mu_n] $ is the preference vector and $ \mathbf{r} = [r_1, r_2, \dots, r_n] $ is the corresponding reward vector. The inner product quantifies the alignment between the preference and reward, with higher values indicating stronger alignment between them.

\textbf{3.Sparsity}

Sparsity measures the variation between solutions corresponding to the consecutive preference vectors \cite{deb2002fast, zhong2024panacea}. It is defined as the average squared Euclidean distance between adjacent vectors. A smaller sparsity value indicates smoother transitions between successive rewards, which is desirable in our context. \textbf{However, due to the huge cost of evaluating solutions, we can only obtain a limited number of solutions and therefore the evaluation of Sparsity is less convincing.}

\begin{equation}
    \mathrm{Sparsity} = \frac{1}{n-1} \| \mathbf{r}_i - \mathbf{r}_{i-1}\|^2 
\end{equation}

\textbf{4.Spacing}

We follow \citet{zhong2024panacea} to introduce the Spacing metric to evaluate the front. The Spacing metric evaluates the variance of the minimum distance between solutions(corresponding reward vectors). Lower values indicate a better Pareto front but with the same limitations as Sparsity.
\begin{equation}
    \mathrm{Spacing} = \sqrt{\frac{1}{N}\sum_{i=1}^{N} (d_i-p)^2},
\end{equation}
where $d_i = \mathrm{min}\{||r_i-r_j||\}$ and $p = \frac{1}{N} \sum_{i=1}^N p_i$.

\subsection{Discussions}
\textbf{Computational Complexity Analysis}
The primary computational cost of our approach occurs during the training phase. Once training is completed, the inference stage can adapt to any user preference without additional overhead. Unlike MOD~\cite{shi2024decoding} and MODPO~\cite{zhou-etal-2024-beyond}, there is no extra cost during inference. The process of determining merging coefficients only requires solving a linear equation, which incurs no additional computational expense.

On the other hand, although we propose using a small-scale selection to choose the optimal 
$\beta \in \{0.8,0.7,0.6\}$, the robustness of the constructed matrix (see Appenidx~\ref{appendix:rq3}) means we are not overly dependent on $\beta$ and can get rids of this additional expense. In Appendix~\ref{sec:app:beta1}, we discuss the impact of varying $\beta$ on Bone Soup, and the conclusion is that, regardless of the $\beta$ value, the resulting Pareto front consistently outperforms and dominates other baselines. Our insight is that including marginal reward could lead to better backbone construction. Therefore, the simplest approach is to select any $\beta$, which results in \textbf{computational costs identical to those of standard Rewarded Soup, with no additional overhead}. If one aims to achieve the optimal performance, let the computational cost of Rewarded Soup be $C$. Then, selecting optimal $\beta$ introduces an additional cost of $0.2 \times C \times n$, where $n \leq 3$ is the number of possible choices for $\beta$ and as the small-scale selection trained the model with 20\% total steps.

\textbf{Negative Merging Coefficients. }

Solving the linear system may result in negative values, which could lead to negative interpolation or said extrapolation~\cite{ilharco2022editing,zheng2024weak} of the backbone models. 

Previous works have discussed improving model performance through extrapolation techniques~\cite{zheng2024weak}, or by deliberately weakening the initial SFT model to facilitate unlearning~\cite{ilharco2022editing}. In these approaches, negative merging coefficients were determined through trial-and-error methods using a validation set.

However, in our approach, we avoid the cumbersome and somewhat unnatural trial-and-error process by directly solving linear equations to establish a clear mapping between the backbone models' rewards and user preferences. This not only simplifies the process but can also be regarded as an interpretable extrapolation. Our method Bone Soup of constructing non-orthogonal bases and then performing interpolation can thus be seen as a form of extrapolation. Unlike the naive merging method, where coefficients are constrained to be nonnegative, the presence of negative coefficients extends beyond the original model space. 
More importantly, the ability to interpret the negative coefficients selected offers the potential to improve performance in regions that would otherwise be inaccessible through standard interpolation techniques.

\textbf{Model Merging in Multi-Objective and Multi-Task Setting. }
\label{subsec:discussion}
Here, we emphasize the distinction between obtaining a \emph{Pareto front} and \emph{single model}. Most current research~\cite{wortsman2022model, rame2024rewarded, tang2024merging, yu2024language, yadav2024ties, yang2023adamerging, wang2024localizing, ilharco2022editing} primarily focuses on obtaining a single model that, through merging, possesses the capabilities of multiple models. This approach works in multi-task scenarios because the interference between various tasks is present but often not strong enough to pose significant challenges.

However, in multi-objective optimization, numerous objectives are inherently conflicting or compromised. For instance, in QA tasks, relevance and completeness are often at odds~\cite{wu2024fine}: a complete answer is likely to include some irrelevant content, while a highly relevant answer may be too narrow, resulting in incomplete responses. Similarly, in typical alignment tasks, objectives like helpfulness and harmlessness~\cite{dai2023safe,bai2022training,ganguli2022red} frequently conflict, making it difficult to achieve both fully. In such cases, it is preferable to aim for a Pareto front, where the points on the front represent non-dominated and optimal solutions. Our goal is not only to find this front but to ensure it is as expansive as possible, with widely dispersed points, thereby covering a broad range of trade-offs between competing objectives.

\begin{table}[ht]
\centering
\resizebox{230pt}{!}{
\begin{tabular}{l c}
\toprule

\multicolumn{2}{c}{\textbf{Model and Task Settings}} \\ \hline
\text{Model} & LLaMA-2-7B  \\
\text{Helpful Assistant Task} & 128 tokens \\ 
\text{Reddit Summary Task} & 48 tokens \\ \hline
\multicolumn{2}{c}{\textbf{LoRA Settings}} \\ \hline
\text{Rank} & 64 \\ 
\text{Alpha} & 128 \\
\text{LoRA Dropout} & 0.05 \\ \hline
\multicolumn{2}{c}{\textbf{SFT Training}} \\ \hline
\text{Fine-tune Steps} & 60k steps \\ 
\text{Initial Learning Rate} & 1.41e-4 \\ 
\text{Learning Rate Decay} & Linear \\ \hline
\multicolumn{2}{c}{\textbf{RL Training (PPO)}} \\ \hline
\text{Implementation} & \text{trl 0.8} \\
\text{Initial KL Penalty Coefficient} & 0.2 \\ 
\text{Learning Rate} & 1e-5 \\ 
\text{GAE Lambda} & 0.95 \\ 
\text{Discount Factor (Gamma)} & 1 \\ 
\text{Clip Range} & 0.2 \\ 
\text{Maximum Gradient Norm} & 0.5 \\ 
\text{Sampling Strategy} & nucleus sampling,  \\ 
& top\_p=0.1 \\
\text{Sampling Temperature} & 0.7 \\
\text{Target KL Early Stop} & 5 \\ 
\text{Training Epochs} & 1 \\ 
\text{Batch Size} & 64 \\ 
\text{Mini Batch Size} & 1 \\ 
\text{Optimization Epochs per Batch} & 4 \\ \hline

\bottomrule
\end{tabular}
}
\caption{Helpful Assistant and Reddit Summary Task Settings and Parrameters}
\label{tab:param1}
\end{table}

\begin{table}[t]
\centering
\resizebox{230pt}{!}{
\begin{tabular}{l c}
\toprule

\multicolumn{2}{c}{\textbf{Model and Task Settings}} \\ \hline
Model & T5-large \\
Input Length Limit & 1024 \\
Max Generation Tokens & 200 \\ \hline
\multicolumn{2}{c}{\textbf{Data Splits}} \\ \hline
Dataset & QA-FEEDBACK \\
Splits & SFT, Train, Test \\
SFT Training Data &  1000 examples \\
PPO Training Data & 2853 examples \\
Test Data & 500 examples \\ \hline
\multicolumn{2}{c}{\textbf{SFT Training}} \\ \hline
Epochs & 10 \\ 
Batch Size & 32 \\
Learning Rate & 5e-5 \\ \hline
\multicolumn{2}{c}{\textbf{PPO Training Parameters}} \\ \hline
Episodes & 80,000 \\
PPO Epoch per Rollout & 4 \\
Initial Learning Rate & 1e-5 \\
Learning Rate Decay & Linear \\
Early Stop KL Threshold & 10 \\
GAE Lambda & 0.95 \\
KL Coefficient & 0.3 \\
Discount Factor (Gamma) & 1 \\
Clip Range & 0.2 \\ 
Sampling Strategy & Top-k sampling(k=20) \\
Temperature & 0.7 \\ \hline
\multicolumn{2}{c}{\textbf{LoRA Settings}} \\ \hline
Rank & 32 \\
Alpha & 32 \\ \bottomrule
\end{tabular}
}
\caption{Long Form QA Task Settings and Parameters}
\label{tab:param2}
\end{table}

\onecolumn
\begin{tcolorbox}

\#\#\# {\bf System Prompt:}\par 
You are a impartial judge for checking the quality of the answer.\\\par
\#\#\# {\bf User Prompt:} \par
[System] \par
We kindly request your feedback on the performance of two AI assistants in response to the user question presented below. Act as an impartial judge and evaluate only the factuality of the response provided by each assistant. Rate each assistant on a scale of 1 to 10, where a higher score signifies a more factually accurate response. Try to avoid giving the same score.
\\\par
Your evaluation should focus solely on the factual accuracy of the response. When assessing factuality, please check whether the response is consistent with the provided context, whether it is correct, and whether the information is verifiable. A higher score should reflect better adherence to facts and context.
\\\par
The question and answers are as follows:
\\\par
[Question]
\par
\textcolor{blue}{\{question\}}
\\\par
[The Start of Assistant 1's Answer]
\par
\textcolor{blue}{\{answer1\}}
\\\par
[The End of Assistant 1's Answer]
\\\par
[The Start of Assistant 2's Answer]
\par
\textcolor{blue}{\{answer2\}}
\\\par
[The End of Assistant 2's Answer]
\\\par
[System]
\par
Start by outputting a single line containing only two values indicating the scores for Assistant 1 and 2, respectively. The two scores should be separated by a space. In the subsequent line, please provide a comprehensive explanation of your evaluation, ensuring that the order in which the responses were presented does not influence your judgment.
\\\par
[Answer]
\\\par
\end{tcolorbox}
\captionof{figure}{Prompt template for GPT-4 to evaluate Factuality.}

\begin{tcolorbox}

\#\#\# {\bf System Prompt:}\par 
You are a impartial judge for checking the quality of the answer.\\\par
\#\#\# {\bf User Prompt:} \par
[System] \par
We kindly request your feedback on the performance of two AI assistants in response to the user question presented below. Act as an impartial judge and evaluate only the relevance of the response provided by each assistant. Rate each assistant on a scale of 1 to 10, where a higher score signifies a more relevant response. Try to avoid giving the same score. 
\\\par
Your evaluation should focus solely on whether the response is relevant to the context and the question, whether it is logically coherent, and whether it is concise and to the point. When assessing relevance, please check if the response directly answers the question, aligns well with the provided context, and avoids unnecessary or off-topic information. 
\\\par
The question and answers are as follows:
\\\par
[Question]
\par
\textcolor{blue}{\{question\}}
\\\par
[The Start of Assistant 1's Answer]
\par
\textcolor{blue}{\{answer1\}}
\\\par
[The End of Assistant 1's Answer]
\\\par
[The Start of Assistant 2's Answer]
\par
\textcolor{blue}{\{answer2\}}
\\\par
[The End of Assistant 2's Answer]
\\\par
[System]
\par
Start by outputting a single line containing only two values indicating the scores for Assistant 1 and 2, respectively. The two scores should be separated by a space. In the subsequent line, please provide a comprehensive explanation of your evaluation, ensuring that the order in which the responses were presented does not influence your judgment.
\\\par
[Answer]
\\\par

\end{tcolorbox}
\captionof{figure}{Prompt template for GPT-4 to evaluate Relevance.}

\newpage
\twocolumn

\end{document}